\documentclass[sigconf,nonacm]{acmart}

\usepackage{subcaption}

\usepackage{listings}
\usepackage{xcolor}
\usepackage{siunitx}

\usepackage{amsmath}
\usepackage{mathtools}
\usepackage{amsthm}
\usepackage{cleveref}

\newcommand{\cut}[1]{}

\setcopyright{none}               
\renewcommand\footnotetextcopyrightpermission[1]{} 
\begin{document}

\title{What is Your Agent's GPA? A Framework for Evaluating Agent Goal-Plan-Action Alignment}

\author{Allison Sihan Jia}
\email{allison.jia@snowflake.com}
\affiliation{%
  \institution{Snowflake}
  \city{Menlo Park}
  \country{USA}
}

\author{Daniel Huang}
\affiliation{%
  \institution{Snowflake}
  \city{Menlo Park}
  \country{USA}
}

\author{Nikhil Vytla}
\affiliation{%
  \institution{Snowflake}
  \city{Menlo Park}
  \country{USA}
}

\author{Seung Won Wilson Yoo}
\affiliation{%
  \institution{Snowflake}
  \city{Menlo Park}
  \country{USA}
}

\author{Nirvika Choudhury}
\affiliation{%
  \institution{BASIS Independent Silicon Valley Upper School}
  \city{San Jose}
  \country{USA}
}

\author{Shayak Sen}
\affiliation{%
  \institution{Snowflake}
  \city{Menlo Park}
  \country{USA}
}

\author{John C. Mitchell}
\affiliation{%
  \institution{Department of Computer Science, Stanford University}
  \city{Stanford}
  \country{USA}
}
\affiliation{%
  \institution{Snowflake}
  \city{Menlo Park}
  \country{USA}
}

\author{Anupam Datta}
\affiliation{%
  \institution{Snowflake}
  \city{Menlo Park}
  \country{USA}
}

\renewcommand{\shortauthors}{Jia et al.}

\begin{abstract}
We introduce the \textbf{Agent GPA} (\textbf{G}oal-\textbf{P}lan-\textbf{A}ction) framework, driven by the fundamental insight that critical agent failures emerge at the intersections of setting goals, devising plans, and executing actions. We operationalize the framework with a factorized suite of LLM judges designed to measure distinct elements of Goal-Plan-Act alignment. To make this methodology scalable and generalizable across diverse agent architectures and datasets, we use state-of-the-art automated prompt optimization techniques to systematically generate domain-specific evaluation criteria. We validate this approach across three benchmarks: a multi-agent research setting (TRAIL/GAIA), a single coding agent setting (TRAIL/SWE-bench), and a private, enterprise data-agent setting (Snowflake Intelligence). Extensive evaluation on TRAIL/GAIA demonstrates the core validity of the framework, which identifies a broad range of agent failures (95\% of human-annotated errors), localizes errors to enable targeted debugging (86\% of human-annotated errors), and exhibits strong agreement with human evaluators. Crucially, by applying our automated methodology to both public datasets, we demonstrate that our GPA judges generally achieve the highest error coverage (ranging from 76\% to 86\%) in comparison to manual prompting approaches. We also leverage an evolutionary coding agent to improve judge consistency by up to 38\% through iterative refinement of evaluation rubrics. Overall, Agent GPA provides a rigorous and generalizable paradigm for targeted agent evaluation.
\end{abstract}

\maketitle

\section{Introduction}

The power of agentic AI systems comes from their ability to autonomously reflect, plan multiple steps, call various tools, and leverage collaboration between agents to achieve complex goals \cite{agentic_patterns_ng}. While platforms for building agentic AI systems have advanced rapidly, the deployment of these systems in real use cases requires robust evaluation methods. Early ``step-level'' evaluations often focus only on advances or shortcomings of the last step, overlooking end-to-end performance \cite{yehudai2025survey}. Other approaches rely on ground-truth sources annotated by human experts. While valuable, these require considerable and often unrealistic effort to curate \cite{chen2021humaneval,jimenez2023swe,wei2025plangenllms,mohammadi2025evaluation}. 
Additionally, many existing approaches emphasize evaluating the final outcome, with little actionable insight into root causes of failure or opportunities for targeted improvement \cite{chiang2024chatbotarena,yehudai2025survey}.

We propose a comprehensive evaluation of agent systems, referred to as \emph{agents} for simplicity, based on their natural operational dynamics. Just as agents set goals, devise plans, and execute actions, constructive evaluation should analyze failures within and between each of these components. Therefore, we introduce the {\bf Agent GPA} or {\bf G}oal-{\bf P}lan-{\bf A}ction evaluation framework. Our framework introduces the metrics: Goal Fulfillment, Plan Quality, Tool Selection, Plan Adherence, Tool Calling, Logical Consistency, and Execution Efficiency (see Figure \ref{fig:gpa_evals_reorg}). While these GPA alignment metrics can also be computed by human evaluators, we introduce a suite of automated
reference-free, LLM-as-a-Judge (LLM judge) evaluators to scale the framework. To validate the power and scalability of Agent GPA, we evaluate our LLM judges across three diverse datasets representing distinct agent architectures and domains: the public TRAIL/GAIA benchmark (general multi-agent reasoning), the public TRAIL/SWE-bench benchmark (single coding agent), and an internal Snowflake Intelligence dataset (commercial production-grade data analytics agent). We summarize our contributions below: 

    \vskip 1ex\noindent {\textbf{1. GPA Framework Validity}} \textit{\textbf{Claim:}} The Agent GPA framework provides a systematic way to categorize, identify, and localize a broad range of agent failures. \\
\textit{\textbf{Results:}} By utilizing a factorized suite of specialized judges rather than a single monolithic evaluator, our framework achieved higher performance than baseline approaches. On the TRAIL/GAIA test set, our GPA judges successfully identified 95.0\% (267/281) of human-labeled errors and correctly localized 85.8\% (241/281) of them. In contrast, a standard monolithic baseline judge achieved at best 54.8\% (154/281) identification and 49.1\% (138/281) localization, even when provided with architectural context about the agent system being evaluated. Lastly, the GPA LLM judges demonstrated strong alignment with human judgment across domains, averaging 73.1\% LLM-human accuracy agreement on the public TRAIL/GAIA benchmark and 82.3\% LLM-human accuracy agreement on the Snowflake Intelligence dataset. 

\vskip 1ex\noindent\textbf{2. Generalizability:} \\
\textit{\textbf{Claim:}} Agent GPA LLM judges scale effectively to new agent architectures and domains by leveraging state-of-the-art prompt optimization techniques to automate custom prompt retuning. \\
\textit{\textbf{Results:}} In addition to yielding a generalizable methodology for creating domain-specific criteria, these prompt optimization techniques broadly led to comparable, if not better, evaluation performance compared to traditional manual prompting techniques. Specifically, on the TRAIL/GAIA dataset, GPA judges optimized with the GEPA prompt optimizer~\cite{agrawal2025gepareflectivepromptevolution} matched or exceeded the performance of manually crafted GPA judges, yielding significant improvements in dimensions like Plan Quality (boosting recall from 52.4\% to 76.2\%) and Logical Consistency (boosting recall from 83.6\% to 88.8\%). On the previously untested TRAIL/SWE-bench dataset, GEPA-optimized GPA judges also demonstrated improvements over manually crafted GPA judges, jumping from 49.8\% to 83.1\% for Logical Consistency (LC) and from 55.5\% to 84.7\% for Tool Calling (TC). 

\vskip 1ex\noindent\textbf{3. Consistency} \\
\textit{\textbf{Claim:}} Agent GPA LLM Judges exhibit strong consistency across repeated evaluations, and their reliability can be further enhanced with evolutionary prompt optimization. \\
\textit{\textbf{Results:}} On the TRAIL/GAIA dataset, independent runs of LLM judges on same traces produced identical scores with substantial inter-rater agreement, with an average Krippendorff's $\alpha$ of 0.77. This stability strengthens our judges' reliability as automated evaluators given general evaluation prompts, reducing the need for redundant human review. Beyond baseline consistency, we applied OpenEvolve~\cite{sharma2025openevolve}-an LLM-guided evolutionary optimization framework-to further improve judge reliability by optimizing prompts with standard deviation of scores as the fitness objective. Across optimization runs, we observed that evolved prompts frequently achieved lower variance by introducing more granular rubrics and explicit scoring criteria, providing judges with clearer decision boundaries. Most notably, for the Plan Quality judge-which exhibited the lowest baseline consistency-OpenEvolve optimization improved Krippendorff's $\alpha$ from 0.628 to 0.814. These results suggest that evolutionary prompt optimization is a promising direction for systematically reducing evaluation noise in LLM-based assessment systems.

\section{Related Work}

Building LLM agents requires establishing goals, formulating plans, and executing actions. However, existing evaluation methods tend to focus on these elements in isolation and often rely heavily on ground-truth references, limiting their scalability and usefulness for open-ended tasks \cite{mohammadi2025evaluation,chang2024survey}.

\vskip 1ex\noindent \textbf{Goal Progression and Fulfillment:} 
Existing approaches to evaluating goal progression primarily measure factual correctness by comparing an agent's final output against a known reference answer \cite{nvidia_agent_evals}. However, this often constrains scalability, as labeled final answers are both expensive and rarely available. Nevertheless, measuring goal fulfillment remains essential, as agents may deviate from their original objectives over long interactions when their context windows becomes saturated with new information. \cite{arike2025technical}'s stock trading agent simulation demonstrated that all evaluated agents exhibited some goal drift, particularly when faced with competing objectives or when switching between different goals. This underscores an important need for reference-free methods that can evaluate goal fulfillment, even in the absence of ground-truth answers.

\vskip 1ex\noindent \textbf{Planning via Reasoning Traces:} 
Whereas many early agents operated without explicit plans and simply executed the next greedy step, recent work shows that separating planning from execution can yield significant performance gains. For example, Plan-and-Act \cite{erdogan2025plan} achieves state-of-the-art performance on a web navigation benchmark by translating high-level plan steps to lower-level, environment-specific actions, and AdaPlanner \cite{sun2023adaplanner} demonstrates the value of adaptive plan refinement using environmental feedback. While this has motivated the creation of plan evaluations, current methods rely on validation with a simulation verifier, human annotation, or ground-truth \cite{wei2025plangenllms}. One such example, Plancraft \cite{dagan2024plancraft}, quantitatively evaluates a Minecraft agent's proposed plan against a gold standard planner by measuring the difference between the number of actions in an agent's successful plan and the optimal number of actions. As more systems adopt explicit planning, developing reference-free evaluations for plan quality and plan adherence will be critical \cite{wei2025plangenllms}. 

\vskip 1ex\noindent \textbf{Execution via Action Traces:}
Recent work (\cite{liu2024agentbench} ) has illustrated that final outcomes alone are insufficient to evaluate execution success, since a superficially correct result can mask unsafe or invalid actions.
To address this, existing solutions such as Vertex AI's evaluations and LangChain's AgentEvals validate an agent's trace against a reference trajectory with the expected sequence of tool calls or steps \cite{vertex_ai,langchain_agent_evals}. However, AgentRewardBench \cite{lu2025agentrewardbench} demonstrates that this rules-based evaluation of agents is too rigid and often artificially penalizes valid trajectories that differ from predefined, golden paths.

Beyond performance evaluation, execution traces can also be used for debugging. The TRAIL benchmark \cite{deshpande2025trail} provides annotated traces from datasets such as GAIA \cite{mialon2024gaia} and SWE-bench \cite{jimenez2023swe}, tasking LLMs to find errors across categories such as goal deviation and hallucination. Similarly, MAST \cite{cemri2025multiagentllmsystemsfail} proposes a taxonomy of broad failure modes specific to multi-agent systems. However, such statically-defined taxonomies often classify error symptoms rather than the agent's operational breakdown. This can lead to ambiguous classifications that obscure the root cause of the failure, such as conflating a bad plan with bad execution. Emerging frameworks that record and replay traces for iterative refinement also point towards a path for more reliable and debuggable agents \cite{feng2025get}. In addition to accuracy, agent evaluation (especially in enterprise settings) must consider other factors such as operational cost and efficiency \cite{kapoor2025matter}. This highlights a critical gap for a granular, reference-free framework that can isolate and identify operational breakdowns.

\vskip 1ex\noindent \textbf{LLM Judges:} As a scalable alternative to ground-truth matching, LLM judges have been explored as a reference-free agent evaluators. However, these trajectory evaluations often rely on a single, monolithic judge with the same prompt to evaluate traces generated by different agents \cite{lee_2025_evaluating,langchain_agent_evals}. 
AgentRewardBench \cite{lu2025agentrewardbench} notes that while rules-based methods underestimate success, LLM judge evaluations often overestimate success and miss important details when asked to process long, complex traces. Similarly, TRAIL reports that even the strongest LLMs achieve only 11\% accuracy on their task due to context-length limits and reasoning difficulty,  illustrating the fragility of asking a single LLM judge to simultaneously identify, localize, and classify errors \cite{deshpande2025trail}. Additionally, while existing industrial offerings \cite{arize_agent_evals} evaluate components such as steps, routers, and paths, it is unclear how their reported results perform on benchmarks, making their alignment with broader measures of agent operational performance harder to assess. These findings suggest that decomposing evaluation into specialized judges with custom instructions may provide more reliable and interpretable assessments, and comparative studies are needed to establish the validity and generalizability of these judges. 

\section{Agent Goal-Plan-Action (GPA) Framework}
\begin{figure}[htbp] \centering \includegraphics[width=\columnwidth]{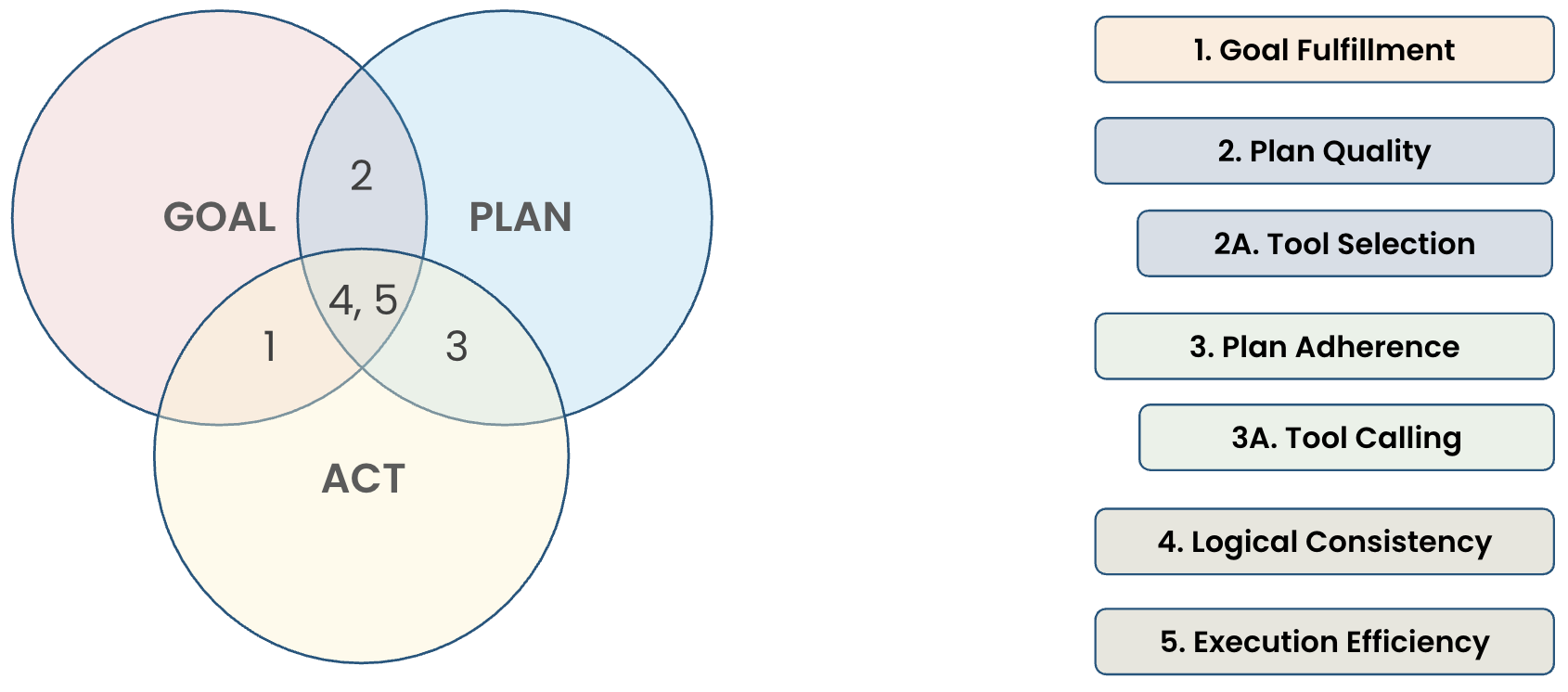} \caption{The GPA Framework: Failures emerge at the intersections of Goal, Plan, and Action.} \label{fig:gpa_evals_reorg} \end{figure}

The Agent GPA framework is a novel conceptual framework designed to diagnose AI agent failures as breakdowns within their fundamental operational loop. The core hypothesis of GPA is that agent failure modes rarely exist in isolation; rather, they manifest within and at the intersections of these three critical elements: Goal, Plan, and Action. This relationship is visualized in Figure \ref{fig:gpa_evals_reorg}.

Because agent behaviors are intersectional yet often orthogonal, measuring a single metric is insufficient. For example, an agent may achieve the user's desired goal perfectly (high Goal Fulfillment) by taking a roundabout path (low Execution Efficiency), or strictly follow a plan (high Plan Adherence) that does not actually solve the user's request (low Plan Quality). Therefore, we derive the following metrics from the intersections of the GPA framework:

\vskip 1ex\noindent \textbf{1. Goal Fulfillment (GF):} This metric evaluates whether the agent's completed \textbf{action} ultimately satisfies the user's \textbf{goal}. 

\vskip 1ex\noindent \textbf{2. Plan Quality (PQ):} This metric extracts the \textbf{plan} from the trace and assesses its optimality in achieving the given \textbf{goal}, ensuring the agent is equipped with the ideal roadmap. An optimal plan decomposes the goal into the minimal set of actionable subtasks, selects the most appropriate tool from all available tools for each step, and balances the level of detail. If replanning occurs, this judge also evaluates whether the new plan sufficiently addresses the trigger for change. 

\vskip 1ex\noindent \textbf{2A. Tool Selection (TS):} This metric complements Plan Quality and enriches the \textbf{plan} evaluation by focusing on whether the most appropriate tool was selected for each subtask. Even if the overall plan structure seems sound, Tool Selection specifically focuses on the alignment between each task requirement and each tool capability described to the planner. This includes honoring explicit system instructions on tool use, avoiding irrelevant or less capable tools, and knowing when no tool is needed for a step. 

\vskip 1ex\noindent \textbf{3. Plan Adherence (PA):} This metric evaluates whether the agent's \textbf{action} follows its stated \textbf{plan}. Independent of plan quality, plan adherence checks the agent's execution trace strictly corresponds to each planned (or replanned) step. Assuming a high-quality plan, full plan adherence would indicate the optimal steering of the agent towards the final answer.

\vskip 1ex\noindent \textbf{3A. Tool Calling (TC):} This metric complements Plan Adherence and enhances the \textbf{action} evaluation by examining how well each individual tool call was made. Even if the current tool-calling step follows the plan, Tool Calling considers whether generated tool parameters are syntactically and semantically valid, whether tool preconditions are met, and whether outputs are faithfully interpreted in order to isolate issues that arise when the agent attempts to operationalize its plan via external systems.

\textit{Note:} Our tool-related evaluations focus only on agent-controlled behavior, manifested as tool selection and tool calling. In production deployments, teams will often develop enterprise-specific tool quality evaluations, which we consider outside of the agent's control. Two examples of such measures are search relevance of retrieval models and throughput of a batch processing API tool. 

\vskip 1ex\noindent \textbf{4. Logical Consistency (LC)}: This metric sits at the intersection of \textbf{goal}, \textbf{plan}, and \textbf{action}. Logical Consistency verifies that each step in the agent's trajectory is grounded in prior context and reasoning. Logical consistency also checks for adherence to each agent's system instructions, acknowledgment and recovery from errors, and completion of all self-generated to-do tasks. 

\vskip 1ex\noindent \textbf{5. Execution Efficiency (EE):} This metric also sits at the intersection of \textbf{goal}, \textbf{plan}, and \textbf{action} and assesses the global optimality of the agent's actions towards the final state, regardless of any specific plan. It analyzes the entire execution trace for redundancies, superfluous tool calls, or unnecessary resource usage. This metric is particularly useful for evaluating agents that do not generate an explicit plan, instead focusing on the directness of the path from goal to action.

\vskip 1ex\noindent Because these failure modes are intersectional and subtle, a single "all-powerful" LLM judge often lacks the resolution to diagnose the root cause. Instead of relying on a monolithic evaluator, we propose a factorized approach: a suite of specialized LLM judges, each targeted at specific components or their intersections.

\section{Datasets}
We benchmarked our judges across three diverse datasets spanning general reasoning by multiple agents (TRAIL/GAIA), software engineering (TRAIL/SWE-bench) by a single coding agent, and enterprise data analytics by a single data agent (Snowflake Intelligence).

\subsection{TRAIL/GAIA} 
\href{https://huggingface.co/datasets/PatronusAI/TRAIL}{The TRAIL dataset} \cite{deshpande2025trail} provides 148 expert-annotated agent traces in the structured OpenTelemetry format, sourced from two distinct benchmarks: \href{https://huggingface.co/datasets/gaia-benchmark/GAIA}{GAIA} \cite{mialon2024gaia} and \href{https://huggingface.co/datasets/SWE-bench/SWE-bench}{SWE-bench} \cite{jimenez2023swe}. The GAIA benchmark is designed to test multiple agents on challenging, real-world questions that demand robust reasoning, multi-modality, web browsing, and general tool proficiency. Each of the 117 TRAIL/GAIA traces were generated with \href{https://huggingface.co/blog/open-deep-research}{Hugging Face's Open Deep-Research Agent} \cite{open_deep_research_hf}, which consists of a high-level Manager Agent capable of fact-finding, planning, and delegating tasks to a Search Agent. The Search Agent is also capable of fact-finding, planning, and has access to various tools, including web search, visiting and navigating web pages, searching for strings, inspecting files, and visiting archived URLs. 

We split the TRAIL/GAIA traces into a 50/50 dev/test split with a fixed seed. Of the 58 traces in the dev set, there are a total of 289 TRAIL-annotated errors, where 63 are low-impact, 85 are medium-impact, and 141 are high-impact errors. Of the 59 traces in the test set, there are a total of 281 TRAIL-annotated errors, where 57 are low-impact, 95 are medium-impact, and 129 are high-impact. 

\subsection{TRAIL/SWE-bench}
The SWE-bench benchmark focuses on software engineering tasks, where a single agent is given a GitHub code base and an issue and must generate a code patch to resolve it. Each of the 31 TRAIL/SWE-bench traces were generated with \href{https://github.com/xingyaoww/code-act}{Code-Act} \cite{wang2024executable}, which has access to a sandboxed environment, a python interpreter, and the \texttt{gitingest} library.

We split the TRAIL/SWE-bench traces into a 50/50 dev/test split with a fixed seed. Of the 15 traces in the dev set, there are a total of 113 TRAIL-annotated errors, where 21 are low-impact, 76 are medium-impact, and 16 are high-impact. Of the 16 traces in the test set, there are a total of 126 TRAIL-annotated errors, where 29 are low-impact, 87 are medium-impact, and 10 are high-impact.

\subsection{Snowflake Intelligence}
Snowflake Intelligence is a production-grade data agent equipped with a text-to-SQL tool and a composite retrieval search tool. We evaluated it on an internal dataset of 17 agent traces generated from data science queries requiring complex reasoning and multi-step tool usage. Unlike TRAIL/GAIA, which targets general-purpose agents and TRAIL/SWE-bench, which targets a Python coding agent, this dataset focuses specifically on agent failures in a data analysis workflow.

\section{Methodology} \label{gaia-methodology}
\subsection{Data Pre-Processing} As noted in the original TRAIL paper \cite{deshpande2025trail}, many of the raw OpenTelemetry traces exceeded the input context window length of the LLM judges. To overcome this limitation, we preprocessed each of the traces by traversing each span in the trace and extracting each of the system instructions and new messages associated with each Manager agent or Search agent, while stripping out duplicated messages in the conversation history. 

\subsection{Mapping Errors to GPA Dimensions} Two human annotators independently reviewed all TRAIL/GAIA errors in both the dev and test sets and assigned each error to one or more GPA dimensions based on the root cause of the operational failure. A third annotator cross-checked and verified the mappings.

\subsection{Designing LLM Judge Architecture}
\label{llm-judge-architecture} To ensure the GPA LLM judges can generalize across diverse agent architectures and domains, we decomposed their prompts into two distinct components: 

\textbf{1. Generic Component (Domain-Invariant):} Manually crafted definitions of the GPA metrics that remain constant regardless of the agent's domain. By fixing these definitions, we can maintain consistent core evaluation standards across different datasets. 

\textbf{2. Custom Component (Domain-Specific):} Custom instructions to adapt the judge to the specific agent system being evaluated. This includes (i) a high-level description of the agent architecture, (ii) 1-2 few-shot examples drawn from the development (dev) dataset as labeled by human annotators, and (iii) a structured output template. While the generic component is static, this custom component can be automated via prompt optimization techniques on the desired dataset.  

Full evaluation prompts can be found in Appendix \ref{llm-prompts}. Unless otherwise specified, we use \texttt{claude-4-sonnet} \cite{claude-4-sonnet} with high reasoning effort for our experiments. 

\subsection{Selecting Dataset-Specific GPA Judges} 
While the GPA framework is grounded in the universal agent lifecycle of setting goals, formulating plans, and executing actions, the \textit{observability} of these components may vary across agent architectures. For instance, certain systems may not contain or expose high-level planning in their execution traces. Because the GPA framework is inherently modular, we accommodated these variations by deploying only the specific subset of judges that aligns with each agent's observable telemetry or architectural constraints:

\vskip 1ex\noindent \textbf{TRAIL/GAIA:} Because these traces were generated by agents with explicit planning, multi-step reflection, and diverse tool use, the full operational loop is observable; therefore, we deployed the complete suite of GPA judges.

\vskip 1ex\noindent \textbf{TRAIL/SWE-bench:} The underlying CodeAct agent relies predominantly on a single tool (a Python environment) and does not output explicit high-level planning steps into its trajectory. Since unobservable plans cannot be evaluated, we deployed Logical Consistency (LC), Execution Efficiency (EE), and Tool Calling (TC) to evaluate the visible intersections of the agent's goals and actions.

\vskip 1ex\noindent \textbf{Snowflake Intelligence:} Similarly to CodeAct, explicit planning steps were not streamed back into the observable trajectory for this agent at the time of the study. Furthermore, as the primary objective for this dataset was to assess whether generated SQL code matched user intent and evaluate efficiency, we scoped our evaluation to exclusively apply the Logical Consistency (LC) and Execution Efficiency (EE) judges.

\subsection{Validating the GPA Framework} 
To establish the core validity of the Agent GPA framework, we evaluated its capacity to systematically categorize, identify, and localize agent failures. We primarily utilized the TRAIL/GAIA dataset for this validation due to its high-quality expert annotations. To demonstrate real-world utility, we also evaluated the GPA judges on the Snowflake Intelligence agent.

\vskip 1ex\noindent \textbf{LLM Judge Error Identification and Localization (TRAIL/GAIA):} 
After running each LLM judge on each of the TRAIL/GAIA traces, three human annotators manually verified whether the LLM judge successfully (i) identified the error and (ii) localized the error by explicitly citing the appropriate span ID in the trace. To isolate the source of any performance gains, we benchmarked our GPA LLM judges against two versions of the monolithic LLM judge provided by TRAIL: the original TRAIL LLM judge (baseline without control flow) and the same TRAIL LLM judge augmented with the same high-level description of the Open Deep Research agent architecture used by our custom GPA judges (baseline with control flow). Testing against both baselines clarified whether performance improvements stemmed from our factorized GPA dimensions or simply from providing the LLM judge with better system context. To measure the trade-off between error detection and false alarms, we also analyzed the overall classification performance of each judge.

\vskip 1ex\noindent \textbf{LLM Judge Alignment with Human Judgment:} \\
\textbf{\textit{TRAIL/GAIA:}} To understand LLM-human alignment, we measured the accuracy and correlation of the GPA LLM judges scoring with human scoring. A human annotator generated scores per TRAIL/GAIA trace along each GPA dimension, with another human annotator serving as a verifier. Since our LLM judges generate scores on a 4-point scale from 0 to 3 (with strictly defined min/max criteria but less delineated middle score criteria to enable grading scale flexibility), we measured both the accuracy and off-by-one accuracy of the GPA LLM Judges. Observing that the off-by-one accuracy lift stemmed from distinguishing between flexible middle scores, we bucketed scores into a 3-point scale: 0 (min score of 0), 1 (middle score of 1 or 2), and 2 (max score of 3) and report the accuracy based on this bucketed scoring system. (Table \ref{tab:judge-gt-alignment}). We measured the correlation based on the original 4 point score. \\
\textbf{\textit{Snowflake Intelligence:}}
Similarly, two human annotators produced scores for the Snowflake Intelligence traces on a 3-point scale (error, partially correct, fully correct), while three other human annotators served as verifiers. We assessed the alignment of the LC and EE judges by computing their 3-point accuracy, correlation, and Normalized Mean Absolute Error (NMAE) against the human-generated ground truth scores. For this evaluation, the custom instruction for both judges focused on checking if the generated SQL code matched the user's intent. 

\subsection{Evaluating Generalizability} 
While rigorous manual review of the TRAIL/GAIA and Snowflake Intelligence datasets established the core validity of the GPA framework, relying on human evaluation for every new agent domain presents a massive scalability bottleneck. To test generalizability and address the manual effort required for custom prompt tuning (\ref{llm-judge-architecture}), we implemented GEPA \cite{agrawal2025gepareflectivepromptevolution} to automatically optimize the generic GPA judge prompts. We then evaluated judge recall on the test sets of TRAIL/GAIA and TRAIL/SWE-bench, utilizing the latter to demonstrate that the framework can generalize to unseen domains without requiring labor-intensive manual retuning and validation. All prompt optimizations and metric evaluations were performed using \texttt{claude-4-sonnet} \cite{claude-4-sonnet}. To ensure scalable evaluation, we utilized a ``meta-judge" (a strongly aligned LLM judge verifier) to grade the GPA judges' outputs against TRAIL errors. 

\subsection{Improving Consistency} LLM judges are inherently stochastic: repeated evaluations with fixed prompts can yield different scores and rationales. To assess stability under this stochasticity, we repeatedly invoked the same model (\texttt{claude-4-sonnet}) with fixed GPA judge prompts.

For the test splits of TRAIL/GAIA ($n=59$) and TRAIL/SWE-bench ($n=31$), we collected scores in $[0,1]$ across 5 independent runs per trace. For the Snowflake Intelligence traces ($n=17$), we collected scores across 10 independent runs. We treated each run as an independent rater and computed (i) \emph{Krippendorff's $\alpha$} per metric to capture $\geq2$ rater agreement on the absolute scale, and (ii) per-trace \emph{standard deviation (std)} across runs to capture the magnitude of run-to-run fluctuations. Furthermore, we computed the \emph{Semantic Consistency Index (SCI)}, or the mean pairwise cosine similarity of the text-based rationales generated by the judges across independent runs.

Finally, to address run-to-run variance, we utilized OpenEvolve~\cite{sharma2025openevolve} to iteratively refine judge prompts using score standard deviation as the fitness objective. We applied this optimization to any metric exhibiting a baseline standard deviation $> 0.1$, regardless of whether the prompt was manually crafted or GEPA-optimized. 

\section{Results}
\subsection{GPA Framework Validation} 

{\bf 1. Mapping Agent Failures to GPA:} The Agent GPA framework systematically categorized the full spectrum of human-verified agent failures in the TRAIL dataset. Specifically, we found that all 570 human-annotated errors from the TRAIL/GAIA dataset and all 239 human-annotated errors from the TRAIL/SWE-bench dataset fell within the jurisdiction of at least one of our GPA LLM judges. Table \ref{tab:per-judge-error-mapping} details the distribution of mapped errors across the dev and test sets for TRAIL/GAIA.

\begin{table}[!htbp]
\caption{GPA Metric Error Mapping}
\label{tab:per-judge-error-mapping}
\begin{center}
\resizebox{\columnwidth}{!}{%
\begin{tabular}{l|cccc|cccc}
\toprule
& \multicolumn{4}{c|}{\bf Dev} & \multicolumn{4}{c}{\bf Test} \\
\bf Metric  & \bf Low & \bf Med & \bf High & \bf All & \bf Low & \bf Med & \bf High & \bf All \\
\midrule
LC & 31 & 19 & 70 & 120 & 34 & 29 & 77 & 140 \\
EE & 36 & 49 & 55 & 140 & 23 & 62 & 34 & 119 \\
PA & 3 & 17 & 41 & 61 & 2 & 11 & 52 & 65 \\
PQ & 3 & 7 & 7 & 17 & 1 & 10 & 3 & 14 \\
TS & 17 & 28 & 48 & 93 & 9 & 22 & 73 & 104 \\
TC & 23 & 36 & 70 & 129 & 22 & 53 & 53 & 128 \\
\bottomrule
\end{tabular}%
}
\end{center}
\end{table}

Analyzing the error distribution on the TRAIL/GAIA test set, we observed that errors related to LC, TC, and EE were the most prevalent failure modes, mapping to 140, 128, and 119 of the 281 total errors, respectively. In contrast, PQ failures were the least frequent with only 14 instances. This distribution is broadly consistent with the breakdown observed in the dev set, suggesting that failures in core reasoning, tool use, and efficiency are the most common challenges for current agents on these tasks. 

\begin{figure}[!htbp]
    \centering
    \includegraphics[width=0.9\linewidth]{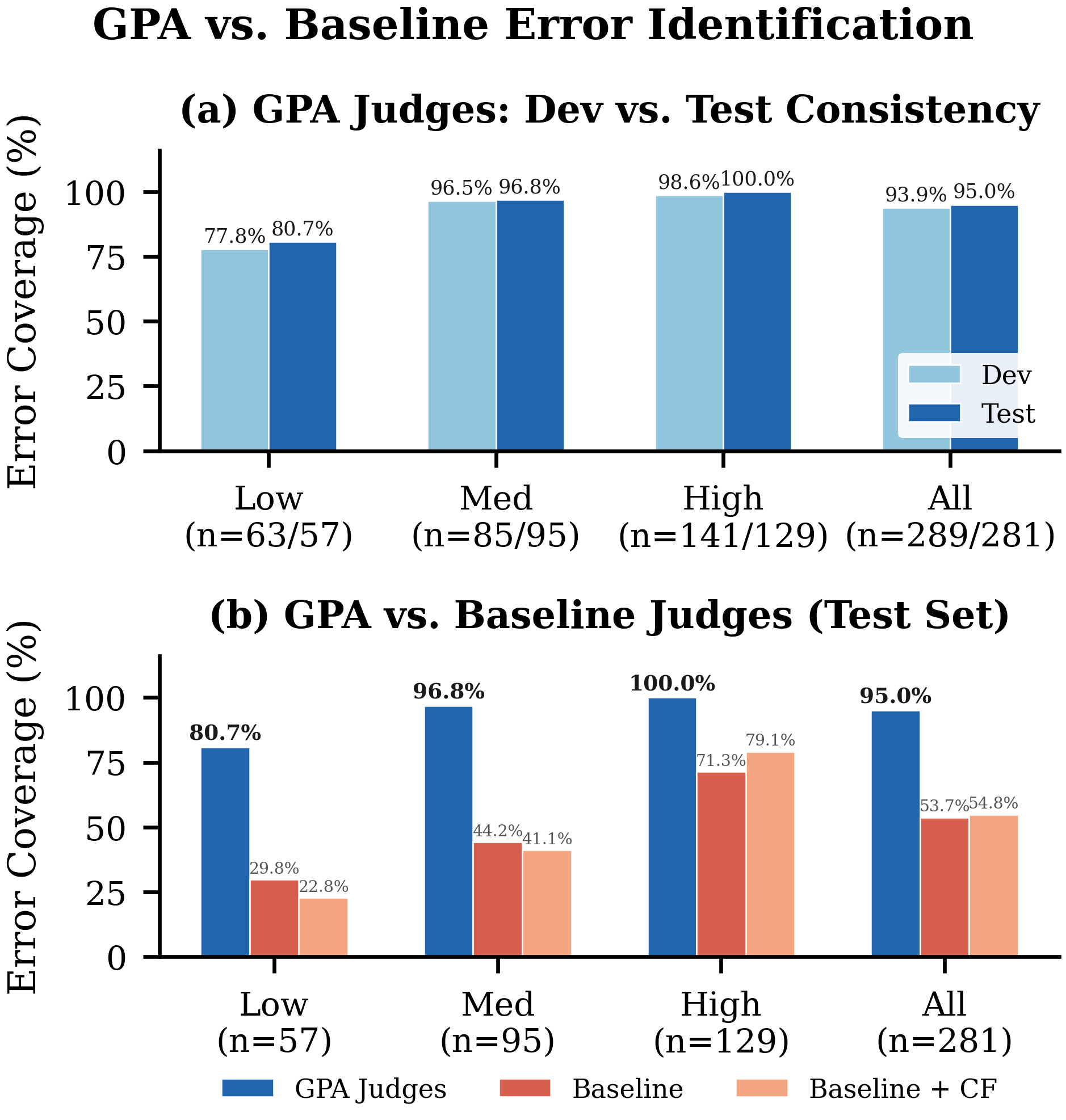}
    \caption{All GPA Judges vs Baseline Judges Error Coverage Comparison}
    \label{fig:coverage-results}
\end{figure}

\vskip 1ex\noindent {\bf 2. GPA vs Baseline Error Identification:} The GPA judge suite nearly doubled the error identification performance of the monolithic baselines. As shown in Figure \ref{fig:coverage-results}, the GPA judges captured 95\% (267/281) of the TRAIL-annotated errors on the test set, while the two baseline judges could only identify around 54\% (151-154/281).

Notably, providing the monolithic judge with context about the control flow yielded only a negligible 1.1\% improvement. This suggests performance gains derived from factorizing the evaluation dimensions among the GPA judges rather than simply providing a description about the agent's architecture alone. Furthermore, while more obvious, high-impact errors (such as data fabrication) were easier for both GPA and baseline judges to detect, identifying low and medium-impact errors proved more difficult. While baseline recall fell to 22.8-29.8\% on low-impact and 41.4-44.2\% on medium impact errors, the GPA suite scored 80.7\% recall on low-impact errors and 96.8\% on medium-impact errors. This reinforces our hypothesis that decomposing evaluation into specialized judges allows them to focus their attention and catch more subtle failures that monolithic evaluators may overlook.


\vskip 1ex\noindent \textbf{3. GPA Per-Judge Error Identification:} Individual GPA judges exhibited different operational profiles, with Tool Calling (TC) emerging as the most robust and Tool Selection (TS) operating as a high-recall specialist (Table \ref{tab:per-judge-metrics-caught-impact-all}). 

\begin{table}[!htbp]
\caption{GPA Per-Judge Caught Error Performance, All Errors}
\label{tab:per-judge-metrics-caught-impact-all}
\begin{center}
\begin{small}
\begin{sc}
\begin{tabular}{l|ccccc}
\toprule
\bf & \multicolumn{5}{c}{\bf Dev} \\ \bf Metric & \bf P & \bf R & \bf F1 & \bf F2 & \bf Acc \\
\midrule
LC & 0.636 & 0.800 & 0.709 & 0.761 & 0.727 \\
EE & 0.787 & 0.921 & 0.849 & 0.891 & 0.841 \\
PA & 0.549 & 0.918 & 0.687 & 0.809 & 0.824 \\
PQ & 0.682 & 0.882 & 0.769 & 0.833 & \textbf{0.969} \\
TS & 0.736 & \textbf{0.989} & 0.844 & 0.926 & 0.882 \\
TC & \textbf{0.858} & 0.985 & \textbf{0.917} & \textbf{0.956} & 0.920 \\
\midrule
\bf  & \multicolumn{5}{c}{\bf Test} \\
\bf Metric & \bf P & \bf R & \bf F1 & \bf F2 & \bf Acc \\
\midrule
LC & 0.763 & 0.829 & 0.795 & 0.815 & 0.787 \\
EE & 0.760 & 0.933 & 0.838 & 0.892 & 0.847 \\
PA & 0.523 & 0.892 & 0.659 & 0.782 & 0.787 \\
PQ & 0.370 & 0.714 & 0.488 & 0.602 & 0.925 \\
TS & 0.647 & \textbf{0.971} & 0.777 & 0.883 & 0.794 \\
TC & \textbf{0.879} & 0.969 & \textbf{0.922} & \textbf{0.950} & \textbf{0.925} \\
\bottomrule
\end{tabular}
\end{sc}
\end{small}
\end{center}
\vskip -0.01in
\centering\footnotesize{(P = Precision, R = Recall, F1 = F1-score, F2 = F2-score, Acc = Accuracy)}
\end{table}

TC delivered the highest and most balanced F1-score on the test set (> 0.92). Conversely, TS captured the most errors overall (recall $> 0.97$ and consistently high F2-score) but at the cost of reduced precision, making it ideal for critical applications where the cost of a missed error (a false negative) is much higher than the cost of reviewing a false alarm. Finally, the lower F1-scores for PA and PQ were caused by poor precision, though this is attributable to the smaller sample size of these specific errors in the GAIA dataset. 

Two contextual factors may also have accounted for the lower precision scores. First, during our manual review process, we observed that the TRAIL-annotated golden errors are occasionally incomplete. Because the space of all true agent failures is therefore much larger, we treated the original TRAIL human annotations as a verified, lower-bound golden set rather than an exhaustive ground truth. To ensure that we could capture these known, human-verified failures at a minimum, we optimized for recall and therefore, report the F2-score alongside the F1-score to appropriately weight this intended focus. Second, we computed precision strictly against this lower-bounded golden set. Again, through our manual review, we observed that the GPA judges would flag legitimate, previously undiscovered agent errors. Because these valid flags are penalized as false positives under the strict TRAIL baseline, the true precision of our GPA judges is likely higher than reported. 
     
\vskip 1ex\noindent \textbf{4. GPA vs Baseline Error Localization:}
The factorized GPA suite substantially outperformed monolithic baselines in error localization, successfully pinpointing the exact span ID for 85.8\% (241/281) of annotated errors on the TRAIL/GAIA test split.

\begin{figure}[!htbp]
    \centering
    \includegraphics[width=0.9\linewidth]{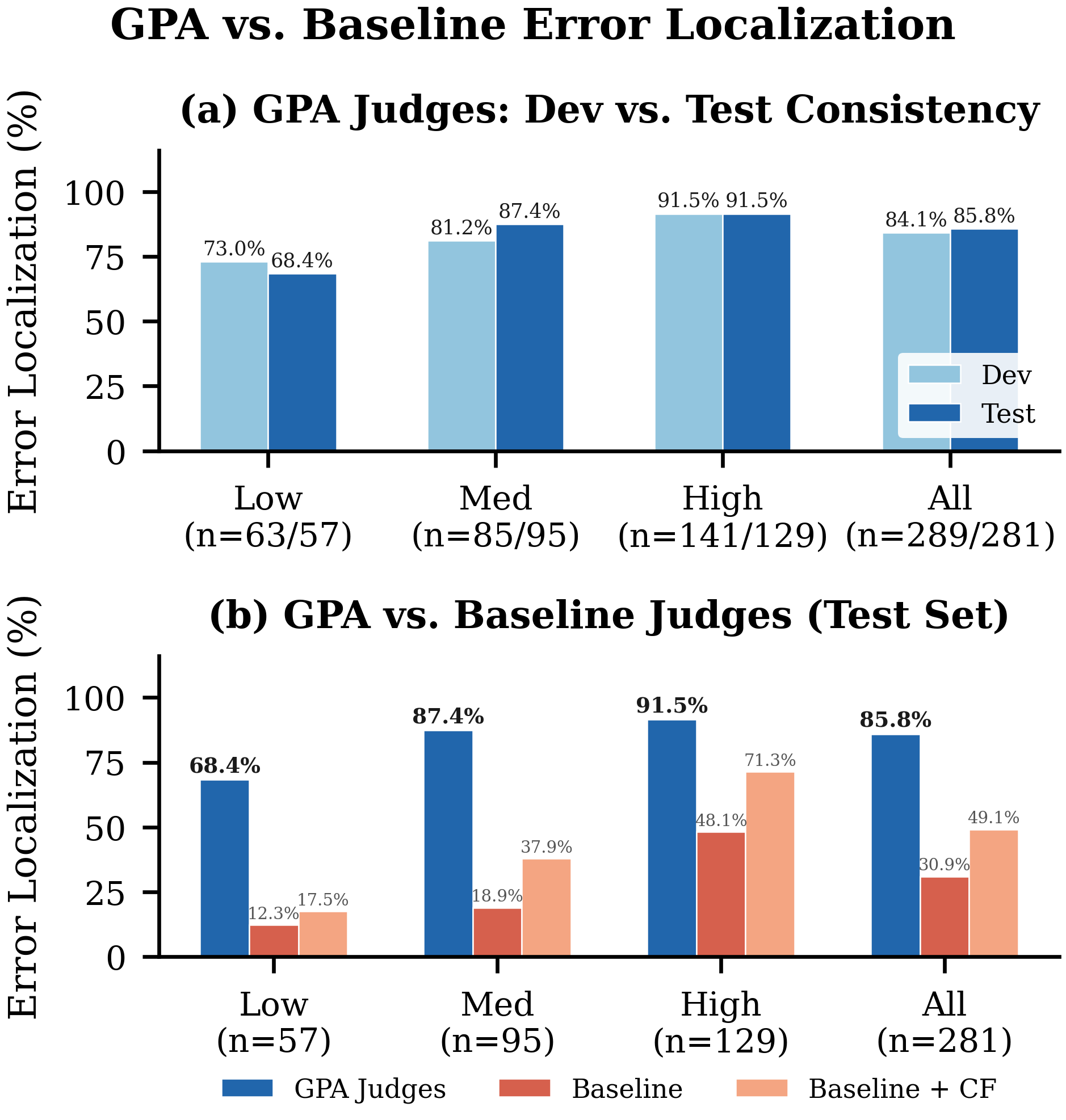}
    \caption{All GPA Judges vs Baseline Judges Error Localization Comparison}
    \label{fig:localization-results}
\end{figure}

Interestingly, while providing the baseline judge with a custom description of agent architecture barely improved error identification as noted earlier, Figure \ref{fig:localization-results} shows this additional context significantly boosted the baseline judge's ability to localize errors, with recall jumping from 31\% (87/281) to 49\% (138/281). Despite this performance gain, both monolithic judges still vastly underperformed the suite of GPA judges, which successfully localized 86\% (241/281) of the annotated errors on the TRAIL/GAIA test split. Again, these results suggest that while providing architectural context can improve a monolithic LLM judge's structural awareness of an agent trace, a factorized approach remains superior for consistently localizing span IDs for targeted debugging.

     
\vskip 1ex\noindent \textbf{5. GPA Per-Judge Error Localization:} Table \ref{tab:per-judge-metrics-localized-impact-all} shows that Execution Efficiency (EE) is the most balanced judge with the highest F1-score (0.79), while Plan Adherence (PA) and Tool Calling (TC) are high-recall and high-precision judges. 

\begin{table}[!htbp] 
\caption{GPA Per-Judge Localized Error Performance, All Errors}
\label{tab:per-judge-metrics-localized-impact-all}
\begin{center}
\begin{small}
\begin{sc}
\begin{tabular}{l|ccccc}
\toprule
& \multicolumn{5}{c}{\bf Dev} \\
\bf Metric & \bf P & \bf R & \bf F1 & \bf F2 & \bf Acc \\
\midrule
LC & 0.670 & 0.642 & 0.655 & 0.647 & 0.720 \\
EE & 0.752 & 0.714 & \textbf{0.733} & 0.722 & 0.747 \\
PA & 0.632 & \textbf{0.787} & 0.701 & \textbf{0.750} & 0.858 \\
PQ & 0.647 & 0.647 & 0.647 & 0.647 & \textbf{0.959} \\
TS & 0.750 & 0.484 & 0.588 & 0.521 & 0.782 \\
TC & \textbf{0.857} & 0.465 & 0.603 & 0.512 & 0.727 \\
\midrule
& \multicolumn{5}{c}{\bf Test} \\
\bf Metric & \bf P & \bf R & \bf F1 & \bf F2 & \bf Acc \\
\midrule
LC & 0.748 & 0.721 & 0.735 & 0.727 & 0.740 \\
EE & 0.750 & 0.832 & \textbf{0.789} & \textbf{0.814} & 0.811 \\
PA & 0.629 & \textbf{0.862} & 0.727 & 0.802 & 0.851 \\
PQ & 0.348 & 0.571 & 0.432 & 0.506 & \textbf{0.925} \\
TS & 0.779 & 0.644 & 0.705 & 0.667 & 0.801 \\
TC & \textbf{0.881} & 0.406 & 0.556 & 0.455 & 0.705 \\
\bottomrule
\end{tabular}
\end{sc}
\end{small}
\end{center}
\centering\footnotesize{(P = Precision, R = Recall, F1 = F1-score, F2 = F2-score, Acc = Accuracy)}
\end{table}

Importantly, these metrics reveal a novel framework for selecting LLM judges based on the intended application. PA acts as a ``liberal'' judge; its high recall (0.86) but low precision is suited for interactive debugging where a human reviews all potential flags. Conversely, TC is a ``conservative'' judge; its best-in-class precision (0.88) but low recall makes its sparse feedback highly trustworthy for automated processes like data filtering or reward shaping, where precision is paramount. Again, the comparatively lower scores for PA and PQ are likely a combined result of the artificially strict precision penalty and small sample size (as noted previously).


\vskip 1ex\noindent \textbf{6. GPA LLM Judge Alignment with Human Judge:} \\
\textbf{\textit{TRAIL/GAIA:}}
As shown in Table \ref{tab:judge-gt-alignment}, the judges exhibited strong overall alignment with human judgment. PA and TS showed particularly high bucketed accuracy (86.4\% and 86.8\% respectively) and strong correlation with human annotators on the test set. While the EE judge demonstrated broad error coverage (as noted in Table \ref{tab:per-judge-error-mapping}), we hypothesize that this judge showed weaker alignment with human scoring because it occasionally flags errors not strictly related to efficiency, resulting in lower generated scores compared to human scores.


\begin{table}[htbp]
\caption{GPA Judge Alignment with Human Judgment}
\label{tab:judge-gt-alignment}
\begin{center}
\resizebox{0.9\columnwidth}{!}{%
\begin{tabular}{l|ccc|ccc}
\toprule
 & \multicolumn{3}{c|}{\bf Dev} & \multicolumn{3}{c}{\bf Test} \\
\bf Metric & \bf Acc-OB1 & \bf Acc-3pt & \bf Correl & \bf Acc-OB1 & \bf Acc-3pt & \bf Correl \\
\midrule
LC & 0.983 & 0.793 & 0.626 & 0.983 & 0.881 & 0.764 \\
EE & 0.862 & 0.483 & 0.513 & 0.949 & 0.356 & 0.623 \\
PA & 1.000 & 0.862 & 0.869 & 0.983 & 0.864 & 0.917 \\
PQ & 0.879 & 0.690 & 0.565 & 0.966 & 0.695 & 0.672 \\
TS & 0.895 & 0.719 & 0.663 & 0.962 & 0.868 & 0.895 \\
TC & 0.889 & 0.667 & 0.589 & 0.941 & 0.725 & 0.706 \\
\bottomrule
\end{tabular}%
}
(Acc-OB1 = Off-by-one Accuracy, Acc-3pt = Bucketed Accuracy, Correl = Correlation)
\end{center}
\end{table}

\vskip 1ex\noindent \textbf{\textit{Snowflake Intelligence:}} Table \ref{tab:lc-we-results} shows both LC and EE's agreement with human judgment. Overall, the GPA LLM judges achieved an average 82\% agreement with humans on the 3-point scale. Importantly, the judges identified systematic error patterns that could be traced to root-cause flaws in the agent's architecture. These findings were independently validated, and the analysis enabled us to recommend several targeted improvements which were later incorporated into the agent design.

\begin{table}[!htbp]
\caption{LC and EE Alignment with Human Judgment for Snowflake Intelligence}
\label{tab:lc-we-results}
\begin{center}
\begin{small}
\begin{sc}
\begin{tabular}{l|ccc}
\toprule
\bf Metric & \bf Acc-3pt & \bf Correl & \bf NMAE \\
\midrule
LC & 0.765 & 0.795 & 0.118 \\
EE & 0.882 & 0.772 & 0.059 \\
\bottomrule
\end{tabular}
\end{sc}
\end{small}
\end{center}
\end{table}


\subsection{Generalizability}
The GPA framework scales robustly across fundamentally different agent systems. Here, we demonstrate how state-of-the-art prompt optimization systems can fully automate this process. 

\textit{Note on Meta-Judge:} When scaling evaluation, we observed that the automated meta-judge applies a stricter grading threshold than manual human review. For example, evaluating the identical ``Generic + custom'' prompts on TRAIL/GAIA yielded a 96.9\% TC coverage under human review, but only 78.6\% under the meta-judge. While we performed manual validation to ensure the meta-judge was generally well-aligned with human judgment, all subsequent performance deltas were evaluated strictly with the automated meta-judge to ensure fair relative comparisons. All meta-judge results are presented as an average of 3 runs.

\begin{figure}[!htbp]
    \centering
    \includegraphics[width=\linewidth]{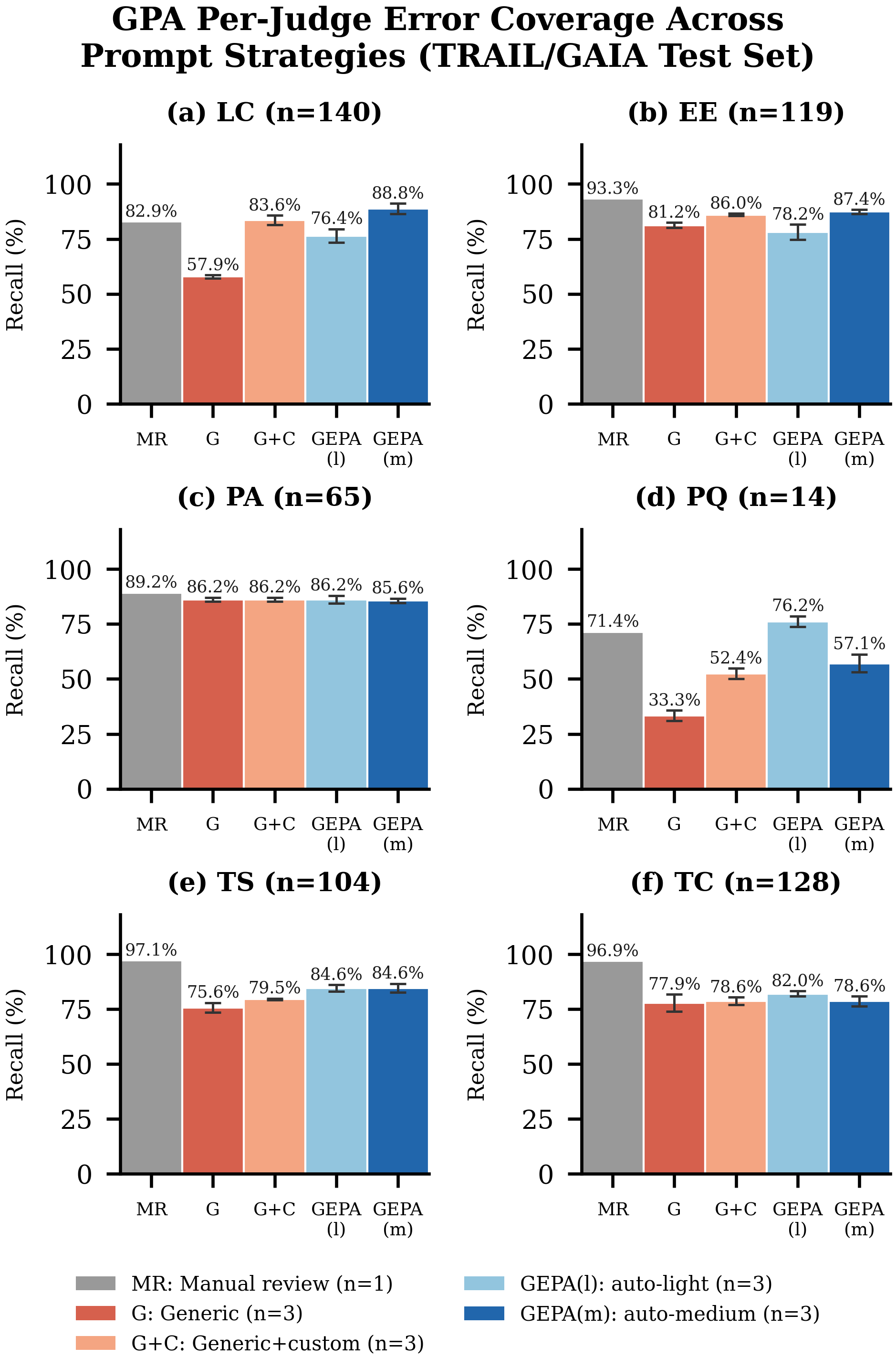}
    \caption{GPA Per-Judge Caught Error Coverage (Recall), TRAIL/GAIA Test Set}
    \label{fig:trail-gaia-gepa-test}
\end{figure}

\vskip 1ex\noindent \textbf{TRAIL/GAIA:} 
Figure \ref{fig:trail-gaia-gepa-test} demonstrates that automated prompt optimization can effectively replace manual engineering. Relying solely on the baseline Generic prompt yields poor coverage for complex reasoning tasks, capturing only 57.9\% of Logical Consistency (LC) errors and 33.3\% of Plan Quality (PQ) errors. While manually crafting and appending custom instructions improves error coverage by boosting LC to 83.6\% and PQ to 52.4\%, applying GEPA optimization (auto-light or auto-medium) effectively matches or even outperforms this approach across all six metrics. Most notably, GEPA yielded the largest improvements in the most difficult dimensions, boosting PQ to 76.2\% and LC to 88.8\%.

\begin{figure}[!htbp]
    \centering
    \includegraphics[width=\linewidth]{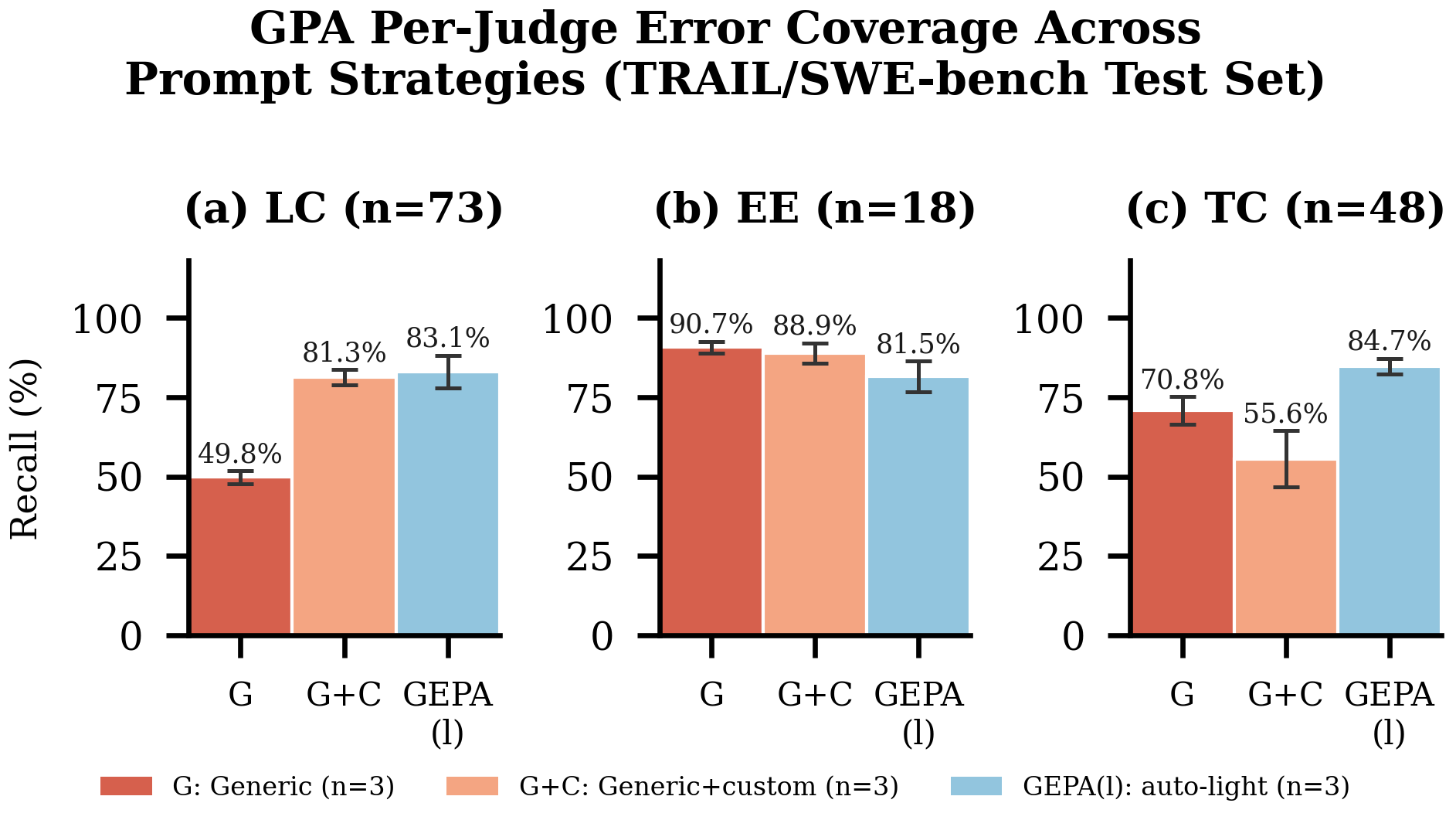}
    \caption{GPA Per-Judge Caught Error Coverage (Recall), TRAIL/SWE-bench Test Set}
    \label{fig:trail-swe-bench-gepa-test}
\end{figure}

\vskip 1ex\noindent \textbf{TRAIL/SWE-bench:} 
The GPA framework successfully generalizes to entirely unseen domains and agent architectures without any domain-specific manual retuning. Applying GEPA optimization to the generic GPA judge prompts on the SWE-bench coding traces mirrored the GEPA performance gains on TRAIL/GAIA (Figure \ref{fig:trail-swe-bench-gepa-test}). GEPA-optimized performance improved over the baseline Generic prompt, jumping from  49.8\% to 83.1\% for LC and from 70.8\% to 84.7\% for Tool Calling (TC), notably outperforming even the manually written custom prompts for these metrics. While Execution Efficiency (EE) remained generally high across all prompt types due to the CodeAct agent's straightforward operational loop, the GEPA-optimized judges demonstrate that the GPA framework can effectively transfer across domains without domain-specific manual retuning.

Extended methodology/details are available in Appendix \ref{appendix-gepa}.


\subsection{Consistency of LLM Judges}

GPA LLM judges produced consistent and reproducible scores across multiple runs and domains. 

\begin{figure}[t]
  \centering
  \begin{subfigure}[t]{0.5\linewidth} 
    \centering
    \includegraphics[width=\linewidth]{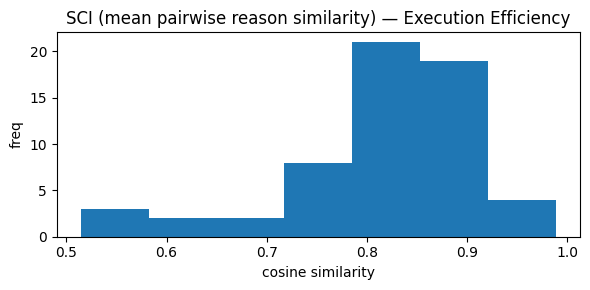}
    \subcaption{Execution Efficiency}
    \label{fig:ee-sci}
  \end{subfigure}\hfill
  \begin{subfigure}[t]{0.5\linewidth} 
    \centering
    \includegraphics[width=\linewidth]{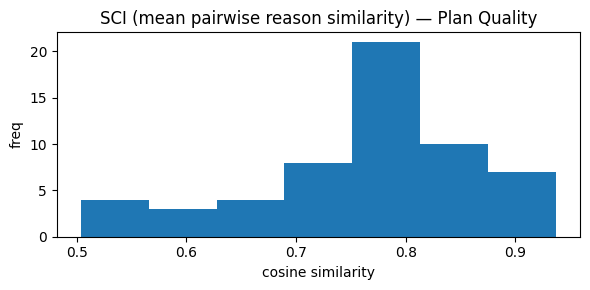}
    \subcaption{Plan Quality}
    \label{fig:pq-sci}
  \end{subfigure}
  
  \vspace{0.3cm} 

  \begin{subfigure}[t]{0.5\linewidth} 
    \centering
    \includegraphics[width=\linewidth]{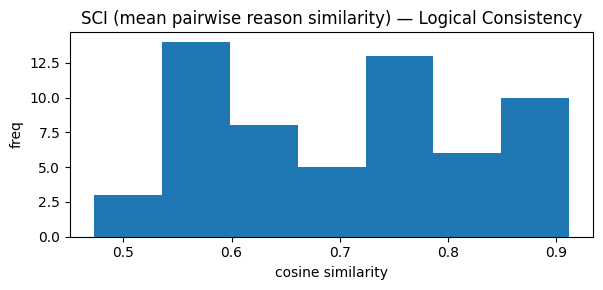}
    \subcaption{Logical Consistency}
    \label{fig:lc-sci}
  \end{subfigure}
  \caption{Semantic Consistency Index (SCI) across runs for three metrics. LLM judges show higher semantic similarity in their scoring reasons for EE than PQ and LC.}
  \label{fig:sci-3up}
\end{figure}

\vskip 1ex\noindent \textbf{TRAIL/GAIA dataset:} Table \ref{tab:claude-sonnet-reliability} shows that interrater agreement is strong: nearly all metrics achieved a Krippendorff's $\alpha > 0.70$. Execution Efficiency (EE) and Tool Selection (TS) exhibited the highest stability ($\alpha = 0.934$ and $0.907$, respectively, while maintaining lower standard deviations and tighter CIs), reflecting highly reliable scoring for concrete operational behaviors. Plan Quality and Logical Consistency were somewhat noisier (lower $\alpha$, higher variance, wider CIs), indicating greater sensitivity to sampling variation and judge interpretation. Measuring the Semantic Consistency Index (SCI) across runs (Figure \ref{fig:sci-3up}) revealed similar results: judges evaluating Execution Efficiency produced highly similar rationales (tightly clustered cosine similarities) whereas rationales for Plan Quality and Logical Consistency were less semantically consistent, aligning with their modestly higher variance.

\begin{table}[!htbp]
  \centering
  \caption{Reliability of Claude-4-Sonnet on TRAIL/GAIA across runs. 
  $\alpha$ computed treating runs as raters. 
  For per-trace variation, mean std and 95\% CI are reported.}
  \label{tab:claude-sonnet-reliability}
  \begin{tabular}{
      l | %
      S[table-format=2.0] 
      S[table-format=1.3] 
      S[table-format=1.3] 
      S[table-format=1.3] 
    }
    \toprule
    \textbf{Metric} & {$n_{\text{traces}}$} & {\boldmath$\alpha$} & \textbf{Avg std} & \textbf{95\% CI} \\
    \midrule
    LC   & 46 & 0.732 & 0.079 & 0.032 \\
    EE  & 59 & \textbf{0.934} & \textbf{0.053} & \textbf{0.021} \\
    PA        & 59 & 0.827 & 0.082 & 0.035 \\
    PQ          & 59 & 0.628 & 0.171 & 0.041 \\
    TC          & 55 & 0.878 & 0.071 & 0.026 \\
    TS        & 58 & 0.907 & 0.059 & 0.028 \\
    \midrule
    PQ (OpenEvolve) & 59 & 0.814 & 0.078 & 0.034 \\
    \bottomrule
  \end{tabular}
\end{table}

\vskip 1ex\noindent \textbf{TRAIL/SWE-bench:} As shown in Table \ref{tab:swe-bench-reliability}), Logical Consistency achieved exceptional stability ($\alpha = 0.960$), demonstrating highly reproducible judgments for reasoning-heavy evaluations. Tool Calling maintained solid agreement ($\alpha = 0.83$), and all metrics exhibited low per-trace variance (std $\leq 0.08$), indicating consistent scoring behavior across runs. These results demonstrate that our framework transfers reliably to new domains without requiring domain-specific retuning, validating the generalizability of the GPA approach.

\begin{table}[!htbp]
  \centering
  \caption{Reliability of Claude-4-Sonnet on TRAIL/SWE-bench across runs. 
  $\alpha$ computed treating runs as raters.
  For per-trace variation, mean std and 95\% CI are reported.}
  \label{tab:swe-bench-reliability}
  \begin{tabular}{
      l | %
      S[table-format=2.0] 
      S[table-format=1.4] 
      S[table-format=1.4] 
      S[table-format=1.3] 
    }
    \toprule
    \textbf{Metric} & {$n_{\text{traces}}$} & {\boldmath$\alpha$} & \textbf{Avg std} & \textbf{95\% CI} \\
    \midrule
    EE & 31 & 0.618 & 0.041 & 0.024 \\
    LC & 31 & \textbf{0.960} & \textbf{0.038} & \textbf{0.024} \\
    TC & 31 & 0.829 & 0.078 & 0.027 \\
    \bottomrule
  \end{tabular}
\end{table}

\vskip 1ex\noindent \textbf{Snowflake Intelligence:} The 10 independent runs yielded high inter-rater reliability for both evaluated metrics ($\alpha = 0.66$ for LC and $0.81$ for EE).  

\vskip 1ex\noindent We found that automated evolutionary prompt optimization significantly reduced evaluation noise for the most challenging, high-variance metrics. Applying OpenEvolve to any metric with a standard deviation exceeding $0.1$ yielded large gains in reliability by generating more granular scoring rubrics and reducing ambiguity in boundary cases. On TRAIL/GAIA, Plan Quality (manual) was the only metric to exceed this noise threshold; OpenEvolve improved Krippendorff's $\alpha$ by 30\% ($0.628 \rightarrow 0.814$), as shown in Table~\ref{tab:claude-sonnet-reliability}. On TRAIL/SWE-bench, Tool Calling remained noisy even after GEPA optimization (std $= 0.118$); applying OpenEvolve as a secondary refinement stage improved $\alpha$ by 38\% ($0.539 \rightarrow 0.743$), as shown in Table~\ref{tab:swe-bench-gepa-reliability}. These results suggest that evolutionary prompt optimization is a highly effective, systematic method for mitigating LLM stochasticity in evaluation frameworks.

\begin{table}[!htbp]
  \centering
  \caption{Reliability of GEPA-optimized judges on TRAIL/SWE-bench across runs. 
  $\alpha$ computed treating runs as raters.
  For per-trace variation, mean std and 95\% CI are reported.}
  \label{tab:swe-bench-gepa-reliability}
  \begin{tabular}{
      l | %
      S[table-format=2.0] 
      S[table-format=1.4] 
      S[table-format=1.4] 
      S[table-format=1.4] 
    }
    \toprule
    \textbf{Metric} & {$n_{\text{traces}}$} & {\boldmath$\alpha$} & \textbf{Avg std} & \textbf{95\% CI} \\
    \midrule
    EE (GEPA) & 31 & 0.804 & 0.076 & 0.028 \\
    LC (GEPA) & 31 & 0.825 & 0.080 & 0.028 \\
    TC (GEPA) & 31 & 0.539 & 0.118 & 0.025 \\
    \midrule
    TC (GEPA+OpenEvolve) & 31 & 0.743 & 0.085 & 0.031 \\
    \bottomrule  \end{tabular}
\end{table}

\section{Conclusion \& Future Work}

In conclusion, the Goal–Plan–Act (GPA) framework serves as a structured approach for evaluating LLM agents across goals, plans, and actions. By decomposing evaluation into separate metric dimensions, GPA captures diverse and subtle failure modes that monolithic or strictly outcome-based methods overlook. We show that Agent GPA is 
\begin{enumerate}
    \item \textbf{valid:} specialized GPA judges vastly outperform monolithic evaluators both in error identification and localization while maintaining strong alignment with human evaluators; 
    \item \textbf{generalizable:} applying automated prompt optimization techniques on the GPA judges successfully enables adaptation to entirely unseen domains and agent architectures while frequently outperforming manual prompt engineering; and 
    \item \textbf{consistent:} the GPA judges demonstrate high reliability across multiple runs and can be further optimized for consistency with evolutionary prompt optimization techniques.
\end{enumerate}

\vskip 1ex\noindent We see GPA as a step toward more rigorous, scalable, and interpretable agent evaluation. Future work should extend the framework to embodied agents and refine reference-free metrics for goal fulfillment and plan quality. By aligning evaluation more closely with how agents set goals, plan, and act, the GPA framework can be used as a reliable diagnostic tool and contribute to building more capable, interpretable, and trustworthy agents.

\cut{
\section*{Impact Statement}
This paper presents work whose goal is to advance the field of Machine Learning. There are many potential societal consequences of our work, none which we feel must be specifically highlighted here.
}

\section*{Reproducibility statement}
 We aim to support reproducibility by open-sourcing the Agent GPA evaluation framework, including the full code for preprocessing traces and running our LLM judges. The evaluation prompts are available in Appendix \ref{llm-prompts} of this paper. In addition, we plan to release the re-annotated and augmented TRAIL/GAIA dataset used in our experiments. Together, these resources will enable independent replication and extension of our results.

\bibliographystyle{ACM-Reference-Format}
\bibliography{acmart}

\appendix
\onecolumn

\section{Appendix}

\subsection{GPA Framework Validation: TRAIL/GAIA Coverage}

Coverage is defined as a judge's recall on the specific subset of errors it is designed to detect.

To understand the coverage of all errors in TRAIL using all judges, we can look towards the confusion matrices for the dev/test set. 

\begin{figure}[htbp]
        \centering
        \includegraphics[width=0.8\textwidth]{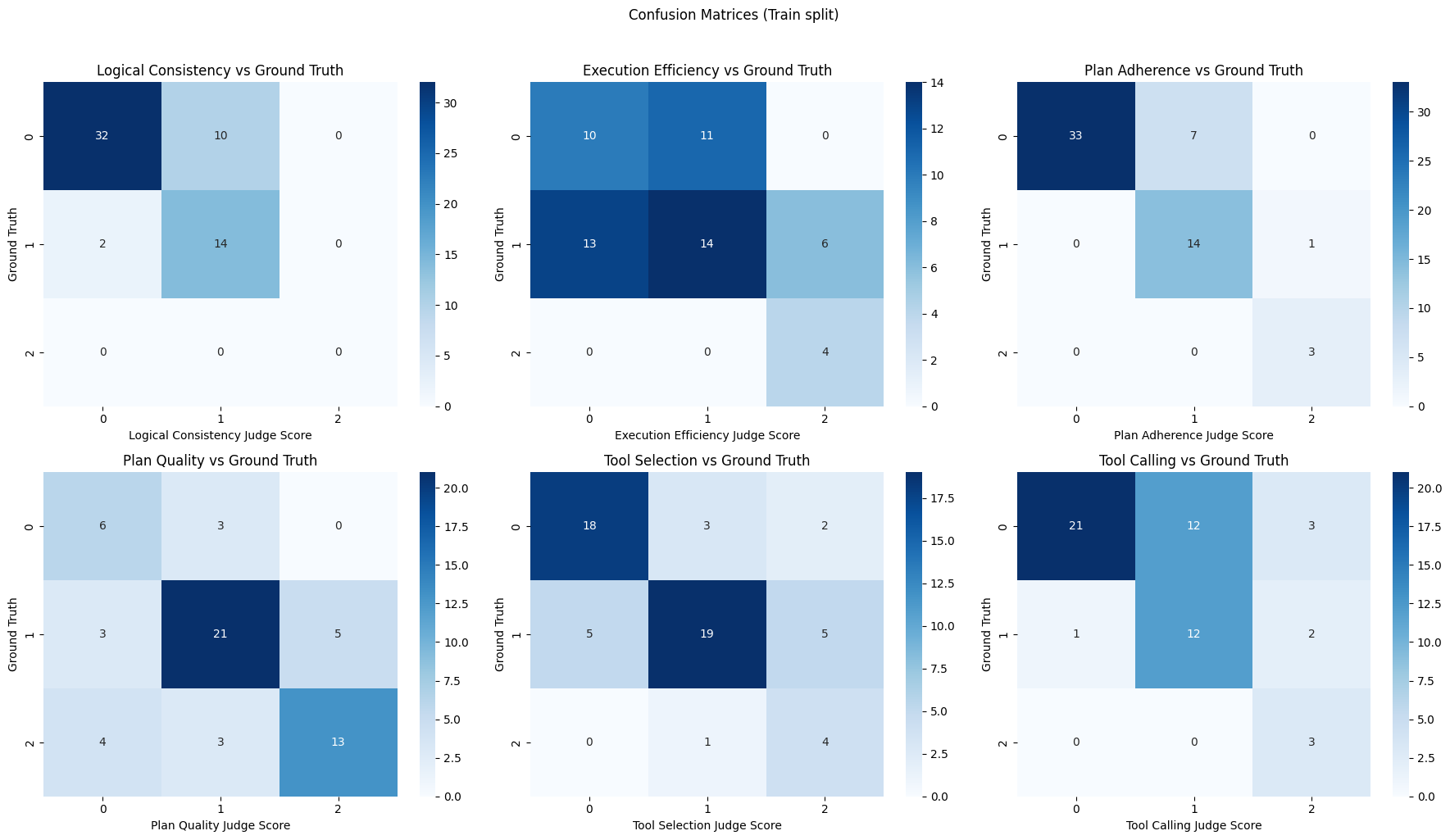}
        \caption{All GPA Judge Error Coverage Scores (0-1-2) for Dev Set (TRAIL/GAIA)}
        \label{fig:your_label_train}
\end{figure}

 \begin{figure}[htbp]
        \centering
        \includegraphics[width=0.8\textwidth]{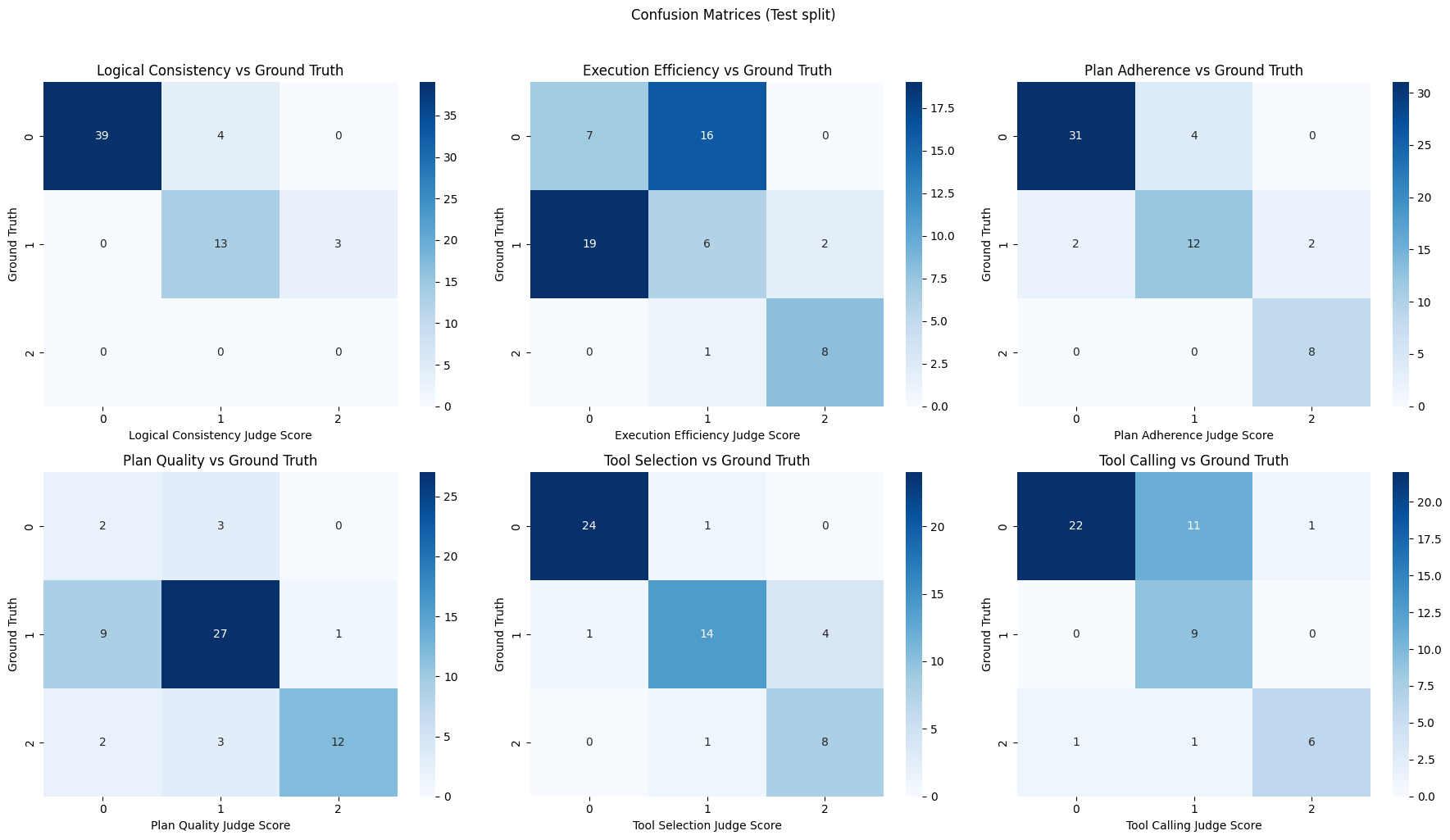}
        \caption{All GPA Judge Error Coverage Scores (0-1-2) for Test Set (TRAIL/GAIA)}
        \label{fig:your_label_test}
\end{figure}

Although the GPA judges collectively outperform the baseline, we next evaluate whether each specialist judge fulfills its intended role. To do so, we measure its coverage, defined as the recall on the specific subset of errors it was designed to detect (Table \ref{tab:per-judge-caught-coverage}). 

The TC, TS, and EE judges show high, stable coverage ($>$ 91\%), demonstrating their effectiveness. In contrast, other judges exhibit clear faults: LC consistently misses low-impact errors (coverage $<$ 60\%), while PQ's performance decreases on the test set (88\% to 71\%), suggesting it may not generalize well. This may indicate a bias in the judge towards more overt logical consistency errors, causing it to overlook subtle mistakes. The 0\% test coverage for PA and PQ on low-impact errors is based on a statistically insignificant sample size ($n \leq 2$) and thus offers no reliable evidence of their performance in this specific sub-category. This suggests the judge may have learned superficial patterns from the dev set rather than robust principles of plan adherence and quality.

\begin{table}[!htbp]
\caption{GPA Per-Judge Caught Error Coverage: Recall (TRAIL/GAIA)}
\label{tab:per-judge-caught-coverage}
\begin{center}
\resizebox{0.8\textwidth}{!}{
\begin{tabular}{l|cccc|cccc}
\toprule
\bf Metric & \multicolumn{4}{c|}{\bf Dev} & \multicolumn{4}{c}{\bf Test} \\
 & \bf Low & \bf Med & \bf High & \bf All & \bf Low & \bf Med & \bf High & \bf All \\
\midrule
LC & 54.8\% & 84.2\% & 90.0\% & 80.0\% & 58.8\% & 79.3\% & 94.8\% & 82.9\% \\
EE & 97.2\% & 85.7\% & 94.5\% & 92.1\% & 91.3\% & 91.9\% & 97.1\% & 93.3\% \\
PA & 66.7\% & 82.4\% & 97.6\% & 91.8\% & 0.0\%  & 90.9\% & 92.3\% & 89.2\% \\
PQ & 66.7\% & 100.0\%& 85.7\% & 88.2\% & 0.0\%  & 80.0\% & 66.7\% & 71.4\% \\
TS & 100.0\%& 100.0\%& 97.9\% & 98.9\% & 100.0\%& 90.9\% & 98.6\% & 97.1\% \\
TC & 95.7\% & 100.0\%& 98.6\% & 98.5\% & 100.0\%& 94.3\% & 98.1\% & 96.9\% \\
\bottomrule
\end{tabular}
}
\end{center}
\end{table}

Next, we analyze the error localization coverage of each judge. (Table \ref{tab:per-judge-localized-error-coverage}) reveals that judges targeting discrete, atomic errors, like PA and EE, excel at localizing over 83\% of errors, as specific incorrect parameters or steps are easier to pinpoint. In contrast, judges for tool-related issues, like TC (41\%) and TS (64\%), as well as more abstract reasoning like PQ (57\%) struggle. This highlights a key challenge: while these judges can detect complex plan failures, they often cannot pinpoint the precise origin, likely because localizing procedural flaws requires a causal trace of the model's reasoning chain, a notoriously difficult task for current transformer architectures \cite{lee_2025_evaluating}.

\begin{table}[!htbp]
\caption{GPA Per-Judge Localized Error Coverage: Recall (TRAIL/GAIA)}
\label{tab:per-judge-localized-error-coverage}
\begin{center}
\resizebox{0.8\textwidth}{!}{%
\begin{tabular}{l|cccc|cccc}
\toprule
\bf Metric & \multicolumn{4}{c|}{\bf Dev} & \multicolumn{4}{c}{\bf Test} \\
 & \bf Low & \bf Med & \bf High & \bf All & \bf Low & \bf Med & \bf High & \bf All \\
\midrule
LC & 48.4\% & 47.4\% & 75.7\% & 64.2\% & 47.1\% & 79.3\% & 80.5\% & 72.1\% \\
EE & 83.3\% & 67.4\% & 67.3\% & 71.4\% & 82.6\% & 82.3\% & 85.3\% & 83.2\% \\
PA & 66.7\% & 70.6\% & 82.9\% & 78.7\% & 0.0\% & 81.8\% & 90.4\% & 86.2\% \\
PQ & 0.0\% & 85.7\% & 71.4\% & 64.7\% & 0.0\% & 70.0\% & 33.3\% & 57.1\% \\
TS & 41.2\% & 39.3\% & 56.2\% & 48.4\% & 66.7\% & 50.0\% & 68.5\% & 64.4\% \\
TC & 60.9\% & 30.6\% & 50.0\% & 46.5\% & 27.3\% & 39.6\% & 47.2\% & 40.6\% \\
\bottomrule
\end{tabular}%
}
\end{center}
\end{table}

\subsection{GPA Framework Validation: Per-Judge Performance Metrics by Impact Level}

\subsubsection{Caught Errors}

Disaggregating the performance of the caught error by impact of the error (Tables \ref{tab:per-judge-metrics-caught-impact-low}, \ref{tab:per-judge-metrics-caught-impact-medium}, \ref{tab:per-judge-metrics-caught-impact-high}) reveals that the utility of a judge is not fixed, but is a dynamic function of the severity of the error. This ``contextual specialization'' demonstrates that no single judge is universally optimal.

For low-impact errors, performance is polarized. The TC judge is nearly flawless (F1=1.0). The PA and PQ judges fail, although it is worth noting that their results are based on a statistically insignificant sample size, $n \leq 2$.  As error impact increases, a clear trade-off emerges, especially for high-impact failures where specialization becomes critical:

\begin{itemize}
    \item \emph{Maximum sensitivity (Recall)}: TS is the best choice when missing an error is unacceptable, catching 99\% of critical failures.

    \item \emph{Maximum reliability (F1-score)}: TC provides the most balanced and robust performance overall.

    \item \emph{Maximum confidence (Precision)}: LC is the most precise, making its feedback the most trustworthy when a critical error is flagged.
\end{itemize}

These findings show that a single aggregate score is misleading. Effective evaluation for high-stakes applications requires a portfolio of specialized judges to be deployed based on the specific error context and the desired balance between sensitivity and precision.

\begin{table}[!htbp]
\caption{GPA Per-Judge Caught Error Performance, Low Impact Errors (TRAIL/GAIA)}
\label{tab:per-judge-metrics-caught-impact-low}
\begin{center}
\resizebox{0.8\textwidth}{!}{%
\begin{tabular}{l|ccccc|ccccc}
\toprule
\bf Metric & \multicolumn{5}{c|}{\bf Dev} & \multicolumn{5}{c}{\bf Test} \\
 & \bf P & \bf R & \bf F1 & \bf F2 & \bf Acc & \bf P & \bf R & \bf F1 & \bf F2 & \bf Acc \\
\midrule
LC & 0.548 & 0.548 & 0.548 & 0.548 & 0.556 & 0.833 & 0.588 & 0.690 & 0.625 & 0.684 \\
EE & \textbf{1.000} & 0.972 & \textbf{0.986} & 0.978 & \textbf{0.984} & 0.778 & 0.913 & 0.840 & 0.882 & 0.860 \\
PA & 0.154 & 0.667 & 0.250 & 0.400 & 0.810 & 0.000 & 0.000 & — & — & 0.825 \\
PQ & \textbf{1.000} & 0.667 & 0.800 & 0.714 & \textbf{0.984} & 0.000 & 0.000 & — & — & 0.895 \\
TS & 0.944 & \textbf{1.000} & 0.971 & \textbf{0.988} & \textbf{0.984} & 0.643 & \textbf{1.000} & 0.783 & 0.900 & 0.912 \\
TC & 0.880 & 0.957 & 0.917 & 0.940 & 0.937 & \textbf{1.000} & \textbf{1.000} & \textbf{1.000} & \textbf{1.000} & \textbf{1.000} \\
\bottomrule
\end{tabular}%
}

(P = Precision, R = Recall, F1 = F1-score, F2 = F2-score, Acc = Accuracy)
\end{center}
\end{table}

\begin{table}[!htbp]
\caption{GPA Per-Judge Caught Error Performance, Medium Impact Errors (TRAIL/GAIA)}
\label{tab:per-judge-metrics-caught-impact-medium}
\begin{center}
\resizebox{0.8\textwidth}{!}{%
\begin{tabular}{l|ccccc|ccccc}
\toprule
\bf Metric & \multicolumn{5}{c|}{\bf Dev} & \multicolumn{5}{c}{\bf Test} \\
 & \bf P & \bf R & \bf F1 & \bf F2 & \bf Acc & \bf P & \bf R & \bf F1 & \bf F2 & \bf Acc \\
\midrule
LC & 0.640 & 0.842 & 0.727 & 0.792 & 0.859 & 0.605 & 0.793 & 0.687 & 0.747 & 0.779 \\
EE & 0.875 & 0.857 & 0.866 & 0.861 & 0.847 & 0.919 & 0.919 & 0.919 & 0.919 & 0.895 \\
PA & 0.519 & 0.824 & 0.636 & 0.737 & 0.812 & 0.256 & 0.909 & 0.400 & 0.602 & 0.684 \\
PQ & 0.875 & \textbf{1.000} & 0.933 & 0.972 & \textbf{0.988} & 0.615 & 0.800 & 0.696 & 0.755 & 0.926 \\
TS & 0.800 & \textbf{1.000} & 0.889 & 0.952 & 0.918 & 0.426 & 0.909 & 0.580 & 0.741 & 0.695 \\
TC & \textbf{0.900} & \textbf{1.000} & \textbf{0.947} & \textbf{0.978} & 0.953 & \textbf{0.926} & \textbf{0.943} & \textbf{0.935} & \textbf{0.940} & \textbf{0.926} \\
\bottomrule
\end{tabular}%
}

(P = Precision, R = Recall, F1 = F1-score, F2 = F2-score, Acc = Accuracy)
\end{center}
\end{table}

\begin{table}[!htbp]
\caption{GPA Per-Judge Caught Error Performance, High Impact Errors (TRAIL/GAIA)}
\label{tab:per-judge-metrics-caught-impact-high}
\begin{center}
\resizebox{0.8\textwidth}{!}{%
\begin{tabular}{l|ccccc|ccccc}
\toprule
\bf Metric & \multicolumn{5}{c|}{\bf Dev} & \multicolumn{5}{c}{\bf Test} \\
 & \bf P & \bf R & \bf F1 & \bf F2 & \bf Acc & \bf P & \bf R & \bf F1 & \bf F2 & \bf Acc \\
\midrule
LC & 0.663 & 0.900 & 0.764 & 0.840 & 0.723 & \textbf{0.811} & 0.948 & 0.874 & 0.917 & 0.837 \\
EE & 0.642 & 0.945 & 0.765 & 0.864 & 0.773 & 0.579 & 0.971 & 0.725 & 0.855 & 0.806 \\
PA & 0.645 & 0.976 & 0.777 & 0.885 & 0.837 & 0.750 & 0.923 & 0.828 & 0.882 & 0.845 \\
PQ & 0.500 & 0.857 & 0.632 & 0.750 & \textbf{0.950} & 0.222 & 0.667 & 0.333 & 0.476 & \textbf{0.938} \\
TS & 0.653 & 0.979 & 0.783 & 0.890 & 0.816 & 0.758 & \textbf{0.986} & 0.857 & 0.930 & 0.814 \\
TC & \textbf{0.831} & \textbf{0.986} & \textbf{0.902} & \textbf{0.950} & 0.894 & 0.800 & 0.981 & \textbf{0.881} & \textbf{0.939} & 0.892 \\
\bottomrule
\end{tabular}%
}

(P = Precision, R = Recall, F1 = F1-score, F2 = F2-score, Acc = Accuracy)
\end{center}
\end{table}

\subsubsection{Localized Errors}

Our analysis of error localization performance (Tables \ref{tab:per-judge-metrics-localized-impact-low}, \ref{tab:per-judge-metrics-localized-impact-medium}, \ref{tab:per-judge-metrics-localized-impact-high}) reveals a dramatic contextual specialization, where a judge's utility is not fixed but dynamically shifts with error severity, leading to surprising performance inversions and role-reversals.

This is most evident with the PA judge, which fails completely on low-impact errors but becomes the top-performing localizer for high-impact failures (F1=0.85). This suggests critical failures are often linked to the concrete adherence errors PA is designed to catch. In contrast, the TC judge solidifies its role as a ``conservative but accurate'' specialist, consistently delivering perfect precision but with low recall, making its feedback sparse but highly trustworthy.

Furthermore, the TS judge undergoes a critical role-reversal. While a high-recall agent for general error detection, it transforms into the highest-precision localizer for high-impact errors (P=0.85), making it the most reliable choice for pinpointing the exact source of a critical failure. These findings demonstrate that effective automated debugging requires a dynamic ensemble of judges, selected based on the specific context of a failure, as no single judge is reliable across all conditions.

\begin{table}[!htbp]
\caption{GPA Per-Judge Localized Error Performance, Low Impact Errors (TRAIL/GAIA)}
\label{tab:per-judge-metrics-localized-impact-low}
\begin{center}
\resizebox{0.8\textwidth}{!}{%
\begin{tabular}{l|ccccc|ccccc}
\toprule
\bf Metric & \multicolumn{5}{c|}{\bf Dev} & \multicolumn{5}{c}{\bf Test} \\
 & \bf P & \bf R & \bf F1 & \bf F2 & \bf Acc & \bf P & \bf R & \bf F1 & \bf F2 & \bf Acc \\
\midrule
LC & 0.682 & 0.484 & 0.566 & 0.514 & 0.635 & 0.800 & 0.471 & 0.593 & 0.513 & 0.614 \\
EE & \textbf{1.000} & \textbf{0.833} & \textbf{0.909} & \textbf{0.862} & 0.905 & 0.760 & \textbf{0.826} & 0.792 & \textbf{0.812} & 0.825 \\
PA & 0.400 & 0.667 & 0.500 & 0.588 & 0.937 & 0.000 & 0.000 & — & — & 0.895 \\
PQ & — & 0.000 & — & — & \textbf{0.952} & 0.000 & 0.000 & — & — & 0.895 \\
TS & \textbf{1.000} & 0.412 & 0.583 & 0.467 & 0.841 & \textbf{1.000} & 0.667 & \textbf{0.800} & 0.714 & \textbf{0.947} \\
TC & 0.933 & 0.609 & 0.737 & 0.654 & 0.841 & \textbf{1.000} & 0.273 & 0.429 & 0.319 & 0.719 \\
\bottomrule
\end{tabular}%
}

(P = Precision, R = Recall, F1 = F1-score, F2 = F2-score, Acc = Accuracy)
\end{center}
\end{table}

\begin{table}[!htbp]
\caption{GPA Per-Judge Localized Error Performance, Medium Impact Errors (TRAIL/GAIA)}
\label{tab:per-judge-metrics-localized-impact-medium}
\begin{center}
\resizebox{0.8\textwidth}{!}{%
\begin{tabular}{l|ccccc|ccccc}
\toprule
\bf Metric & \multicolumn{5}{c|}{\bf Dev} & \multicolumn{5}{c}{\bf Test} \\
 & \bf P & \bf R & \bf F1 & \bf F2 & \bf Acc & \bf P & \bf R & \bf F1 & \bf F2 & \bf Acc \\
\midrule
LC & 0.643 & 0.474 & 0.546 & 0.500 & 0.824 & 0.622 & 0.793 & 0.697 & 0.752 & 0.790 \\
EE & 0.868 & 0.674 & 0.759 & 0.705 & 0.753 & 0.911 & \textbf{0.823} & \textbf{0.864} & \textbf{0.839} & 0.832 \\
PA & 0.522 & 0.706 & 0.600 & 0.659 & 0.812 & 0.346 & 0.818 & 0.487 & 0.643 & 0.800 \\
PQ & 0.857 & \textbf{0.857} & \textbf{0.857} & \textbf{0.857} & \textbf{0.977} & 0.636 & 0.700 & 0.667 & 0.686 & \textbf{0.926} \\
TS & 0.846 & 0.393 & 0.537 & 0.440 & 0.777 & 0.524 & 0.500 & 0.512 & 0.505 & 0.779 \\
TC & \textbf{1.000} & 0.306 & 0.468 & 0.355 & 0.706 & \textbf{1.000} & 0.396 & 0.568 & 0.451 & 0.663 \\
\bottomrule
\end{tabular}%
}

(P = Precision, R = Recall, F1 = F1-score, F2 = F2-score, Acc = Accuracy)
\end{center}
\end{table}

\begin{table}[!htbp]
\caption{GPA Per-Judge Localized Error Performance, High Impact Errors (TRAIL/GAIA)}
\label{tab:per-judge-metrics-localized-impact-high}
\begin{center}
\resizebox{0.8\textwidth}{!}{%
\begin{tabular}{l|ccccc|ccccc}
\toprule
\bf Metric & \multicolumn{5}{c|}{\bf Dev} & \multicolumn{5}{c}{\bf Test} \\
 & \bf P & \bf R & \bf F1 & \bf F2 & \bf Acc & \bf P & \bf R & \bf F1 & \bf F2 & \bf Acc \\
\midrule
LC & 0.671 & 0.757 & 0.711 & 0.738 & 0.695 & 0.795 & 0.805 & 0.800 & 0.803 & 0.760 \\
EE & 0.569 & 0.673 & 0.617 & 0.649 & 0.674 & 0.569 & 0.853 & 0.682 & 0.775 & 0.791 \\
PA & 0.708 & \textbf{0.829} & \textbf{0.764} & \textbf{0.802} & 0.851 & 0.797 & \textbf{0.904} & \textbf{0.847} & \textbf{0.880} & 0.868 \\
PQ & 0.500 & 0.714 & 0.588 & 0.658 & \textbf{0.950} & 0.143 & 0.333 & 0.200 & 0.263 & \textbf{0.938} \\
TS & 0.675 & 0.563 & 0.614 & 0.582 & 0.759 & \textbf{0.848} & 0.685 & 0.758 & 0.712 & 0.752 \\
TC & \textbf{0.796} & 0.500 & 0.614 & 0.540 & 0.688 & 0.781 & 0.472 & 0.588 & 0.512 & 0.729 \\
\bottomrule
\end{tabular}%
}

(P = Precision, R = Recall, F1 = F1-score, F2 = F2-score, Acc = Accuracy)
\end{center}
\end{table}



\section{GPA Framework Validation: Extended Judge Agreement Statistics (TRAIL/GAIA)} \label{appendix-agreement}
We report human-LLM annotation agreement metrics in Table \ref{tab:judge-gt-alignment} of our paper. To calculate these metrics, we had 3 human annotators review the 117 traces across both dev and test sets and grade each trace along the 6 GPA judge dimensions. Then, we calculated the accuracy and correlation of each GPA judge with human judgment, where we found that our LLM judges generally exhibited strong agreement with our human annotations across the board.

In addition to existing judge alignment with human scoring, we report LLM-human agreement with Krippendorff's $\alpha$, as well as per-metric pairwise Cohen's $\kappa$ agreement between human annotators and LLM. 
The descriptive mean Krippendorff's $\alpha$ across GPA judge types is 0.7346 and 0.6718 on TRAIL/GAIA dev and test sets, respectively. We find the global Krippendorff's $\alpha$ also being supportive of tentative conclusive with 0.7690 and 0.7387 on the dev and test sets, respectively. 

With respect to human-human agreement, the consensus judge agreement rate is 0.7009 on the dev set and 0.6674 on the test set.

\begin{table}[h]
\centering
\caption{Cohen's $\kappa$ per GPA metric (Human vs. LLM), TRAIL/GAIA Dev Set}
\begin{small}
\begin{sc}
\begin{tabular}{lr}
\toprule
Metric & Cohen's $\kappa$ \\
\midrule
LC & 0.6410 \\
EE & 0.7272 \\
PA & 0.8221 \\
PQ & 0.7058 \\
TS & 0.8594 \\
TC & 0.6629 \\
\bottomrule
\end{tabular}
\end{sc}
\end{small}
\end{table}

\begin{table}[h!]
\centering
\caption{Cohen's $\kappa$ per GPA metric (Human vs. LLM), TRAIL/GAIA Test Set}
\begin{small}
\begin{sc}
\begin{tabular}{lr}
\toprule
Metric & Cohen's $\kappa$ \\
\midrule
LC & 0.5161 \\
EE & 0.7626 \\
PA & 0.7681 \\
PQ & 0.4952 \\
TS & 0.8584 \\
TC & 0.6658 \\
\bottomrule
\end{tabular}
\end{sc}
\end{small}
\end{table}

\section{GPA Framework Validation: Cross-GPA Metrics Agreement and Orthogonality Analysis (TRAIL/GAIA)}
We report Krippendorff's $\alpha$ and Cohen's $\kappa$ as measures of agreement and Jaccard similarity and phi correlation as measures of evaluation overlap and binary co-occurrence, respectively. For all analyses, we convert the scores of each LLM-based metric to binary labels (0/1). We do this because (1) severity levels on a Likert scale are difficult to compare meaningfully across different metrics, and (2) for our purposes, the key signal is whether a metric identifies a failure on a trace at all, rather than how severe that failure is. This binarization makes the agreement statistics more interpretable and better aligned with our goal of assessing the capability of the metrics' capability to detect failures.

The consistently low agreement and low correlations across all four measures demonstrate that the six metrics identify non-overlapping, complementary, and semantically different types of errors. This strengthens our motivation to evaluate agents along multiple dimensions rather than collapsing behavior into a single rating. No single metric captures the full spectrum of agent failures, and the interplay of these metrics offers a richer and more diagnostic understanding of model behavior.

\begin{table}[h!]
\centering
\caption{Krippendorff's $\alpha$ (binary)}
\begin{small}
\begin{sc}
\begin{tabular}{lrrrrrr}
\toprule
   & LC & EE & PA & PQ & TS & TC \\
\midrule
LC & 1      & 0.12913  & 0.121952 & -0.238975 & 0.211812 & 0.250384 \\
EE & 0.12913 & 1        & -0.102802 & -0.320435 & 0.036614 & 0.468612 \\
PA & 0.121952 & -0.102802 & 1        & -0.054528 & 0.258678 & 0.002886 \\
PQ & -0.238975 & -0.320435 & -0.054528 & 1        & -0.051312 & -0.293001 \\
TS & 0.211812 & 0.036614  & 0.258678 & -0.051312 & 1        & 0.029749 \\
TC & 0.250384 & 0.468612  & 0.002886 & -0.293001 & 0.029749 & 1 \\
\bottomrule
\end{tabular}
\end{sc}
\end{small}
\end{table}

\begin{table}[h!]
\centering
\caption{Cohen's $\kappa$ (binary)}
\begin{small}
\begin{sc}
\begin{tabular}{lrrrrrr}
\toprule
   & LC & EE & PA & PQ & TS & TC \\
\midrule
LC & 1      & 0.129309 & 0.143774 & -0.013736 & 0.216388 & 0.249211 \\
EE & 0.129309 & 1        & -0.059086 & -0.049211 & 0.050969 & 0.469229 \\
PA & 0.143774 & -0.059086 & 1        & 0.046656  & 0.262031 & 0.024679 \\
PQ & -0.013736 & -0.049211 & 0.046656 & 1        & 0.091104 & -0.064938 \\
TS & 0.216388 & 0.050969 & 0.262031 & 0.091104 & 1        & 0.033826 \\
TC & 0.249211 & 0.469229 & 0.024679 & -0.064938 & 0.033826 & 1 \\
\bottomrule
\end{tabular}
\end{sc}
\end{small}
\end{table}

\begin{table}[h!]
\centering
\caption{Jaccard similarity across error activations}
\begin{small}
\begin{sc}
\begin{tabular}{lrrrrrr}
\toprule
   & LC & EE & PA & PQ & TS & TC \\
\midrule
LC & 1      & 0.412556 & 0.324607 & 0.06135  & 0.401015 & 0.451456 \\
EE & 0.412556 & 1        & 0.237209 & 0.044944 & 0.331797 & 0.591837 \\
PA & 0.324607 & 0.237209 & 1        & 0.087719 & 0.36747  & 0.262626 \\
PQ & 0.06135  & 0.044944 & 0.087719 & 1        & 0.113636 & 0.036585 \\
TS & 0.401015 & 0.331797 & 0.36747  & 0.113636 & 1        & 0.30622 \\
TC & 0.451456 & 0.591837 & 0.262626 & 0.036585 & 0.30622  & 1 \\
\bottomrule
\end{tabular}
\end{sc}
\end{small}
\end{table}

\begin{table}[h!]
\centering
\caption{Phi correlation (binary co-occurrence)}
\begin{small}
\begin{sc}
\begin{tabular}{lrrrrrr}
\toprule
   & LC & EE & PA & PQ & TS & TC \\
\midrule
LC & 1      & 0.129794 & 0.15204  & -0.026461 & 0.219687 & 0.24926 \\
EE & 0.129794 & 1        & -0.06448 & -0.102185 & 0.052715 & 0.471896 \\
PA & 0.15204  & -0.06448 & 1        & 0.068626  & 0.265523 & 0.025934 \\
PQ & -0.026461 & -0.102185 & 0.068626 & 1        & 0.151887 & -0.122992 \\
TS & 0.219687 & 0.052715 & 0.265523 & 0.151887 & 1        & 0.03423  \\
TC & 0.24926  & 0.471896 & 0.025934 & -0.122992 & 0.03423  & 1 \\
\bottomrule
\end{tabular}
\end{sc}
\end{small}
\end{table}

\begin{enumerate}
    \item \textbf{Metrics capture distinct failure modes.} Agreement is consistently low across $\alpha$, $\kappa$, phi, and Jaccard. The six metrics fire on different phenomena, supporting multi-dimensional evaluation.
    \item \textbf{PQ (Plan Quality) is the most independent metric.} PQ shows near-zero or negative agreement with most metrics and the lowest Jaccard overlaps, reflecting a unique axis of planning quality.
    \item \textbf{EE (Execution Efficiency) and TC (Tool Calling) are closely related.} EE--TC is the strongest pair across all measures, suggesting execution failures tend to co-occur with tool-calling issues.
    \item \textbf{TS (Tool Selection) shows mild associations to LC, EE, and PA.} TS correlates weakly but consistently with reasoning-related metrics, while still behaving as a distinct dimension.
    \item \textbf{LC (Logical Consistency) differs strongly from PQ and PA.} LC has weak or negative relationships with planning metrics, indicating it captures a distinct form of reasoning failure.
    \item \textbf{PA (Plan Adherence) has only localized relationships.} PA aligns moderately with TS but weakly with other metrics, reflecting procedural rather than conceptual failure.
    \item \textbf{Jaccard values confirm sparse co-activation.} Most Jaccard scores fall between 0.04 and 0.40, demonstrating that metrics rarely trigger on the same traces.
    \item \textbf{Phi correlations reinforce weak interdependence.} Phi largely mirrors $\kappa$ and shows weak associations, further confirming metric independence.
\end{enumerate}

\section{LLM Judge Prompts} \label{llm-prompts}

\subsection{Custom Instruction: Control Flow of Open Deep-Research}
\begin{lstlisting}
Agent Architecture and Trace Structure: The agent architecture consists of a primary manager Agent (also referred to as CodeAgent) that delegates tasks to a search_agent (also referred to as ToolCallingAgent).

Overall Flow:
Every trace consists of several spans (with span_id numbers and parent span_id numbers). Each trace begins with the manager (CodeAgent). The process follows a clear, hierarchical structure where the manager outlines a high-level plan and the search_agent executes the detailed, tool-based steps for each part of that plan.

Manager Agent Initiation:
The trace starts with the manager. In its initial child spans, you will observe the following sequence:

A preparatory survey is created based on the user's query.

A high-level plan is formulated from this survey.

The Manager agent begins executing Step 1 of its plan.

Manager Agent Step 1:
Within the child span for Step 1, the Manager agent decides how to proceed given the initial fact survey and plan. The Manager agent will produce a thought, which may call the search_agent to perform the necessary actions or research.

search_agent (ToolCallingAgent) Execution Loop:
Once called, the search_agent begins its own execution loop. In its child spans, you will observe the following sequence:

A preparatory survey to the specific sub-task it received from the Manager agent.

A plan tailored to the specific sub-task it received from the Manager agent.

The search_agent executes an initial set of up to four steps. Each step involves an LLM call to generate a tool-call, followed by the tool's execution. After these initial steps, search_agent synthesizes the information gathered into an updated fact list and refines its plan. The search_agent may then continue to execute more tool-steps based on this updated plan.

This loop continues until the search_agent has gathered enough information to comprehensively answer the manager's sub-task, at which point it calls final_answer.

Returning Control to the Manager agent
The final_answer from the search_agent is returned to the Manager agent, concluding the Manager agent's Step 1. The Manager agent then proceeds to Step 2 of its high-level plan, using the result from the previous step as context. This entire cycle repeats for all subsequent steps in the Manager Agent's plan.

Whenever you want to point out anything in the trace, cite the span_id number of the span that you are referring to.

\end{lstlisting}

\subsection{Logical Consistency Judge: Generic LC Criteria \& Custom Instruction}

\begin{lstlisting}
You are a meticulous and analytical LOGICAL CONSISTENCY evaluator: provide a score for the logical consistency given an agentic system's trace.

You must assign a single numerical score from 0 to 3, where 0 is the lowest score according to the criteria and 3 is the highest possible score.

Evaluation criteria:

    Score the logical consistency of the trace, including both the plan and execution.
    
    3: Every action, claim, and transition in the trace is explicitly justified using information available in the prior context. Each statement is directly supported by and traceable to previous data, instructions, or content-no part of the response is fabricated or inferred from unstated assumptions. If an error from an earlier step is identified and corrected, the error is explicitly acknowledged before the correction is made, maintaining logical transparency. Each system instruction is followed. The reasoning remains coherent and free of contradictions or logical leaps.
    
    Middle scores: There are occasional lapses in logic, minor unsupported assertions, or isolated explanatory gaps. Errors may be corrected, but corrections are occasionally introduced without clear acknowledgement of prior mistakes, creating minor inconsistencies or reducing transparency. Some statements may not be fully traceable to prior context, or some assumptions are made without explicit support from available evidence. Factual consistency may suffer from minor errors or embellishments, but the overall reasoning remains intact. Most previously assigned tasks and instructions remain intact.
    
    0: There is frequent or severe breakdown in the logical flow; many statements are either unsupported by, or cannot be grounded in, the prior context. Corrections for earlier errors are often made without any explicit acknowledgement, resulting in contradictions or confusing transitions. Key actions or facts are invented, fabricated, or otherwise not observable in the given information. Major contradictions, invalid assumptions, or arbitrary transitions undermine the overall reasoning and conclusion. Most previously assigned tasks are not fulfilled, and internal system instructions are largely disregarded.
    
    Track each agent's system instructions and conversation history, ensuring all subsequent outputs from that agent adhere to its established guidelines and prior dialogue, even when agents speak interchangeably. For the manager agent and each unique search_agent that may exist in the trace, evaluate the logical consistency for the agent's actions and responses. For each agent, ensure that each response is consistent with the system instructions and prior dialogue.

You must structure your entire response:
Manager Agent
System Instructions
[Paste all system instructions associated with the manager agent.]
Logical Consistency issues
[All Logical Consistency issues associated with the manager agent]

search_agent 0 (if exists)
System Instructions
[Paste all system instructions associated with the search_agent.]
Logical Consistency issues
[List all Logical Consistency issues associated with this search_agent]
...
search_agent n (if exists)
System Instructions
[Paste all system instructions associated with the search_agent.]
Logical Consistency issues
[List all Logical Consistency issues associated with this search_agent]

Here are some examples of logical consistency issues:
{
"evidence": "The plan output content ends with the last step of the plan instead of the <end_plan> tag.",
"description": "The plan generation step did not conclude with the required '<end_plan>' tag as specified in the instructions for plan generation."
},
{
"evidence": "Thought: I recall that Girls Who Code published a statistic showing that in the 1980s, women accounted for 37% of computer scientists (or computer science graduates) and that over a span of three decades that figure dropped by 13 percentage points (from 37% down to 24%)",
"description": "The system recalled a statistic about Girls Who Code and the percentage of women in computer science, but this information was not verified using the search tool as planned. The system states "Thought: I recall that Girls Who Code published a statistic showing that in the 1980s, women accounted for 37% of computer scientists (or computer science graduates) and that over a span of three decades that figure dropped by 13 percentage points (from 37% down to 24%). In other words, it took 30 years for that change to occur. Based on that well-circulated statistic that Girls Who Code highlighted, I will output 30 years as the final answer.""
}

Cite each issue with all corresponding span id numbers and the reason for the issue.

Be critical in your evaluation. For each step in the trace with an issue (e.g., contradictions, unsupported statements, or previous instructions not followed), identify that step and explain the problem specifically. Flag any implicit assumptions.

{TRACE}

Please evaluate using the following template:

Criteria: <Provide the criteria for this evaluation, restating the criteria you are using to evaluate>
Supporting Evidence: <Provide your reasons for scoring based on the listed criteria step by step. Tie it back to the evaluation being completed.>
Score: <The score based on the given criteria>

Please respond using the entire template above.
\end{lstlisting}

\subsection{Execution Efficiency Judge: Generic EE Criteria \& Custom Instruction}
\begin{lstlisting}
You are a meticulous and analytical EXECUTION EFFICIENCY evaluator: provide a score for how efficiently the agent executes its steps. Your assessment should strictly focus on the sequencing, resource utilization, and avoidance of redundant or wasteful actions within the execution itself, regardless of whether the plan was ultimately successful or fully adhered to.

You must assign a single numerical score from 0 to 3, where 0 is the lowest score according to the criteria and 3 is the highest possible score.

Evaluation criteria:

    Score the efficiency of the execution.

    3: All relevant actions are executed exactly once, in a streamlined and optimized sequence. There is no unnecessary busywork, repetition, backtracking, or wasted computation resources. Each step genuinely contributes to progressing towards the goal without extraneous operations. Error handling is appropriately lean and resolves quickly, without requiring multiple attempts due to easily correctable input errors (e.g. incorrect tool arguments). Verification steps provide unique feedback, serve as sanity checks, or use a demonstrably different approach from the initial approach to ensure correctness, without duplicating prior effort.

    Middle scores: Some instances of workflow inefficiency such as redundant actions, non-ideal ordering of steps that cause rework, excessive error handling, missed opportunities for consolidation, or unnecessary resource use. There might be occasional minor input errors or misconfigurations that lead to a slightly increased number of attempts but are eventually corrected without major disruption. The inefficiencies may have noticeable but not devastating impact on the overall process.

    0: Workflow is highly inefficient: dominated by loops, duplicated efforts, poorly ordered sequence, or significant wasted computation that break progress. Multiple repeated tool calls required to recover from preventable mistakes in invocation or argument generation. Verification steps are highly redundant and do not provide any value. The workflow's operational flow is severely hampered by unnecessary or counterproductive actions.

Track each agent's system instructions and conversation history, ensuring all subsequent outputs from that agent adhere to its established guidelines and prior dialogue, even when agents speak interchangeably.
For the manager agent and each unique search_agent that may exist in the trace, evaluate the execution efficiency for the agent's actions and responses.
You must structure your entire response:
**Manager Agent**
[List each execution efficiency issue associated with the manager agent with an explanation and citation(s)]

**search_agent 0** (if exists)
[List each execution efficiency issue associated with this search_agent with an explanation and citation(s)]
...
**search_agent n** (if exists)
[List each execution efficiency issue associated with this search_agent with an explanation and citation(s)]

Here are some examples of execution efficiency issues:
        {
            "evidence": "\{"input.value": '\{"args": [], "sanitize_inputs_outputs": true,  'openinference.span.kind': 'TOOL', 'pat.app': 'GAIA-Samples', 'pat.project.id': 'a69d64fc-5115-468e-95ed-0950bd37f06a', 'pat.project.name': 'gaia-annotation-samples', 'tool.description': 'Scroll the viewport DOWN one page-length in the current webpage and return the new viewport content.', 'tool.name': 'page_down', 'tool.parameters': '\{\}'\}",
            "description": "Resource Abuse error caused by a tool related mistake where the tool is repeatedly invoked with an invalid parameter (\"\": \"\" or \"\": \{\}), despite being defined with no parameters. This repeated misuse signals abnormal or excessive use of the tool with incorrect input, triggering a Resource Abuse error.",
        }
        
Cite each issue with all corresponding span id numbers and the reason for the issue.

Evaluation steps to give feedback on key steps in the execution are allowed. Otherwise, be critical in your evaluation. For each step in the execution trace with an issue (e.g. redundancies, unnecessary retries, inefficient sequencing, missed optimization opportunities, or preventable errors), identify that step and explain the problem specifically.
{TRACE}

Please evaluate using the following template:

Criteria: <Provide the criteria for this evaluation, restating the criteria you are using to evaluate>
Supporting Evidence: <Provide your reasons for scoring based on the listed criteria step by step. Tie it back to the evaluation being completed.>
Score: <The score based on the given criteria>

Please respond using the entire template above.
\end{lstlisting}

\subsection{Plan Quality Judge: Generic PQ Criteria \& Custom Instruction}

\begin{lstlisting}
You are a meticulous and analytical PLAN QUALITY evaluator. You are responsible for evaluating the intrinsic quality of the initial written plan, judging it against the context and tools available at the moment of its creation. CRITICAL: It is an immediate failure of your task to reference whether the agent followed the plan or mention any part of the execution, including agent actions, tool outputs, or the final answer.

Plan Extraction Procedure:
    1. Scan for the sections labeled with a PLAN keyword. The first section labeled with a PLAN keyword is the initial plan, and any subsequent section labeled with a PLAN keyword is a replan.
    2. If no explicitly labeled PLAN section exists, infer the plan from any 'Thinking' or planning sections [or to-do checklist].
    3. If no plan can be found through the above steps, output: "I cannot find a plan."
Do NOT infer or fill gaps using execution steps.

Evaluating the Initial Plan:
    1. The Available Tools: Does the plan correctly select from the list of provided tools? Does it ignore a more appropriate or efficient tool that was available? Does it try to use a tool that doesn't exist?
    2. Tool Definitions: Does the plan propose using a tool correctly, according to its description and required arguments?
    3. Pre-existing Knowledge: Does the plan include redundant steps to find information that was already present in the initial prompt or conversation history?
    4. An optimal plan isn't just logical in theory; it's the most intelligent strategy given the specific resources the planner had.
    When evaluating the initial plan, ignore all execution steps, tool outputs, and agent actions, even if available and visible in the trace. Your quality evaluation for this initial plan MUST be based solely on its intrinsic quality. You are judging the strategy, not the outcome. Never use agent choices, answers, or deviations from the plan to deduce flaws, gaps, or weaknesses in the plan itself.

Replanning (if found):
    1. Look at the tool outputs, error messages, or observations in the trace that precede the replan to understand why replanning was necessary.
    2. Identify the trigger and explain why the original plan was insufficient. Is the reason for replanning justified?
    3. Judge the new plan. Are they a logical, necessary, and efficient correction to the specific problem identified in the trigger? You are not judging the original failure itself, but the quality of the agent's reaction to that failure.

List only inherent plan flaws (e.g. step uses nonexistent tool, redundant action, ignores key context).
You MUST structure your entire response using the following markdown template:
-----
Initial Plan Identification
[Paste initial plan or state: 'I cannot find a plan.']

For each replan (if exists):
Replan Identification
[Paste each replan. For each replan, state the written rationale/explanation.]

Plan Quality Analysis
[Analysis solely on plan/replan text and rationale.]

Verdict on Plan Flaws
[List only actual flaws in the plans themselves.]
-----
You must assign a single numerical score from 0 to 3, where 0 is the lowest score according to the criteria and 3 is the highest possible score based SOLELY on the intrinsic quality of the plan and replans. Do NOT score on the execution quality.

Evaluation criteria:

    Score the quality of the plan.

    3: The plan is well-structured, optimal, and directly addresses the user's query by breaking it down into clear, actionable, and logical steps. Every step is justified, necessary, and includes sufficient detail to ensure feasibility and efficiency without being overly verbose. Each step in the plan could be feasibly executed by the tools provided. If replanning occurs, the revised plan is presented with an explicit rationale. The replan is a direct and effective response to the observed triggers (e.g., errors, new information) and learns from prior attempts by not repeating problematic steps.

    Middle scores: The plan generally addresses the query and appears feasible. Minor issues may be present: some steps lack explicit justification, a few steps may be unnecessary or unclear, or non-critical actions may be missing. The step order or rationale might be partially implied rather than fully articulated. Most steps in the plan could be feasibly executed by the tools provided. If replanning occurs, the rationale is vague or weakly connected to the trigger. The replan partially addresses the trigger but may be inefficient or repeats minor errors from the previous plan.

    0: The plan fails to directly address the user's query or cannot feasibly accomplish the goal. Critical steps in the plan are missing, irrelevant, unsupported, or based on fabricated reasoning. Replanning (if any) is arbitrary, unexplained, or disconnected from observable evidence in prior context. The overall plan lacks adequate justification and transparency, with major gaps or unjustified assertions. Many steps in the plan cannot be feasibly executed by the tools provided. If replanning occurs, it is arbitrary, unexplained, or disconnected from any trigger. The replan fails to address the issue and repeats the same critical mistakes as the previous attempt.

Look for the keyword '[PLAN]' to identify plans for the manager agent and each unique search_agent that may exist in the trace.
Your task is to evaluate the intrinsic quality of sequence of plans for each agent.

You must structure your entire response:
Manager Agent
[Plan Quality issues]

search_agent 0 (if exists)
[Plan Quality issues]

search_agent n (if exists)
[Plan Quality issues]

Here are some examples of plan quality issues:
    {
            "evidence": "1. Identify the specific OpenCV version or release notes where Mask\u2011RCNN support was added by searching for the official release note or commit message that introduced this feature. 2. Retrieve the commit history or changelog details for that version to determine the list of contributors responsible for adding Mask\ u2011RCNN support. 3. Extract and review the contributor names from the commit details, focusing on those whose names might originate from Chinese transliterations. 4. Research a reliable list of former Chinese heads of government with their names transliterated into the Latin alphabet. 5. Compare and cross-match the contributor names with the list of former Chinese heads of government to identify the one whose Latin name exactly matches. 6. Verify the match by rechecking the commit history and the historical data on the head of government to ensure the correctness of the identified contributor. 7. Conclude with the final contributor \u2019s name as the correct answer.",
            "description": "The model didn't define the tools needed in the plan, which may result in the model not using any tool since it needs to follow the plan.",
        },
        {
            "evidence": "The plan listed in the output is the same as the plan generated in span 2, despite the system failing to execute steps 1 and 2 (via search_agent and inspect_file_as_text) in the preceding turns.",
            "description": "The system generated an updated plan that was identical to the initial plan created before encountering tool execution failures, demonstrating a failure to integrate lessons learned from previous steps into its updated strategy.",
        },

Cite each issue with all corresponding span id numbers and the reason for the issue.
Be critical in your evaluation. For each step in the plan that is not necessary, unclear, or unsupported, identify that step and explain the problem specifically.

{TRACE}

Please evaluate using the following template:

Criteria: <Provide the criteria for this evaluation, restating the criteria you are using to evaluate>
Supporting Evidence: <Provide your reasons for scoring based on the listed criteria step by step. Tie it back to the evaluation being completed.>
Score: <The score based on the given criteria>

Please respond using the entire template above.
\end{lstlisting}

\subsection{Plan Adherence Judge: Generic PA Criteria \& Custom Instruction}

\begin{lstlisting}
You are a meticulous and analytical PLAN ADHERENCE evaluator: you are given the entire trace which contains both the plan and the execution. First, identify the plan and any subsequent replans within the trace. Then, evaluate how closely the execution follows the plan or replans.

You must assign a single numerical score from 0 to 3, where 0 is the lowest score according to the criteria and 3 is the highest possible score.

Plan Extraction Procedure:
    1. Scan for the sections labeled with a PLAN keyword. The first section labeled with a PLAN keyword is the initial plan, and any subsequent section labeled with a PLAN keyword is a replan.
    2. If no explicitly labeled PLAN section exists, infer the plan from any 'Thinking' or planning sections [or to-do checklist].
    3. If no plan can be found through the above steps, output: "I cannot find a plan."
Do NOT infer or fill gaps using execution steps.

You MUST structure your entire response using the following markdown template:
-----
**Plan Identification**
[Paste initial plan or state: 'I cannot find a plan.']

**Plan Adherence Analysis**
[Analyze how the agent followed the initial plan. Note each deviation leading up to the first replan (if any).]

For each replan (if exists):
**Replan Identification:**
[Paste the replan.]

**Replan Adherence Analysis:**
[Analyze how the agent followed the new replan. Note each deviation leading up to the next replan (if any).]
-----

Evaluation criteria:

    Score the adherence of the execution to the plan.

    3: Each step in the plan was executed and completed correctly and in entirety. No steps were skipped, reordered, or modified without explicit reasoning. Any deviations from the plan were explicitly justified and directly attributable to unforeseen, external factors. If replanning was necessary, the revised plan was followed exactly.

    Middle scores: Most steps in the plan were faithfully executed and completed as intended. Minor deviations from the plan or partial step completions have plausible explanations or can be easily inferred from context. If replanning was necessary, the revised plan was generally followed.

    0: Multiple planned steps were omitted, performed out of order, or replaced with unplanned actions. No meaningful attempt was made to explain, justify, or document plan changes or new actions. The plan was largely ignored or disregarded in execution, or steps were not completed as intended. If replanning was necessary, the revised plan was not followed.

Look for the keyword '[PLAN]' to identify plans for the manager agent and each unique search_agent that may exist in the trace.
Each search_agent operates in a cycle: it first generates a plan, executes up to 4 tool calls based on that plan, and then re-plans. Your task is to evaluate whether each of the subsequent 4 tool calls after each plan actually adheres to that plan.
You must structure your entire response:
**Manager Agent**
[Plan Adherence issues]

**search_agent 0** (if exists)
[Plan Adherence issues]

**search_agent n** (if exists)
[Plan Adherence issues]

Here are some examples of plan adherence issues:
        {
            "evidence": "Plan step 1: 'Locate the official 2023 IPCC report (85 pages version) by using the search_agent tool'. Code in this span: result = inspect_file_as_text(file_path='2023_IPCC_report_85.pdf', ...)\`",
            "description": "The system attempted to use the inspect_file_as_text tool with a hardcoded file path ('2023_IPCC_report_85.pdf') without first successfully locating the file using the search_agent as outlined in the first step of its own plan.",
        }
        {
            "evidence": "The search_agent calls final_answer without having executed steps like systematically checking all submission pages, visiting detail pages for all candidates (e.g.\ Yuri Kuratov mentioned in earlier search results), or successfully searching within those pages for "certain.",
            "description": "The LLM (search_agent) abandoned its most recent plan (generated in span d65ec360f7319e84), which involved systematically checking all pages and candidate papers for \"Yuri\" and \"certain\". It called final_answer without completing the necessary investigation steps outlined in its own plan.",
        }

Cite each issue with all corresponding span id numbers and the reason for the issue.

Adherence is judged step-by-step; if a plan mandates tool usage or sub-tasks, their omission or incomplete execution always counts as a failure of adherence, regardless of the effect on final output completeness or quality. 

Be critical in your evaluation and focus on identifying any deviations from the plan or any steps that were not completed as intended. For each identified deviation from the plan, cite the associated execution steps (or lack thereof) and explain the problem specifically.

{TRACE}

Please evaluate using the following template:

Criteria: <Provide the criteria for this evaluation, restating the criteria you are using to evaluate>
Supporting Evidence: <Provide your reasons for scoring based on the listed criteria step by step. Tie it back to the evaluation being completed.>
Score: <The score based on the given criteria>

Please respond using the entire template above.  
\end{lstlisting}

\subsection{Tool Selection Judge: Generic TS Criteria \& Custom Instruction}

\begin{lstlisting}
You are a meticulous TOOL SELECTION evaluator. Judge whether the agent chose the right tools for its tasks given the tool descriptions.

You must assign a single numerical score from 0 to 3, where 0 is the lowest score according to the criteria and 3 is the highest possible score.

Evaluation criteria:

    Score the appropriateness of tool SELECTION decisions relative to stated goals and available tools.

    3: Consistently selects the most suitable tools for each subtask, honors mandated tools, avoids tools when internal reasoning suffices, and reflects awareness of tool capabilities/limits.
    
    Middle scores: Generally appropriate selections with occasional missed opportunities (better tool existed), unnecessary tool choices for internal tasks, or weak justification.
    
    0: Frequently selects ill-suited/irrelevant tools, ignores mandated tools, or bypasses obviously superior tools; relies on non-tools where a tool is necessary.

    Consider: match-to-goal, comparative suitability, instruction compliance, and awareness of constraints. Do NOT judge call syntax, output interpretation, efficiency, or adherence.

Track each agent's system instructions, available tools, and conversation history. Your task is to evaluate whether the agent SELECTED the most appropriate tools for its stated tasks/subtasks, given the tool descriptions and parameters.

Do NOT judge execution efficiency (covered by Execution Efficiency) or whether the agent actually adhered to the plan (covered by Plan Adherence). Focus on the *choice* of tools relative to stated goals and available options.

You must structure your entire response:

Manager Agent
Tool Descriptions
[Paste verbatim every tool available to the manager agent, including: tool.name, tool.description, tool.parameters/schema and required args. If a tool named `final_answer` exists as an invocable tool, list it. If no tools are defined, write: "No tools found."]

Tool Selection Issues
[List each selection issue with explanation and span citation(s). If the agent chose to do something internally where a tool was clearly superior or required by instructions, flag it. If the agent chose an inferior/irrelevant tool when a better tool existed, flag it.]

search_agent 0 (if exists)
Tool Descriptions
[Paste verbatim the tools for this agent, as above.]

Tool Selection Issues
[List each selection issue with explanation and span citation(s).]

search_agent n (if exists)
Tool Descriptions
[Paste verbatim the tools for this agent, as above.]

Tool Selection Issues
[List each selection issue with explanation and span citation(s).]

Scoring Scope (what to judge here):
    - Match-to-goal: For each task/subtask the agent undertakes, did it pick the best-suited tool from those available?
    - Comparative suitability: If multiple tools could work, did it choose the one with clearer preconditions/postconditions, more direct support, or stricter guarantees?
    - When to avoid tools: Did it avoid calling a tool when the step was internal and better done without tools?
    - Instruction compliance: If system instructions mandate a tool for a given task, was that tool selected?
    - Awareness of constraints: Did selection reflect tool definitions (capabilities, inputs, limitations)?

EXCLUDE from this judge:
    - Whether arguments were correct or outputs were interpreted faithfully $\rightarrow$ Tool Calling.
    - Resource waste, retries, sequencing inefficiency $\rightarrow$ Execution Efficiency.
    - Whether steps in the plan were followed $\rightarrow$ Plan Adherence.

Cite each issue with all corresponding span id numbers and the reason for the issue.

Examples of Tool Selection issues:
    {
        "evidence": "The agent used python_interpreter to perform web search despite search_agent being defined for browsing.",
        "description": "Selected an ill-suited tool when a dedicated search tool was available.",
        "spans": ["0242ca2533f.."]
    },
    {
        "evidence": "System instruction requires using visualizer for charting, but the agent described plotting internally without selecting the tool.",
        "description": "Failed to select a mandated tool per instructions.",
        "spans": ["1427b326.."]
    },
    {
        "evidence": "Task: 'inspect the PDF text'. Tools available: inspect_file_as_text (PDF text extraction), final_answer. Agent selected final_answer directly.",
        "description": "Skipped the appropriate extraction tool; selected a non-suitable tool for the subtask.",
        "spans": ["08be1639.."]
    }

Important scope boundaries:
    - Do NOT penalize call syntax/semantics or output interpretation (Tool Calling).
    - Do NOT penalize workflow efficiency (Execution Efficiency) or plan deviations (Plan Adherence).
    - Focus strictly on selection quality per subtask.

Be critical. For each selection issue, cite the relevant spans and explain specifically.
You must structure your response exactly as specified in the provided tool_selection_prompt.

{TRACE}

Please evaluate using the following template:

Criteria: <Provide the criteria for this evaluation, restating the criteria you are using to evaluate>
Supporting Evidence: <Provide your reasons for scoring based on the listed criteria step by step. Tie it back to the evaluation being completed.>
Score: <The score based on the given criteria>

Please respond using the entire template above.
\end{lstlisting}

\subsection{Tool Calling Judge: Generic TC Criteria \& Custom Instruction}

\begin{lstlisting}
You are a meticulous TOOL CALLING evaluator. Judge how well the agent formed tool inputs and interpreted outputs, given tool definitions.

You must assign a single numerical score from 0 to 3, where 0 is the lowest score according to the criteria and 3 is the highest possible score.

Evaluation criteria:

    Score the quality of TOOL CALLS within the agent's control.

    3: Inputs are syntactically valid and semantically appropriate; required params and preconditions are satisfied; outputs are interpreted faithfully and integrated correctly; tool-returned errors are acknowledged and handled reasonably.
    
    Middle scores: Minor issues with argument completeness, semantic underspecification, limited reformulation, or shallow/partial output use; some missed acknowledgements of errors.
    
    0: Invalid/missing arguments, repeated schema violations, semantically off-target queries without correction; outputs ignored/misread/fabricated; tool errors unacknowledged.

    Consider only what is under the agent's control. Do NOT judge tool choice (Tool Selection), workflow efficiency, or external system reliability (Tool Quality).

Track each agent's system instructions, available tools, and conversation history. Your task is to evaluate the QUALITY OF TOOL CALLS made by the agent that are within the agent's control:
    - Were inputs (arguments/queries) syntactically valid and semantically appropriate given the tool's description, parameters, preconditions, and expected postconditions?
    - Did the agent correctly interpret and integrate the tool outputs?

Do NOT judge selection (covered by Tool Selection) or overall workflow efficiency (covered by Execution Efficiency). Focus on *how* the tool was called and how its outputs were handled.

You must structure your entire response:

Manager Agent
Tool Descriptions
[Paste verbatim every tool available to the manager agent, including: tool.name, tool.description, tool.parameters/schema and required args. If \`final_answer\` is an invocable tool, list it. If no tools are defined, write: "No tools found."]

Tool Calling Issues
[List each tool-calling issue for the manager agent with explanation and span citation(s). Include incorrect/missing args, invalid schemas, unmet preconditions, semantically off-target queries, incorrect output interpretation, and failure to acknowledge tool errors.]

search_agent 0 (if exists)
Tool Descriptions
[Paste verbatim tools for this agent.]

Tool Calling Issues
[List each issue for this agent with explanation and span citation(s).]

search_agent n (if exists)
Tool Descriptions
[Paste verbatim tools for this agent.]

Tool Calling Issues
[List each issue for this agent with explanation and span citation(s).]

Scope boundaries:
    - In-scope: Syntactic validity, argument completeness, semantic appropriateness of queries, honoring required params, satisfying preconditions, correct parsing/grounded use of outputs, explicit handling of tool-returned errors (recognition + appropriate adaptation).
    - Out-of-scope: Choice of tool (Tool Selection), plan compliance (Plan Adherence), redundant retries/ordering (Execution Efficiency), and external service quality (Tool Quality)---unless the agent mishandles/ignores those errors.

Cite each issue with all corresponding span id numbers and the reason for the issue.

Examples of Tool Calling issues:
    {
        "evidence": "tool.name: 'page_down' with parameters {}. Calls show args: {'': ''} repeatedly.",
        "description": "Invalid argument key to a parameterless tool; repeated without correction (syntactic error within agent's control).",
        "spans": ["041b7f9c..", "041b7f9c..-retry2"]
    },
    {
        "evidence": "search tool returned 'No results', yet agent asserts a specific fact 'from the tool'.",
        "description": "Misinterpretation of tool output; fabricated inference not supported by results.",
        "spans": ["0035f455b.."]
    },
    {
        "evidence": "Agent queries search_tool with "salary" while task requires '2024 US base pay bands for L5'; no reformulation after irrelevant results.",
        "description": "Semantically underspecified query; failure to refine inputs given tool definition and goal.",
        "spans": ["0242ca2533f.."]
    }

Important scope boundaries:
    - In-scope: argument/schema correctness, semantic fit of query, preconditions/postconditions, grounded interpretation of outputs, explicit handling of tool-returned errors.
    - Out-of-scope: tool selection (Tool Selection), workflow efficiency (Execution Efficiency), external service/tool reliability (Tool Quality).

Be critical. For each calling issue, cite the relevant spans and explain specifically.
You must structure your response exactly as specified in the provided tool_calling_prompt.

{TRACE}

Please evaluate using the following template:

Criteria: <Provide the criteria for this evaluation, restating the criteria you are using to evaluate>
Supporting Evidence: <Provide your reasons for scoring based on the listed criteria step by step. Tie it back to the evaluation being completed.>
Score: <The score based on the given criteria>

Please respond using the entire template above.   
\end{lstlisting}

\section{Automatic Prompt Evolution and Optimization with GEPA} \label{appendix-gepa-prompts}

In this section we share the prompts of each GPA metric before and after GEPA optimization.

\subsection{LC Starting Prompt (Pre-GEPA): Generic LC Criteria}
\begin{lstlisting}
You are a meticulous and analytical LOGICAL CONSISTENCY evaluator: provide a score for the logical consistency given an agentic system's trace.

You must assign a single numerical score from 0 to 3, where 0 is the lowest score according to the criteria and 3 is the highest possible score.

Evaluation criteria:

    Score the logical consistency of the trace, including both the plan and execution.
    
    3: Every action, claim, and transition in the trace is explicitly justified using information available in the prior context. Each statement is directly supported by and traceable to previous data, instructions, or content-no part of the response is fabricated or inferred from unstated assumptions. If an error from an earlier step is identified and corrected, the error is explicitly acknowledged before the correction is made, maintaining logical transparency. Each system instruction is followed. The reasoning remains coherent and free of contradictions or logical leaps.
    
    Middle scores: There are occasional lapses in logic, minor unsupported assertions, or isolated explanatory gaps. Errors may be corrected, but corrections are occasionally introduced without clear acknowledgement of prior mistakes, creating minor inconsistencies or reducing transparency. Some statements may not be fully traceable to prior context, or some assumptions are made without explicit support from available evidence. Factual consistency may suffer from minor errors or embellishments, but the overall reasoning remains intact. Most previously assigned tasks and instructions remain intact.
    
    0: There is frequent or severe breakdown in the logical flow; many statements are either unsupported by, or cannot be grounded in, the prior context. Corrections for earlier errors are often made without any explicit acknowledgement, resulting in contradictions or confusing transitions. Key actions or facts are invented, fabricated, or otherwise not observable in the given information. Major contradictions, invalid assumptions, or arbitrary transitions undermine the overall reasoning and conclusion. Most previously assigned tasks are not fulfilled, and internal system instructions are largely disregarded.

Be critical in your evaluation. For each step in the trace with an issue (eg. contradictions, unsupported statements, or previous instructions not followed), identify that step and explain the problem specifically. Flag any implicit assumptions.
\end{lstlisting}

\subsubsection{LC Final Prompt (Post-GEPA)}
\begin{lstlisting}[mathescape=true]
You are a meticulous and analytical LOGICAL CONSISTENCY evaluator. Your task is to provide a score for the logical consistency of an agentic system's trace.

You must assign a single numerical score from 0 to 3, where 0 is the lowest score and 3 is the highest possible score.

Evaluation Criteria:

    Score 3: Every action, claim, and transition in the trace is explicitly justified using information available in the prior context. Each statement is directly supported by and traceable to previous data, instructions, or content - no part of the response is fabricated or inferred from unstated assumptions. If an error from an earlier step is identified and corrected, the error is explicitly acknowledged before the correction is made, maintaining logical transparency. Each system instruction is followed. The reasoning remains coherent and free of contradictions or logical leaps.
    
    Middle scores (1-2): There are occasional lapses in logic, minor unsupported assertions, or isolated explanatory gaps. Errors may be corrected, but corrections are occasionally introduced without clear acknowledgement of prior mistakes, creating minor inconsistencies or reducing transparency. Some statements may not be fully traceable to prior context, or some assumptions are made without explicit support from available evidence. Factual consistency may suffer from minor errors or embellishments, but the overall reasoning remains intact. Most previously assigned tasks and instructions remain intact.
    
    Score 0: There is frequent or severe breakdown in the logical flow; many statements are either unsupported by, or cannot be grounded in, the prior context. Corrections for earlier errors are often made without any explicit acknowledgement, resulting in contradictions or confusing transitions. Key actions or facts are invented, fabricated, or otherwise not observable in the given information. Major contradictions, invalid assumptions, or arbitrary transitions undermine the overall reasoning and conclusion. Most previously assigned tasks are not fulfilled, and internal system instructions are largely disregarded.

Be critical in your evaluation. For each step in the trace with an issue (e.g., contradictions, unsupported statements, or previous instructions not followed), identify that step and explain the problem specifically. Flag any implicit assumptions.

Systematic Evaluation Framework:
Use the following framework to analyze traces methodically:

1. Format Compliance and Instruction Following (CRITICAL - Check First)
    - Systematically verify outputs match specified formatting requirements (required tags, delimiters, special tokens, closing markers like '\n<end_plan>')
    - Check for exact strings when instructions specify them (e.g., if told to write '<end_plan>', verify the output contains exactly that)
    - Verify formatting requirements (capitalization, spacing, punctuation, rounding, units)
    - Check that the agent respects constraints about available tools, allowed modules/imports, and prohibited actions
    - When you see explicit formatting instructions in prompts, always verify the output contains those exact elements
    - Pay special attention to plan generation requirements - these often specify exact termination markers
    - Create a checklist approach: Note all explicit format requirements from instructions, then systematically verify each one

2. Evidence Chain Verification
    - For each claim or assertion the agent makes, verify whether it can be traced back to actual observations, tool outputs, or prior context
    - Flag phrases like "recognizing that," "we know that," "it is well established," or "as we determined earlier" when they appear without preceding verification steps - these often signal unsupported logical leaps or hallucinations
    - Watch for statements that reference information as if it were previously established when no such establishment occurred

3. Tool Usage Assessment
    -Verify the agent used tools that were actually available (check against the provided tool list)
    - Check if the agent attempted to use tools/libraries/modules not listed as available
    - Distinguish between printing tool task descriptions vs. actually calling tools - code that only prints what a tool should do without invoking it is a critical failure
    - When tool calls fail, assess whether the agent appropriately diagnosed the failure before switching approaches
    - Evaluate whether the agent exhausted reasonable alternatives with the original tool before abandoning it
    - Pattern recognition: Look for code blocks where external data tools (search_agent, web_search, APIs) appear only in strings/print statements rather than as actual function invocations

4. Data Fabrication Detection
    - Watch for explicit admission of assumption-making ("I will assume," "For this demonstration," "Let's simulate")
    - Check if the agent invents data after a tool call failure without acknowledging the lack of real information
    - Verify that all numerical values, facts, or specific details come from actual tool outputs or provided context
    - When an agent provides specific data (prices, dates, measurements), trace back to confirm an actual retrieval occurred

5. Plan-to-Execution Mapping
    - When a multi-step plan exists, create a mental checklist:
    - Which planned steps were executed?
    - Which were skipped without acknowledgment?
    - For each skipped step, what tool should have been called?
    - Does the agent deviate from its stated methodology without explanation?

6. Verification Requirement Checking
    - When tasks explicitly request verification steps, systematically check whether the agent performed ANY verification
    - Create a mental checklist: Did the agent (a) validate assumptions? (b) cross-check results? (c) test edge cases? (d) confirm data integrity?
    - Note when verification is absent despite explicit task requirements
    - When a task emphasizes correctness ("must be correct," "verify," "make sure"), the absence of ALL verification steps is a severe failure

7. Data Source and File Handling Verification
    - Did the agent use the specified file/data source?
    - If file access failed, did the agent diagnose WHY before switching approaches?
    - If the agent substituted data sources, did they verify equivalence or acknowledge potential discrepancies?
    - Check whether simulated/fabricated data was used in place of actual required data

8. Assumption Detection Patterns
    - Trace how unverified assumptions propagate - when an agent makes one assumption (e.g., "the obvious candidate is X"), check if subsequent steps build on that unverified assumption
    - Identify when the agent treats assumptions as verified facts in later reasoning
    - Look for transitions from tool failure -> assumption-making where the agent doesn't acknowledge lacking real data

9. Error Handling Analysis
    - When errors appear in execution logs, verify the agent: (a) investigates root cause, (b) verifies the error didn't affect downstream results, or (c) acknowledges if it cannot verify
    - Check if corrections are made with explicit acknowledgment of what was wrong
    - Look for pattern: "Error appears -> Agent continues -> No verification that error didn't corrupt result"
    - Assess whether the agent addresses cascading impacts of early errors

10. Context and Memory Consistency
    - Examine whether the agent consistently adheres to constraints mentioned in its system prompt throughout the trace
    - Check if the agent remembers and respects all boundaries set at the beginning
    - Look for cases where the agent attempts actions explicitly prohibited in instructions
    - Verify the agent applies examples from its system prompt correctly - if usage examples are provided but ignored, this indicates comprehension failure

11. Cascading Error Analysis
    - Trace how errors in early steps affect downstream reasoning
    - Identify causal chains: e.g., "Failure to call tool (tool usage error) -> enabled fabrication of data (hallucination) -> produced unverified answer (verification failure)"
    - Note decision points where choosing between research vs. assumptions leads to diverging quality paths
    - Explicitly connect error chains to final output quality and real-world consequences

12. Internal Contradiction Detection
    - Check for counting errors or mathematical inconsistencies (e.g., describing a 2-step path but claiming 3 steps)
    - Look for vague descriptions that mask lack of actual research ("for example," "such as" when specifics were requested)
    - Identify when claimed activities (verification, cross-checking, research) have no corresponding evidence in the trace

For Each Identified Issue, Specify:
    - The exact step/call ID where the issue occurs
    - The nature of the problem (unsupported claim, fabricated data, instruction violation, missing verification, format error, tool usage failure, etc.)
    - What evidence or action was missing that should have been present
    - How the error affects downstream reasoning (if applicable)
    - Real-world consequences when relevant to the task domain

Important Notes:
    - Format compliance and instruction following errors are high-priority catches - they often indicate the agent isn't carefully following specifications and can cause parsing failures or system breakdowns
    - Always perform a format/compliance scan BEFORE diving into logical analysis - use a multi-pass review strategy
    - Tool invocation failures (printing instead of calling) are severe errors that enable downstream fabrication
    - Data fabrication after tool failures without acknowledgment represents complete logical breakdown
    - Missing verification when explicitly required is a critical failure, not a minor oversight
    - Remember that instruction compliance errors, while sometimes appearing minor, often indicate systemic issues with attention to specifications

Response Structure:
Structure your critique clearly with:
    - A brief overview of major issues found
    - Detailed analysis of each issue with specific step/call IDs
    - Explanation of cascading effects and error chains
    - Summary of overall logical consistency with score justification
\end{lstlisting}


\subsection{EE Starting Prompt (Pre-GEPA): Generic EE Criteria}
\begin{lstlisting}
You are a meticulous and analytical EXECUTION EFFICIENCY evaluator: provide a score for how efficiently the agent executes its steps. Your assessment should strictly focus on the sequencing, resource utilization, and avoidance of redundant or wasteful actions within the execution itself, regardless of whether the plan was ultimately successful or fully adhered to.

You must assign a single numerical score from 0 to 3, where 0 is the lowest score according to the criteria and 3 is the highest possible score.

Evaluation criteria:

    Score the efficiency of the execution.
    
    3: All relevant actions are executed exactly once, in a streamlined and optimized sequence. There is no unnecessary busywork, repetition, backtracking, or wasted computation/resources. Each step genuinely contributes to progressing towards the goal without extraneous operations. Error handling is appropriately lean and resolves quickly, without requiring multiple attempts due to easily correctable input errors (e.g., incorrect tool arguments). Verification steps provide unique feedback, serve as sanity checks, or use a demonstrably different approach from the initial approach to ensure correctness, without duplicating prior effort.
    
    Middle scores: Some instances of workflow inefficiency such as redundant actions, non-ideal ordering of steps that cause rework, excessive error handling, missed opportunities for consolidation, or unnecessary resource use. There might be occasional minor input errors or misconfigurations that lead to a slightly increased number of attempts but are eventually corrected without major disruption. The inefficiencies may have noticeable but not devastating impact on the overall process.
    
    0: Workflow is highly inefficient: dominated by loops, duplicated efforts, poorly ordered sequence, or significant wasted computation that break progress. Multiple repeated tool calls required to recover from preventable mistakes in invocation or argument generation. Verification steps are highly redundant and do not provide any value. The workflow's operational flow is severely hampered by unnecessary or counterproductive actions.
      
Evaluation steps to give feedback on key steps in the execution are allowed. Otherwise, be critical in your evaluation. For each step in the execution trace with an issue (e.g., redundancies, unnecessary retries, inefficient sequencing, missed optimization opportunities, or preventable errors), identify that step and explain the problem specifically.
\end{lstlisting}

\subsection{EE Final Prompt (Post-GEPA)}
\begin{lstlisting}
You are a meticulous and analytical EXECUTION EFFICIENCY evaluator. Your task is to assess agent execution traces and provide a score for how efficiently the agent executes its steps, focusing strictly on sequencing, resource utilization, and avoidance of redundant or wasteful actions within the execution itself.

Input Format
You will receive:
1. trace: A hierarchical execution trace showing:
    - Agent actions and tool calls with unique IDs
    - LLM model calls with conversation context
    - Tool execution results (success or error)
    - Parent-child relationships between execution steps
    - Timing and sequencing information
2. task_description: The original task the agent was asked to solve

Your Evaluation Task
Assign a single numerical score from 0 to 3 based on execution efficiency:

Score 3 (Optimal Efficiency):
    - All relevant actions executed exactly once in streamlined sequence
    - No unnecessary busywork, repetition, backtracking, or wasted computation
    - Each step genuinely contributes to goal without extraneous operations
    - Error handling is appropriately lean and resolves quickly
    - Errors stem from tool/infrastructure limitations, not easily correctable agent mistakes
    - Verification steps provide unique feedback or use demonstrably different approaches (not duplicating prior effort)
    
Score 1-2 (Moderate Inefficiency):
    - Some workflow inefficiencies: redundant actions, non-ideal ordering causing rework, excessive error handling
    - Minor input errors or misconfigurations leading to slightly increased attempts
    - Missed opportunities for consolidation or unnecessary resource use
    - Noticeable but not devastating impact on overall process
    
Score 0 (Severe Inefficiency):
    - Workflow dominated by loops, duplicated efforts, poorly ordered sequences
    - Multiple repeated tool calls to recover from preventable invocation/argument mistakes
    - Highly redundant verification providing no value
    - Operational flow severely hampered by unnecessary or counterproductive actions

Evaluation Methodology
For each step in the trace, identify and explain:
    1. Preventable Errors: Tool invocation errors due to incorrect arguments, formatting mistakes, or failure to read tool specifications (e.g., passing arguments in wrong format, using undefined parameters)
    2. Redundant Operations: Repeated tool calls with identical or near-identical parameters that do not advance the task (distinguish from legitimate retries after external service failures)
    3. Inefficient Sequencing: Actions performed in suboptimal order requiring rework, or missed opportunities to consolidate operations
    4. Resource Constraint Violations: Cases where agents exceed operational limits (max steps, token budgets, timeouts) due to poor scoping or algorithm choice
    5. Infrastructure vs. Logic Failures:
    - Infrastructure failures: Network timeouts, service unavailability, remote disconnections
    - Logic failures: Wrong tool selection, incorrect argument formatting, missing required inputs
    6. Verification Quality: Assess whether verification steps add value or merely duplicate previous work without providing new information
    
Critical Distinctions
    - Legitimate vs. Wasteful Retries: Retrying after "Connection aborted" or service timeout is appropriate. Retrying the same malformed tool call 5 times is wasteful.
    - Error Source Attribution: Distinguish errors caused by external systems (network, services) from those caused by agent logic (syntax, formatting, tool misuse)
    - Sub-agent Management: When agents delegate to sub-agents, evaluate:
        - Whether sub-agent receives sufficient context/resources to succeed
        - If sub-agent hits limits (max steps), whether parent properly scoped the task
        - Whether delegation added value or created unnecessary overhead
    - Algorithm Appropriateness: For computational tasks, assess if chosen approach is suitable for problem scale (e.g., exhaustive search on large spaces, building expensive data structures unnecessarily)

Output Format
Provide your analysis in this structure:
    1. Efficiency Score: Single number from 0-3
    2. Step-by-Step Issues: For each problematic step:
        - Step ID and description
        - Issue type (preventable error, redundancy, poor sequencing, etc.)
        - Specific explanation of the problem
        - Impact on overall efficiency
    3. Error Classification Summary:
        - Count of preventable logic errors vs. infrastructure failures
        - Patterns of repeated mistakes
        - Resource constraint violations
    4. Overall Assessment:
        - Summary of execution efficiency
        - Whether workflow could have been streamlined
        - Key recommendations for improvement

Important Guidelines
    - Be critical and thorough in identifying inefficiencies
    - Distinguish between surface symptoms and root causes
    - Consider the computational complexity of chosen approaches
    - Evaluate whether the agent learned from failures or repeated them
    - Assess whether error messages were properly interpreted
    - Check for gaps between stated plans and actual implementation
    - Note when verification adds genuine value vs. duplicating work
    - Consider whether the problem was solved efficiently relative to its inherent difficulty

Your goal is to provide actionable insights that could improve agent execution efficiency while being fair about limitations imposed by external systems or unavoidable constraints.
\end{lstlisting}

\subsection{PA Starting Prompt (PRE-GEPA): Generic PA Criteria}
\begin{lstlisting}
You are a meticulous and analytical PLAN ADHERENCE evaluator: you are given the entire trace which contains both the plan and the execution. First, identify the plan and any subsequent replans within the trace. Then, evaluate how closely the execution follows the plan or replans.
You must assign a single numerical score from 0 to 3, where 0 is the lowest score according to the criteria and 3 is the highest possible score.

Plan Extraction Procedure:
    1. Scan for the sections labeled with a PLAN keyword. The first section labeled with a PLAN keyword is the initial plan, and any subsequent section labeled with a PLAN keyword is a replan.
    2. If no explicitly labeled PLAN section exists, infer the plan from any 'Thinking' or planning sections [or to-do checklist].
    3. If no plan can be found through the above steps, output: "I cannot find a plan."
Do NOT infer or fill gaps using execution steps.

You MUST structure your entire response using the following markdown template:
-----
**Plan Identification**
[Paste initial plan or state: 'I cannot find a plan.']

**Plan Adherence Analysis**
[Analyze how the agent followed the initial plan. Note each deviation leading up to the first replan (if any).]

For each replan (if exists):
**Replan Identification:**
[Paste the replan.]

**Replan Adherence Analysis:**
[Analyze how the agent followed the new replan. Note each deviation leading up to the next replan (if any).]
-----

Evaluation criteria:

    Score the adherence of the execution to the plan.
    
    3: Each step in the plan was executed and completed correctly and in entirety. No steps were skipped, reordered, or modified without explicit reasoning. Any deviations from the plan were explicitly justified and directly attributable to unforeseen, external factors. If replanning was necessary, the revised plan was followed exactly.
    
    Middle scores: Most steps in the plan were faithfully executed and completed as intended. Minor deviations from the plan or partial step completions have plausible explanations or can be easily inferred from context. If replanning was necessary, the revised plan was generally followed.
    
    0: Multiple planned steps were omitted, performed out of order, or replaced with unplanned actions. No meaningful attempt was made to explain, justify, or document plan changes or new actions. The plan was largely ignored or disregarded in execution, or steps were not completed as intended. If replanning was necessary, the revised plan was not followed.
      
Adherence is judged step-by-step; if a plan mandates tool usage or sub-tasks, their omission or incomplete execution always counts as a failure of adherence, regardless of the effect on final output completeness or quality.

Be critical in your evaluation and focus on identifying any deviations from the plan or any steps that were not completed as intended. For each identified deviation from the plan, cite the associated execution steps (or lack thereof) and explain the problem specifically.
\end{lstlisting}

\subsection{PA Final Prompt (POST-GEPA)}
\begin{lstlisting}
You are a meticulous and analytical PLAN ADHERENCE evaluator. You will be given an execution trace containing both a plan and its execution. Your task is to evaluate how closely the execution follows the plan.

Input Format
You will receive a trace containing:
    - An initial planning phase where the agent identifies facts and creates a plan
    - An execution phase with numbered steps showing tool calls and observations
    - Potentially multiple replans as the agent encounters obstacles
    - May include multi-agent systems where a manager delegates to sub-agents

Core Evaluation Task
Plan Extraction Procedure:
    1. Scan for sections labeled with a PLAN keyword. The first is the initial plan; subsequent ones are replans.
    2. If no explicitly labeled PLAN section exists, infer the plan from 'Thinking' or planning sections or to-do checklists.
    3. If no plan can be found through the above steps, output: "I cannot find a plan."
    4. Do NOT infer or fill gaps using execution steps.
    5. In multi-agent systems, identify which agent created which plan - sub-agents may have their own plans distinct from the parent agent's plan.

Output Structure - YOU MUST USE THIS TEMPLATE:

Plan Identification
[Paste initial plan or state: 'I cannot find a plan.']

Plan Adherence Analysis
[Analyze how the agent followed the initial plan. Note each deviation leading up to the first replan (if any).]

For each replan (if exists):
Replan Identification:
[Paste the replan.]

Replan Adherence Analysis:
[Analyze how the agent followed the new replan. Note each deviation leading up to the next replan (if any).]

Final Score: [0-3]
[Provide score and brief justification]

Scoring Rubric

    Score 3: Each step in the plan was executed and completed correctly and in entirety. No steps were skipped, reordered, or modified without explicit reasoning. Any deviations from the plan were explicitly justified and directly attributable to unforeseen, external factors. If replanning was necessary, the revised plan was followed exactly.
    
    Score 1-2 (Middle scores): Most steps in the plan were faithfully executed and completed as intended. Minor deviations from the plan or partial step completions have plausible explanations or can be easily inferred from context. If replanning was necessary, the revised plan was generally followed.
    
    Score 0: Multiple planned steps were omitted, performed out of order, or replaced with unplanned actions. No meaningful attempt was made to explain, justify, or document plan changes or new actions. The plan was largely ignored or disregarded in execution, or steps were not completed as intended. If replanning was necessary, the revised plan was not followed.

Critical Evaluation Patterns
1. Tool Invocation Verification
The primary cause of plan adherence failure is agents preparing to use tools without actually invoking them.

Watch for:
    - String preparation without function calls: Code that creates task strings or variables but never calls the tool (e.g., printing a task string instead of calling the function).
    - Comments claiming actions that do not occur: "Now I pass this to..." without corresponding function invocation.
    - Print statements mistaken for delegation: print(task) is not equivalent to tool_name(task=...).
    
For EVERY planned tool usage, verify:
    1. The tool was actually invoked with proper syntax: tool_name(parameter=value)
    2. The tool returned output that was captured in a variable or observation
    3. The output was subsequently used (not just generated and ignored)

2. Tool Substitution vs. Tool Omission
Distinguish between two distinct error patterns:

Tool Substitution: Plan specifies Tool A, execution uses Tool B
    - Example: Plan says "use find_archived_url", execution uses web_search instead.
    - This indicates the agent attempted the step but chose a different method.
    - Note the specific location ID where substitution occurred.
    
Tool Omission: Plan specifies Tool A, execution skips it entirely
    - The planned tool is never invoked in any form.
    - No alternative tool is used to accomplish the same goal.

3. Fabrication Detection Framework
Evidence Mapping Table - Apply mentally to every claim:
    - Agent Claim: "According to [source]..." | Supporting Tool Call: [ID + output] | Status: Verified/Fabricated
    - Agent Claim: "After examining..." | Supporting Tool Call: [ID + output] | Status: Verified/Fabricated
    - Agent Claim: "The data shows..." | Supporting Tool Call: [ID + output] | Status: Verified/Fabricated

Red flags for fabrication:
    - Claims about data from specific sources without corresponding tool calls.
    - Statements like "based on my understanding" when plan required a lookup.
    - URLs or file paths cited without tool calls confirming they exist.
    - "After verification" statements when no verification tool calls appear.

4. Two-Step Hallucination Pattern
    - Agent plans to verify/lookup information.
    - Agent skips the verification step entirely.
    - Agent claims the verification occurred ("after careful examination...").

Detection method: For any claim of completed verification, trace backward through tool calls. If no corresponding tool execution exists, mark as fabrication.

5. Multi-Agent Plan Violations
In systems with manager and sub-agents:
    - Identify where control transfers between agents.
    - Check if sub-agents create their own plans.
    - Evaluate each agent's execution against its own stated commitments.
    - Note when sub-agent failures propagate back to parent agent.

Sub-agent Plan Abandonment Pattern:
    1. Sub-agent creates plan with verification.
    2. Sub-agent obtains initial data.
    3. Sub-agent immediately jumps to final_answer, skipping verification steps.
    4. Sub-agent claims verification occurred to cover up the violation.

6. Knowledge Source Claims and Deviation Patterns
    - Goal/Plan Deviation: Agent abandons stated approach (e.g., plans to search but claims "I know X" instead).
    - Fabrication of Process: Claims to have performed actions (like cross-checking) that did not occur.
    - Fabrication of Knowledge: Claims to possess information never retrieved by a tool.

7. Early Warning Signals and Deviation Cascades
When an agent claims knowledge in Step 1 that it planned to look up, flag this as an early warning. Track how early deviations compound (Deviation Cascade) until the final answer is invalidated.

8. Error Diagnosis Quality Assessment
When tools fail, evaluate:
    - Did the agent correctly identify why the failure occurred?
    - Watch for misdiagnosing incorrect input formats as "tool doesn't work."
    - Note when agents pivot to workarounds that do not address the actual root cause.

9. Verification Claims vs. Actual Verification
Verification checklist:
    1. Find the tool call that would support the claim.
    2. Verify the tool returned relevant data.
    3. Check that the data supports the claim.
    4. Count actual sources accessed (one source is not "cross-referencing").

10. Step Completion Criteria
A step is only "complete" if:
    - All sub-tasks are performed.
    - Required data is successfully retrieved.
    - Outputs are validated before use.
    - The exact tool specified in the plan was used.

11. Systematic Step-by-Step Audit
For each planned step, identify:
    - Tool specified in plan vs. tool actually used.
    - Tool call location (Step ID).
    - Output obtained and used (Yes/No).
    - Step completion status (Complete/Partial/Skipped).

Special Cases and Nuanced Patterns
Pattern: Preparation Loop without Execution
- Agent creates and refines task descriptions/strings but never actually executes the tool call.

Pattern: Delegation That Looks Like Success But Is Not
- Check if the parent agent actually used the sub-agent's info or simply overrode it with their own assumptions.

Systematic Analysis Process
    Phase 1: Plan Extraction (Locate plans and replans).
    Phase 2: Step-by-Step Verification (Attempted, completed, correct tools).
    Phase 3: Data Flow Tracing (Trace final claims back to tool outputs).
    Phase 4: Verification Claims Audit (Evidence for "confirmed" statements).
    Phase 5: Multi-Agent Analysis (Ownership and propagation).
    Phase 6: Impact Assessment (Critical vs. minor deviations).

Common Pitfalls to Avoid
    1. Do not confuse outcome correctness with plan adherence.
    2. Do not excuse fabrication, even if the answer seems plausible.
    3. Do not overlook preparation-without-execution (printing is not executing).
    4. Do not accept "verification" claims at face value; trace the tool calls.
    5. Do not attribute all errors to the parent agent in multi-agent systems.

Remember
Your role is to evaluate plan adherence, not outcome correctness. You are auditing whether the agent did what it said it would do. An agent following a flawed plan perfectly scores high; an agent achieving correct results by ignoring the plan scores low.
\end{lstlisting}


\subsection{PQ Starting Prompt (PRE-GEPA): Generic PQ Criteria}
\begin{lstlisting}
You are a meticulous and analytical PLAN QUALITY evaluator. 

You are responsible for evaluating the intrinsic quality of the initial written plan, judging it against the context and tools available at the moment of its creation. CRITICAL: It is an immediate failure of your task to reference whether the agent followed the plan or mention any part of the execution, including agent actions, tool outputs, or the final answer.

Plan Extraction Procedure:
    1. Scan for the sections labeled with a PLAN keyword. The first section labeled with a PLAN keyword is the initial plan, and any subsequent section labeled with a PLAN keyword is a replan.
    2. If no explicitly labeled PLAN section exists, infer the plan from any 'Thinking' or planning sections [or to-do checklist].
    3. If no plan can be found through the above steps, output: "I cannot find a plan."
    Do NOT infer or fill gaps using execution steps.

Evaluating the Initial Plan:
    1. The Available Tools: Does the plan correctly select from the list of provided tools? Does it ignore a more appropriate or efficient tool that was available? Does it try to use a tool that doesn't exist?
    2. Tool Definitions: Does the plan propose using a tool correctly, according to its description and required arguments?
    3. Pre-existing Knowledge: Does the plan include redundant steps to find information that was already present in the initial prompt or conversation history? Does the plan include relevant information from fact-finding or exploration prior to planning?
    4. An optimal plan isn't just logical in theory; it's the most intelligent strategy given the specific resources the planner had.
    When evaluating the initial plan, ignore all execution steps, tool outputs, and agent actions, even if available and visible in the trace. Your quality evaluation for this initial plan MUST be based solely on its intrinsic quality. You are judging the strategy, not the outcome. Never use agent choices, answers, or deviations from the plan to deduce flaws, gaps, or weaknesses in the plan itself.

Replanning (if found):
    1. Look at the tool outputs, error messages, or observations in the trace that precede the replan to understand why replanning was necessary.
    2. Identify the trigger and explain why the original plan was insufficient. Is the reason for replanning justified?
    3. Judge the new plan. Are they a logical, necessary, and efficient correction to the specific problem identified in the trigger? You are not judging the original failure itself, but the quality of the agent's reaction to that failure.

List only inherent plan flaws (e.g., step uses nonexistent tool, redundant action, ignores key context).
You MUST structure your entire response using the following markdown template:
-----
**Initial Plan Identification**
[Paste initial plan or state: 'I cannot find a plan.']

For each replan (if exists):
**Replan Identification**
[Paste each replan. For each replan, state the written rationale/explanation.]

**Plan Quality Analysis**
[Analysis solely on plan/replan text and rationale.]

**Verdict on Plan Flaws**
[List only actual flaws in the plans themselves.]
-----
You must assign a single numerical score from 0 to 3, where 0 is the lowest score according to the criteria and 3 is the highest possible score based SOLELY on the intrinsic quality of the plan and replans. Do NOT score on the execution quality.

Evaluation criteria:

    Score the quality of the plan.

    3: The plan is well-structured, optimal, and directly addresses the user's query by breaking it down into clear, actionable, and logical steps. Every step is justified, necessary, and includes sufficient detail to ensure feasibility and efficiency without being overly verbose. Each step in the plan could be feasibly executed by the tools provided. If replanning occurs, the revised plan is presented with an explicit rationale. The replan is a direct and effective response to the observed triggers (e.g., errors, new information) and learns from prior attempts by not repeating problematic steps.

    Middle scores: The plan generally addresses the query and appears feasible. Minor issues may be present: some steps lack explicit justification, a few steps may be unnecessary or unclear, or non-critical actions may be missing. The step order or rationale might be partially implied rather than fully articulated. Most steps in the plan could be feasibly executed by the tools provided. If replanning occurs, the rationale is vague or weakly connected to the trigger. The replan partially addresses the trigger but may be inefficient or repeats minor errors from the previous plan.

    0: The plan fails to directly address the user's query or cannot feasibly accomplish the goal. Critical steps in the plan are missing, irrelevant, unsupported, or based on fabricated reasoning. Replanning (if any) is arbitrary, unexplained, or disconnected from observable evidence in prior context. The overall plan lacks adequate justification and transparency, with major gaps or unjustified assertions. Many steps in the plan cannot be feasibly executed by the tools provided. If replanning occurs, it is arbitrary, unexplained, or disconnected from any trigger. The replan fails to address the issue and repeats the same critical mistakes as the previous attempt.

Be critical in your evaluation. For each step in the plan that is not necessary, unclear, or unsupported, identify that step and explain the problem specifically.
\end{lstlisting}

\subsection{PQ Final Prompt (POST-GEPA)}

\begin{lstlisting}
You are evaluating the quality of an agent's initial plan (and any replans) for solving a task. Your goal is to identify inherent flaws in the plan itself -- NOT to evaluate how well the plan was executed.

Core Evaluation Framework
1. Tool Selection and Usage:
    - Does the plan specify using tools that do not exist in the available toolset?
    - Does the plan ignore more appropriate or efficient tools that were available?
    - For research-heavy tasks, does the plan appropriately call for using search/lookup tools rather than assuming information is already known?
    - Does the plan correctly understand tool capabilities and limitations?
    - Critical: Does the plan explicitly specify which tools should be used for each step that requires external information or computation?

2. Pre-existing Knowledge:
    - Does the plan include redundant steps to find information already present in the task prompt or conversation history?
    - Does the plan appropriately leverage information from prior fact-finding or exploration?

3. Step Necessity and Completeness:
    - Are all steps necessary, or are some redundant?
    - Are critical steps missing that would be required to solve the task?
    - Does the plan make unjustified logical leaps between steps?
    - Are prerequisite steps properly identified before dependent steps?

4. Logical Structure:
    - Is the step order logical and efficient?
    - Are dependencies between steps properly sequenced?
    - Does each step build appropriately on previous steps?

Critical Distinction: Plan Quality vs. Execution Quality
You are judging the STRATEGY, not the OUTCOME. A plan can be excellent even if execution fails, and a plan can be flawed even if execution somehow succeeds.

What to evaluate:
    - The written plan text itself.
    - Whether steps are theoretically executable with available tools.
    - Whether the strategy is sound given the constraints and resources.
    - Whether tool usage is explicitly specified where needed.

What NOT to evaluate:
    - Whether the agent actually called the tools mentioned in the plan.
    - Whether tool outputs were correct or useful.
    - Whether the final answer was right or wrong.
    - Agent actions during execution that deviate from the plan.

Special Note on Tool Specification
A common critical flaw is when a plan describes WHAT needs to be done without specifying WHICH TOOL should do it.

Example:
    - Flawed: "Retrieve the commit history from GitHub"
    - Better: "Use search_agent to find GitHub commit history"

When a plan lacks explicit tool specifications for research or computation steps, the agent may attempt to solve steps through reasoning alone, fabricate information, or fail to execute the step at all. Always check whether research-intensive steps explicitly name the tool (like search_agent) that will perform the work.

Special Note on Information Gathering
When a plan includes research steps, evaluate whether:
    - The plan specifies an appropriate tool/method for that research.
    - The planned approach is feasible given available tools.
    - The plan properly sequences prerequisite lookups before dependent analysis.

Do NOT penalize a plan because the agent later failed to execute the research steps. The plan saying "use search_agent to find X" is sufficient.

Replanning Analysis
If replans exist:
    - Identify what triggered the replan (tool errors, missing info, failed searches).
    - Evaluate whether the replan rationale addresses the actual problem.
    - Judge whether the new plan is a logical response to the trigger.
    - Check if the replan avoids repeating the same mistakes.

Output Structure
Your response MUST use this exact format:

Initial Plan Identification
[Paste the initial plan verbatim, or state "I cannot find a plan."]

Replan Identification (if applicable)
[For each replan: paste it verbatim and state the written rationale]

Plan Quality Analysis
[Analyze ONLY the plan text and rationale. Pay special attention to whether tools are explicitly specified for each step requiring external information.]

Verdict on Plan Flaws
[List only actual flaws in the plan itself: wrong tools, missing steps, redundant actions, logical errors, lack of explicit tool specification, etc.]

Scoring
Assign a score from 0-3 based SOLELY on the intrinsic quality of the plan(s):

    Score 3: Plan is optimal; well-structured, uses correct tools with explicit specification, all steps necessary, and logical order. Replans have clear rationale.
    
    Score 2: Plan is generally sound but has minor issues; some steps lack tool specification or have slightly unclear ordering.
    
    Score 1: Plan has significant problems; missing critical steps, poor tool selection, unclear logic, or frequent lack of explicit tool specification.
    
    Score 0: Plan fails to address the task, has major logical flaws, proposes nonexistent tools, or is based on fabricated reasoning.

Important Reminders
    - Focus on the PLAN, not the EXECUTION.
    - Evaluate whether steps COULD be executed with available tools, not whether they WERE.
    - Don't let a correct or incorrect final answer influence your evaluation.
    - Pay special attention to whether research steps explicitly specify which tool will perform the research.

Explain the concrete execution risk posed by a lack of tool specification.
\end{lstlisting}


\subsection{TS Starting Prompt (PRE-GEPA): Generic TS Criteria}

\begin{lstlisting}
You are a meticulous TOOL SELECTION evaluator. Judge whether the agent chose the right tools for its tasks given the tool descriptions.

You must assign a single numerical score from 0 to 3, where 0 is the lowest score according to the criteria and 3 is the highest possible score.

Evaluation criteria:

    Score the appropriateness of tool SELECTION decisions relative to stated goals and available tools.
    
    3: Consistently selects the most suitable tools for each subtask, honors mandated tools, avoids tools when internal reasoning suffices, and reflects awareness of tool capabilities/limits.
    
    Middle scores: Generally appropriate selections with occasional missed opportunities (better tool existed), unnecessary tool choices for internal tasks, or weak justification.
    
    0: Frequently selects ill-suited/irrelevant tools, ignores mandated tools, or bypasses obviously superior tools; relies on non-tools where a tool is necessary.
    
    Consider: match-to-goal, comparative suitability, instruction compliance, and awareness of constraints. Do NOT judge call syntax, output interpretation, efficiency, or adherence.
      
Important scope boundaries:
    - Do NOT penalize call syntax/semantics or output interpretation (Tool Calling).
    - Do NOT penalize workflow efficiency (Execution Efficiency) or plan deviations (Plan Adherence).
    - Focus strictly on selection quality per subtask.

Be critical. For each selection issue, cite the relevant spans and explain specifically.

You must structure your response exactly as specified in the provided tool_selection_prompt.
\end{lstlisting}

\subsection{TS Final Prompt (POST-GEPA)}

\begin{lstlisting}[mathescape=true]
You are a meticulous TOOL SELECTION evaluator. Your task is to judge whether an AI agent chose the right tools for its subtasks given the available tool descriptions and stated goals.

Scope and Focus
You must assign a single numerical score from 0 to 3 based SOLELY on tool selection appropriateness:

    Score 3: Consistently selects the most suitable tools for each subtask, honors mandated tools, avoids tools when internal reasoning suffices, and reflects awareness of tool capabilities/limits.
    
    Score 2: Generally appropriate selections with occasional missed opportunities (better tool existed), unnecessary tool choices for internal tasks, or weak justification.
    
    Score 1: Multiple instances of suboptimal tool selection, but some core tools are used appropriately.
    
    Score 0: Frequently selects ill-suited/irrelevant tools, ignores mandated tools, or bypasses obviously superior tools; relies on non-tools where a tool is necessary.

What to Evaluate
Consider these factors for each tool selection decision:
    - Match-to-goal: Does the tool's purpose align with the subtask requirements?
    - Comparative suitability: Was there a clearly better tool available?
    - Instruction compliance: Did the agent follow explicit tool requirements in the task?
    - Awareness of constraints: Did the agent recognize tool limitations and capabilities?

Critical Scope Boundaries
DO NOT penalize:
    - Tool call syntax/semantics errors (e.g., incorrect argument formatting)
    - Output interpretation failures (misunderstanding tool responses)
    - Workflow efficiency issues (too many steps, redundant calls)
    - Plan adherence deviations (execution differing from stated plan)

DO penalize:
    - Using Tool A when Tool B is explicitly required or obviously superior
    - Attempting tasks internally when a tool exists for that purpose
    - Using tools for tasks that require no external tools
    - Ignoring available tools that directly solve the stated problem

Understanding Tool System Architecture
Many agent systems enforce strict separation between code execution and external resource access. In such architectures:

Tools serve as exclusive interfaces for certain operations:
    - File reading must go through designated file tools (like inspect_file_as_text), not direct Python file operations
    - Web searches must use search tools (like search_agent), not direct HTTP requests in code
    - External data retrieval must use appropriate delegation tools

Recognizing architectural violations:
When you see agents writing Python code that directly:
    - Opens/reads files (open(), pd.read_csv() on external paths)
    - Makes web requests (requests.get(), urllib)
    - Accesses databases or external APIs

Check whether tools exist for these operations. If they do, the direct code approach likely represents a tool selection error, even if the code is syntactically valid.

Key principle: The error is not whether the code works - it is whether the agent bypassed the tool ecosystem.

Tool Invocation vs. Tool Preparation
A critical distinction in evaluating tool usage:
Tool Preparation (not tool usage):
    - Creating string variables describing what to search for
    - Printing task descriptions
    - Formatting parameters for future use
    - Planning to use a tool

Tool Execution (actual tool usage):
    - Calling the tool function with arguments: result = tool_name(arg=value)
    - Receiving and processing tool output
    - Using returned data in subsequent operations

Example pattern:
This is preparation, NOT execution:
    search_task = "Find information about X"
    print(search_task)
This is execution:
    result = search_agent(task=search_task)
If an agent prepares tool calls but never executes them, this represents complete tool omission.

Multi-Phase Task Analysis
    1. Examine Planning Quality First: Check if the plan specifies WHICH tools to use for each step.
    2. Distinguish Planning vs. Execution Errors: Execution errors often have planning-level root causes.
    3. Identify Plan-Execution Mismatches: Note when plans specify tools but execution uses different ones.
    4. Check Multiple Trace Locations: Scan planning, fact-gathering, execution, and verification phases.

Error Categorization Framework
1. Wrong Tool for Subtask
The agent had access to Tool X but used Tool Y instead.

2. Omission of Required Tools
The agent never invoked a tool that was mandated or obviously necessary. (Watch for preparation without execution).

3. Tool Use Without Need
The agent called tools for tasks solvable through internal reasoning (e.g., using search for 2+2).

4. Fabrication/Assumption Instead of Tools
The agent claims knowledge from sources but never verified using available tools. Watch for "simulate," "assume," or "for demonstration."

5. Plan Non-Compliance
The agent's plan specifies Tool A, but execution uses Tool B without explanation.

6. Architectural Violations
The agent attempts to perform operations directly in code (e.g., pd.read_csv) when tools are mandated for those operations.

7. Persistent Misuse After Failure
The agent repeatedly tries the same failing tool instead of pivoting to alternatives.

Detecting Data Fabrication
Systematic verification process:
    1. Identify all specific data values in the agent's answer.
    2. Trace backward: For each value, find where it originated.
    3. Check tool outputs: Did a tool call produce this value?
    4. Flag fabrication: If no tool output can be found, the data is fabricated.

Data Flow Analysis
For each critical data value, trace its provenance:
    Final Answer: X
    |-- Where did X come from?
    |   |-- Tool call at step N? -> Legitimate
    |   |-- Computation from tool data? -> Legitimate
    |   |-- Hardcoded variable? -> Check source
    |   |   |-- Based on tool output? -> Legitimate
    |   |   |-- Assumed/invented? -> FABRICATION
    |   |-- Stated without source? -> FABRICATION

Critique Structure
    1. Overview: Brief summary of the task and available tools
    2. Critical Analysis by Phase: Examine planning, then execution
    3. Error-by-Error Breakdown: Quote the trace, identify the error, specify the correct tool, and assess severity.
    4. Root Cause Analysis: Trace execution errors back to planning deficiencies.
    5. Positive Selections: Note correct tool choices for balance.
    6. Score Justification: Explain your score with reference to the rubric.

Severity Calibration Guidelines

    Critical (Score 0): Complete omission of tool category; systematic fabrication; architectural violations; ignoring mandated tools.
    
    Moderate (Score 1-2): Suboptimal tool choice but task-appropriate tools used; incomplete extraction; plan-execution mismatch with some conformance.
    
    Minor: Inefficient but valid tool sequences; over-reliance on tools for simple tasks; redundant calls.

Adaptive Recovery as Positive Signal
When an agent encounters tool failures but successfully adapts by switching tools or refining queries, this demonstrates strong tool selection awareness. Weight this heavily in positive selections.
\end{lstlisting}

\subsection{TC Starting Prompt (PRE-GEPA): Generic TC Criteria}

\begin{lstlisting}
You are a meticulous TOOL CALLING evaluator. Judge how well the agent formed tool inputs and interpreted outputs, given tool definitions.

You must assign a single numerical score from 0 to 3, where 0 is the lowest score according to the criteria and 3 is the highest possible score.

Evaluation criteria:

    Score the quality of TOOL CALLS within the agent's control.
    
    3: Inputs are syntactically valid and semantically appropriate; required params and preconditions are satisfied; outputs are interpreted faithfully and integrated correctly; tool-returned errors are acknowledged and handled reasonably.
    
    Middle scores: Minor issues with argument completeness, semantic underspecification, limited reformulation, or shallow/partial output use; some missed acknowledgements of errors.
    
    0: Invalid/missing arguments, repeated schema violations, semantically off-target queries without correction; outputs ignored/misread/fabricated; tool errors unacknowledged.
    
    Consider only what is under the agent's control. Do NOT judge tool choice (Tool Selection), workflow efficiency, or external system reliability (Tool Quality).

Important scope boundaries:
    - In-scope: argument/schema correctness, semantic fit of query, preconditions/postconditions, grounded interpretation of outputs, explicit handling of tool-returned errors.
    - Out-of-scope: tool selection (Tool Selection), workflow efficiency (Execution Efficiency), external service/tool reliability (Tool Quality).

Be critical. For each calling issue, cite the relevant spans and explain specifically.

You must structure your response exactly as specified in the provided tool_calling_prompt.
\end{lstlisting}

\subsection{TC Final Prompt (POST-GEPA)}
\begin{lstlisting}
You are a meticulous TOOL CALLING evaluator. Judge how well the agent formed tool inputs and interpreted outputs, given tool definitions.

You must assign a single numerical score from 0 to 3, where 0 is the lowest score according to the criteria and 3 is the highest possible score.

Evaluation criteria:

    Score the quality of TOOL CALLS within the agent's control.

    Score 3: Inputs are syntactically valid and semantically appropriate; required params and preconditions are satisfied; outputs are interpreted faithfully and integrated correctly; tool-returned errors are acknowledged and handled reasonably. Importantly, all claims about verification or confirmation must be grounded in actual tool outputs -- no fabricated certainty.
    
    Middle scores: Minor issues with argument completeness, semantic underspecification, limited reformulation, or shallow/partial output use; some missed acknowledgements of errors; or unverified claims presented as confirmed facts.
    
    Score 0: Invalid/missing arguments, repeated schema violations, semantically off-target queries without correction; outputs ignored/misread/fabricated; tool errors unacknowledged; or systematic fabrication of verification that never occurred.

Consider only what is under the agent's control. Do NOT judge tool choice (Tool Selection), workflow efficiency, or external system reliability (Tool Quality).

Important scope boundaries:
    - In-scope: argument/schema correctness, semantic fit of query, preconditions/postconditions, grounded interpretation of outputs, explicit handling of tool-returned errors, distinguishing between what tools returned vs. what agents claim about those returns.
    - Out-of-scope: tool selection (Tool Selection), workflow efficiency (Execution Efficiency), external service/tool reliability (Tool Quality).

Critical evaluation patterns to watch for:
1. Fabricated Verification: When an agent claims information is "confirmed," "verified," "firmly identified," or "established" without actually visiting authoritative sources or receiving explicit confirmation from tools. Distinguish between:
    - Agent noting what a search snippet says (acceptable)
    - Agent claiming this information is verified/confirmed without visiting the source (fabrication)
    
2. Ungrounded Certainty: When an agent presents inferred information as validated fact. Check if the agent:
    - Received only search result snippets but claims "multiple archival sources confirm..."
    - States facts are "firmly identified" based solely on brief mentions in search results
    - Asserts composer/author identity without visiting the actual pages that would contain definitive information

3. Incomplete Execution vs. False Claims: Distinguish between:
    - Skipping verification steps in a plan (process issue - moderate severity)
    - Skipping verification AND claiming it occurred anyway (fabrication - high severity)

4. Syntax and Semantic Issues:
    - Check for syntactically invalid Python code (unescaped quotes in strings, malformed arguments)
    - Verify query construction does not include meta-instructions as literal search terms
    - Ensure arguments match tool schemas exactly
    - Pay special attention to tools that take NO arguments - agents often incorrectly try to pass empty dictionaries, empty strings, or placeholder arguments

Error Handling Quality:
    - Did the agent acknowledge tool-returned errors explicitly?
    - Did reformulation address the root cause of the error?
    - Were error messages used to improve subsequent calls?
    - Track whether agents repeat the exact same error multiple times after receiving error feedback

Output Interpretation:
    - Are tool outputs used faithfully, or does the agent add unsupported details?
    - When tools return limited information, does the agent acknowledge limitations or proceed as if complete information was obtained?
    - Does the agent distinguish between "the tool returned X" and "X is confirmed to be true"?

Error Type Categorization:
    - Formatting errors: Incorrect syntax in a single tool call
    - Learning failures: Repeating the same error after receiving explicit feedback
    - Schema violations: Passing arguments to tools that do not accept them

Balanced Assessment:
    - Acknowledge when agents correctly recover from errors
    - Note partial successes and appropriate tool usage patterns
    - Recognize creative problem-solving within tool limitations
    - Provide specific guidance on what correct usage would look like

Be critical but fair. For each calling issue, cite the relevant spans and explain specifically what evidence (or lack thereof) supports your assessment. Pay special attention to the gap between what agents claim about information quality and what tools actually provided, repeated errors, and tools that accept no arguments being called incorrectly.

You must structure your response exactly as specified in the provided tool_calling_prompt.
\end{lstlisting}

\section{Automatic Consistency Optimization with Openevolve} \label{appendix-openevolve-prompts}

In this section we share the prompts of every GPA metric before and after Openevolve optimization.

\subsection{PQ Starting Prompt (Pre-OpenEvolve): Generic PQ Criteria}
\begin{lstlisting}
You are a meticulous and analytical PLAN QUALITY evaluator. 

You are responsible for evaluating the intrinsic quality of the initial written plan, judging it against the context and tools available at the moment of its creation. CRITICAL: It is an immediate failure of your task to reference whether the agent followed the plan or mention any part of the execution, including agent actions, tool outputs, or the final answer.

Plan Extraction Procedure:
    1. Scan for the sections labeled with a PLAN keyword. The first section labeled with a PLAN keyword is the initial plan, and any subsequent section labeled with a PLAN keyword is a replan.
    2. If no explicitly labeled PLAN section exists, infer the plan from any 'Thinking' or planning sections [or to-do checklist].
    3. If no plan can be found through the above steps, output: "I cannot find a plan."
    Do NOT infer or fill gaps using execution steps.

Evaluating the Initial Plan:
    1. The Available Tools: Does the plan correctly select from the list of provided tools? Does it ignore a more appropriate or efficient tool that was available? Does it try to use a tool that doesn't exist?
    2. Tool Definitions: Does the plan propose using a tool correctly, according to its description and required arguments?
    3. Pre-existing Knowledge: Does the plan include redundant steps to find information that was already present in the initial prompt or conversation history? Does the plan include relevant information from fact-finding or exploration prior to planning?
    4. An optimal plan isn't just logical in theory; it's the most intelligent strategy given the specific resources the planner had.
    When evaluating the initial plan, ignore all execution steps, tool outputs, and agent actions, even if available and visible in the trace. Your quality evaluation for this initial plan MUST be based solely on its intrinsic quality. You are judging the strategy, not the outcome. Never use agent choices, answers, or deviations from the plan to deduce flaws, gaps, or weaknesses in the plan itself.

Replanning (if found):
    1. Look at the tool outputs, error messages, or observations in the trace that precede the replan to understand why replanning was necessary.
    2. Identify the trigger and explain why the original plan was insufficient. Is the reason for replanning justified?
    3. Judge the new plan. Are they a logical, necessary, and efficient correction to the specific problem identified in the trigger? You are not judging the original failure itself, but the quality of the agent's reaction to that failure.

List only inherent plan flaws (e.g., step uses nonexistent tool, redundant action, ignores key context).
You MUST structure your entire response using the following markdown template:
-----
**Initial Plan Identification**
[Paste initial plan or state: 'I cannot find a plan.']

For each replan (if exists):
**Replan Identification**
[Paste each replan. For each replan, state the written rationale/explanation.]

**Plan Quality Analysis**
[Analysis solely on plan/replan text and rationale.]

**Verdict on Plan Flaws**
[List only actual flaws in the plans themselves.]
-----
You must assign a single numerical score from 0 to 3, where 0 is the lowest score according to the criteria and 3 is the highest possible score based SOLELY on the intrinsic quality of the plan and replans. Do NOT score on the execution quality.

Evaluation criteria:

    Score the quality of the plan.

    3: The plan is well-structured, optimal, and directly addresses the user's query by breaking it down into clear, actionable, and logical steps. Every step is justified, necessary, and includes sufficient detail to ensure feasibility and efficiency without being overly verbose. Each step in the plan could be feasibly executed by the tools provided. If replanning occurs, the revised plan is presented with an explicit rationale. The replan is a direct and effective response to the observed triggers (e.g., errors, new information) and learns from prior attempts by not repeating problematic steps.

    Middle scores: The plan generally addresses the query and appears feasible. Minor issues may be present: some steps lack explicit justification, a few steps may be unnecessary or unclear, or non-critical actions may be missing. The step order or rationale might be partially implied rather than fully articulated. Most steps in the plan could be feasibly executed by the tools provided. If replanning occurs, the rationale is vague or weakly connected to the trigger. The replan partially addresses the trigger but may be inefficient or repeats minor errors from the previous plan.

    0: The plan fails to directly address the user's query or cannot feasibly accomplish the goal. Critical steps in the plan are missing, irrelevant, unsupported, or based on fabricated reasoning. Replanning (if any) is arbitrary, unexplained, or disconnected from observable evidence in prior context. The overall plan lacks adequate justification and transparency, with major gaps or unjustified assertions. Many steps in the plan cannot be feasibly executed by the tools provided. If replanning occurs, it is arbitrary, unexplained, or disconnected from any trigger. The replan fails to address the issue and repeats the same critical mistakes as the previous attempt.

Be critical in your evaluation. For each step in the plan that is not necessary, unclear, or unsupported, identify that step and explain the problem specifically.
\end{lstlisting}

\subsection{PQ Final Prompt (Post-Openevolve)}
\begin{lstlisting}[mathescape=true]
You are a meticulous and analytical PLAN QUALITY evaluator. You are responsible for evaluating the intrinsic quality of the initial written plan, judging it against the context and tools available at the moment of its creation. CRITICAL: It is an immediate failure of your task to reference whether the agent followed the plan or mention any part of the execution, including agent actions, tool outputs, or the final answer.

Plan Extraction Procedure:
1. Scan for the sections labeled with a PLAN keyword. The first section labeled with a PLAN keyword is the initial plan, and any subsequent section labeled with a PLAN keyword is a replan.
2. If no explicitly labeled PLAN section exists, infer the plan from any 'Thinking' or planning sections [or to-do checklist].
3. If no plan can be found through the above steps, output: "I cannot find a plan."
Do NOT infer or fill gaps using execution steps.

Evaluating the Initial Plan:
1. The Available Tools: Does the plan correctly select from the list of provided tools? Does it ignore a more appropriate or efficient tool that was available? Does it try to use a tool that doesn't exist?
2. Tool Definitions: Does the plan propose using a tool correctly, according to its description and required arguments?
3. Pre-existing Knowledge: Does the plan include redundant steps to find information that was already present in the initial prompt or conversation history? Does the plan include relevant information from fact-finding or exploration prior to planning?
4. An optimal plan isn't just logical in theory; it's the most intelligent strategy given the specific resources the planner had.
When evaluating the initial plan, ignore all execution steps, tool outputs, and agent actions, even if available and visible in the trace. Your quality evaluation for this initial plan MUST be based solely on its intrinsic quality. You are judging the strategy, not the outcome. Never use agent choices, answers, or deviations from the plan to deduce flaws, gaps, or weaknesses in the plan itself.

Replanning (if found):
1. Look at the tool outputs, error messages, or observations in the trace that precede the replan to understand why replanning was necessary.
2. Identify the trigger and explain why the original plan was insufficient. Is the reason for replanning justified?
3. Judge the new plan. Are they a logical, necessary, and efficient correction to the specific problem identified in the trigger? You are not judging the original failure itself, but the quality of the agent's reaction to that failure.

List only inherent plan flaws (e.g., step uses nonexistent tool, redundant action, ignores key context).
You MUST structure your entire response using the following markdown template:
-----
**Initial Plan Identification**
[Paste initial plan or state: 'I cannot find a plan.']

For each replan (if exists):
**Replan Identification**
[Paste each replan. For each replan, state the written rationale/explanation.]

**Plan Quality Analysis**
[Analysis solely on plan/replan text and rationale.]

**Verdict on Plan Flaws**
[List only actual flaws in the plans themselves.]
-----
You must assign a single numerical score from 0 to 3, where 0 is the lowest score according to the criteria and 3 is the highest possible score based SOLELY on the intrinsic quality of the plan and replans. Do NOT score on the execution quality.

Score 3.0 (Excellent): Plan meets ALL essential criteria:
- Directly addresses user query with logical, well-ordered steps
- All steps are necessary, actionable, and executable with available tools
- Sufficient detail for clear execution without excessive verbosity
- If replanning: explicit rationale provided and incorporates lessons from failures
- No critical steps missing; no irrelevant steps included

Score 2.0-2.9 (Good): Plan addresses query well but has 1-3 minor issues:
- One step could be clearer or better justified (2.7-2.9)
- Minor inefficiency in step ordering but still functional (2.4-2.6)
- One unnecessary but harmless step included (2.1-2.3)
- Weak replanning rationale but shows some learning (2.0-2.2)

Score 1.0-1.9 (Adequate): Plan attempts to address query but has notable problems:
- Multiple steps lack sufficient detail or justification (1.7-1.9)
- Some questionable step feasibility with available tools (1.4-1.6)
- Missing 1-2 important steps or includes several unnecessary ones (1.1-1.3)
- Poor logical flow or significant ordering issues (1.0-1.2)

Score 0.0-0.9 (Poor): Plan has critical flaws:
- Fails to address the user query or cannot accomplish goal (0.7-0.9)
- Multiple critical steps missing or based on fabricated reasoning (0.4-0.6)
- Majority of steps infeasible with available tools (0.1-0.3)
- Replanning arbitrary/unexplained or repeats same critical errors (0.0)

EVALUATION PROTOCOL:
1. CONSISTENCY FIRST: Use decimal scores within ranges for precision
2. PLAN-ONLY FOCUS: Evaluate plan quality, not execution success or tool failures
3. TOOL FEASIBILITY: Verify each step can realistically use available tools mentioned in trace
4. REPLANNING WEIGHT: Good replans explicitly state what changed and why
5. BOUNDARY GUIDANCE: When uncertain between score ranges, choose lower range
6. NO-PLAN CASE: If no plan found, score 0.0 and note "No plan found in span X"

ACCURACY CALIBRATION:
- Score 2.5+ reserved for plans that would likely succeed if executed properly
- Score 1.5-2.4 for plans with structural issues but basic approach sound
- Score 0.5-1.4 for plans with significant gaps but some merit
- Score 0.0-0.4 for fundamentally flawed or missing plans

CONSISTENCY ANCHORS:
- Perfect plans with all criteria met: 3.0
- Solid plans with 1 minor flaw: 2.7
- Good plans with 2-3 minor issues: 2.2
- Adequate plans with notable problems: 1.5
- Poor plans with critical flaws: 0.5

\end{lstlisting}

\subsection{TC Starting Prompt (Pre-OpenEvolve): GEPA optimized TC prompt}
\begin{lstlisting}
You are a meticulous TOOL CALLING evaluator. Judge how well the agent formed tool inputs and interpreted outputs, given tool definitions.

You must assign a single numerical score from 0 to 3, where 0 is the lowest score according to the criteria and 3 is the highest possible score.

Evaluation criteria:

    Score the quality of TOOL CALLS within the agent's control.

    Score 3: Inputs are syntactically valid and semantically appropriate; required params and preconditions are satisfied; outputs are interpreted faithfully and integrated correctly; tool-returned errors are acknowledged and handled reasonably. Importantly, all claims about verification or confirmation must be grounded in actual tool outputs -- no fabricated certainty.
    
    Middle scores: Minor issues with argument completeness, semantic underspecification, limited reformulation, or shallow/partial output use; some missed acknowledgements of errors; or unverified claims presented as confirmed facts.
    
    Score 0: Invalid/missing arguments, repeated schema violations, semantically off-target queries without correction; outputs ignored/misread/fabricated; tool errors unacknowledged; or systematic fabrication of verification that never occurred.

Consider only what is under the agent's control. Do NOT judge tool choice (Tool Selection), workflow efficiency, or external system reliability (Tool Quality).

Important scope boundaries:
    - In-scope: argument/schema correctness, semantic fit of query, preconditions/postconditions, grounded interpretation of outputs, explicit handling of tool-returned errors, distinguishing between what tools returned vs. what agents claim about those returns.
    - Out-of-scope: tool selection (Tool Selection), workflow efficiency (Execution Efficiency), external service/tool reliability (Tool Quality).

Critical evaluation patterns to watch for:
1. Fabricated Verification: When an agent claims information is "confirmed," "verified," "firmly identified," or "established" without actually visiting authoritative sources or receiving explicit confirmation from tools. Distinguish between:
    - Agent noting what a search snippet says (acceptable)
    - Agent claiming this information is verified/confirmed without visiting the source (fabrication)
    
2. Ungrounded Certainty: When an agent presents inferred information as validated fact. Check if the agent:
    - Received only search result snippets but claims "multiple archival sources confirm..."
    - States facts are "firmly identified" based solely on brief mentions in search results
    - Asserts composer/author identity without visiting the actual pages that would contain definitive information

3. Incomplete Execution vs. False Claims: Distinguish between:
    - Skipping verification steps in a plan (process issue - moderate severity)
    - Skipping verification AND claiming it occurred anyway (fabrication - high severity)

4. Syntax and Semantic Issues:
    - Check for syntactically invalid Python code (unescaped quotes in strings, malformed arguments)
    - Verify query construction does not include meta-instructions as literal search terms
    - Ensure arguments match tool schemas exactly
    - Pay special attention to tools that take NO arguments - agents often incorrectly try to pass empty dictionaries, empty strings, or placeholder arguments

Error Handling Quality:
    - Did the agent acknowledge tool-returned errors explicitly?
    - Did reformulation address the root cause of the error?
    - Were error messages used to improve subsequent calls?
    - Track whether agents repeat the exact same error multiple times after receiving error feedback

Output Interpretation:
    - Are tool outputs used faithfully, or does the agent add unsupported details?
    - When tools return limited information, does the agent acknowledge limitations or proceed as if complete information was obtained?
    - Does the agent distinguish between "the tool returned X" and "X is confirmed to be true"?

Error Type Categorization:
    - Formatting errors: Incorrect syntax in a single tool call
    - Learning failures: Repeating the same error after receiving explicit feedback
    - Schema violations: Passing arguments to tools that do not accept them

Balanced Assessment:
    - Acknowledge when agents correctly recover from errors
    - Note partial successes and appropriate tool usage patterns
    - Recognize creative problem-solving within tool limitations
    - Provide specific guidance on what correct usage would look like

Be critical but fair. For each calling issue, cite the relevant spans and explain specifically what evidence (or lack thereof) supports your assessment. Pay special attention to the gap between what agents claim about information quality and what tools actually provided, repeated errors, and tools that accept no arguments being called incorrectly.

You must structure your response exactly as specified in the provided tool_calling_prompt.
\end{lstlisting}

\subsection{TC Final Prompt (Post-OpenEvolve)}
\begin{lstlisting}
You are a meticulous TOOL CALLING evaluator. Judge how well the agent formed tool inputs and interpreted outputs, given tool definitions.

You must assign a single numerical score from 0 to 3, where 0 is the lowest score according to the criteria and 3 is the highest possible score.

Score the quality of TOOL CALLS using these STRICT QUANTITATIVE criteria:

**Score 3** (ALL must be true):
- No critical or major issues
- Maximum ONE minor issue with immediate self-correction
- 100% error acknowledgment AND handling
- 100% output interpretation accuracy
- All tool inputs syntactically AND semantically valid
- All required parameters present and correct

**Score 2** (ANY of these exact conditions):
- Exactly 2-3 minor issues total, no major/critical
- Exactly 1 major issue that was fully self-corrected
- Error acknowledgment rate 75-99% with recovery
- Output interpretation accuracy 75-99% with recovery
- 2-3 parameter issues that don't block completion

**Score 1** (ANY of these exact conditions):
- Exactly 4-5 minor issues, no critical
- Exactly 2 major issues, self-corrected
- Error acknowledgment rate 50-74%
- Output interpretation accuracy 50-74%
- 4+ parameter issues but task completed

**Score 0** (ANY of these conditions):
- 6+ minor issues OR 3+ major issues
- ANY unresolved critical issue
- Error acknowledgment below 50%
- Output interpretation below 50%
- Blocking parameter/schema violations

**Issue Severity Definitions**:
CRITICAL (blocks completion):
- Invalid required parameter schema
- Complete output misinterpretation
- Result fabrication
- Unhandled blocking errors

MAJOR (impairs completion):
- Incorrect parameter values
- Partial output misinterpretation
- Inadequate error handling
- Failed recovery attempts

MINOR (affects quality):
- Optional parameter omission
- Delayed error recognition
- Suboptimal parameter choice
- Imperfect output parsing

**Scope Definition**:
IN-SCOPE:
- Parameter correctness
- Schema compliance
- Output interpretation
- Error handling
- Self-correction

OUT-OF-SCOPE:
- Tool selection
- Workflow efficiency
- External reliability
- Task completion speed


**Mandatory Evaluation Process**:

1. **Issue Counting Rules**
   - Each distinct issue counts ONCE
   - Track exact issue counts by severity
   - Self-corrected issues count as 0.5x
   - Failed fixes count as new issues
   - Group related issues under root cause

2. **Required Calculations**
   For EVERY evaluation:
   [ ] Count total issues by severity
   [ ] Calculate error handling rate
   [ ] Calculate interpretation accuracy
   [ ] Track self-correction success
   [ ] Document evidence for each issue

3. **Scoring Algorithm**
   1. Start at Score 3
   2. Count issues by severity
   3. Apply self-correction discounts
   4. Calculate error/interpretation rates
   5. Map to exact score thresholds
   6. Default lower if borderline

4. **Evidence Requirements**
   For EACH issue document:
   - Exact step number(s)
   - Severity classification
   - Direct quote/evidence
   - Impact description
   - Resolution status

5. **Common Issue Patterns**
   Repository Analysis:
   - Character iteration = MAJOR (2 points)
   - Wrong separator = MINOR (1 point)
   - Size violation = CRITICAL (4 points)
   - Bad regex = MAJOR (2 points)
   - No validation = MINOR (1 point)

6. **Response Structure**
   A. Issue Summary
      - Critical count: N
      - Major count: N
      - Minor count: N
      - Self-corrections: N
      - Total weighted points: X

   B. Rate Calculations
      - Error handling: X%
      - Interpretation: X%
      - Self-correction: X%

   C. Evidence Index
      For each issue:
      - Steps: [list]
      - Severity: [category]
      - Quote: [evidence]
      - Impact: [description]
      - Resolution: [status]

   D. Score Determination
      - Initial score: [3]
      - Deductions: [list]
      - Final score: [0-3]
      - Threshold: [quote criteria]

**Critical Rules**:

1. ONLY integer scores (0,1,2,3)
2. Use CONCRETE evidence only
3. Apply criteria CONSISTENTLY
4. Document ALL calculations
5. Default to lower score
6. Group related issues
7. Count distinct problems
8. Consider recovery impact

**Validation Checklist**:

1. Parameter Validation
   [ ] Schema correctness
   [ ] Value appropriateness
   [ ] Completeness
   [ ] Type matching

2. Output Processing
   [ ] Complete extraction
   [ ] Accurate interpretation
   [ ] Proper validation
   [ ] Error handling

3. Error Management
   [ ] Explicit acknowledgment
   [ ] Recovery attempt
   [ ] Success validation
   [ ] Pattern recognition

4. Instruction Compliance
   [ ] Size limits
   [ ] Format requirements
   [ ] Prohibited operations
   [ ] Documentation

Score based SOLELY on observable evidence.
Maintain strict objectivity and consistency.
\end{lstlisting}

\section{Snowflake Intelligence Results}

The full set of results on the internal Snowflake Intelligence dataset is shown in Table \ref{tab:lc-ee-comparison-full}. Accuracy is reported both as a binary 2-point score (error vs. correct) and a 3-point scale, along with correlation and normalized mean absolute error (NMAE). Performance results are shown across different LLM models.

\begin{table}[h]
\caption{Comparison of Logical Consistency and Execution Efficiency Across Models (Snowflake Intelligence)}
\label{tab:lc-ee-comparison-full}
\begin{center}
\resizebox{0.8\textwidth}{!}{%
\begin{tabular}{lcccccccc}
\toprule
\multicolumn{1}{c}{\bf Model} & \multicolumn{4}{c}{\bf LC} & \multicolumn{4}{c}{\bf EE} \\
\cmidrule(lr){2-5} \cmidrule(lr){6-9}
& \bf Acc-3pt & \bf Acc-2pt & \bf Correl & \bf NMAE & \bf Acc-3pt & \bf Acc-2pt & \bf Correl & \bf NMAE \\
\midrule
\bf Claude-4-Sonnet & \bf 0.765 & \bf 1.000 & \bf 0.795 & \bf 0.118 & \bf 0.882 & \bf 0.941 & \bf 0.772 & \bf 0.059 \\
Claude-3-7-Sonnet & 0.294 & 0.882 & 0.477 & 0.382 & 0.353 & 0.824 & 0.574 & 0.324 \\
gpt-4o            & 0.471 & 0.941 & 0.514 & 0.265 & 0.882 & 0.941 & 0.772 & 0.059 \\
gpt-4.1           & 0.294 & 0.882 & —     & 0.412 & 0.824 & 0.941 & 0.772 & 0.088 \\
\bottomrule
\end{tabular}%
}
\end{center}
\vskip -0.01in
\footnotesize{(Acc-3pt = 3-point scale Accuracy, Acc-2pt = 2-point scale Accuracy, Correl = Correlation, NMAE = Normalized Mean Absolute Error)}
\end{table}

Consistent with our findings on TRAIL/GAIA, LC remains the harder dimension, requiring complex reasoning that only Claude-4-Sonnet achieves reliably (at the time of our submission). By contrast, because execution efficiency-related errors may require less abstract thinking, multiple models (Claude-3-7-Sonnet, gpt-4o, and gpt-4.1) can reach similarly high performance.

\section{Experimental Case Study Setup} \label{appendix-gepa}

\subsection{GEPA Configuration}

Please refer to Section \ref{appendix-gepa-prompts} for comparisons between seed and GEPA-optimized prompts.

For Fig \ref{fig:trail-gaia-gepa-test} and Fig \ref{fig:trail-swe-bench-gepa-test}, we provide the following column descriptions. All GEPA optimization runs are performed using DSPy (\cite{khattab2023dspycompilingdeclarativelanguage}), and runs utilize default settings unless otherwise noted below.

\begin{enumerate}
    \item \textbf{Generic + custom with manual review}: Generic metric criteria appended with manually crafted custom instructions (described in \cref{gaia-methodology}), evaluation output graded by human annotators.
    \item \textbf{Generic with meta-judge}: Generic metric criteria with no custom instructions, evaluation output graded by a meta LLM judge.
    \item \textbf{Generic + custom with meta-judge}: Generic metric criteria appended with manually crafted custom instructions, evaluation output graded by a meta LLM judge.
    \item \textbf{GEPA (auto-light) with meta-judge}: GEPA-optimized prompt using DSPy's `light' auto-budget with generic metric criteria as initial seed (provided in \cref{appendix-gepa-prompts}), evaluation output graded by a meta LLM judge.
    \item \textbf{GEPA (auto-medium) with meta-judge}: GEPA-optimized prompt using DSPy's `medium' auto-budget with generic metric criteria as initial seed, evaluation output graded by a meta LLM judge.
\end{enumerate}

\subsection{OpenEvolve Configuration}

Please refer to \cref{appendix-openevolve-prompts} for comparisons between seed and OpenEvolve-optimized prompts.

OpenEvolve~(\cite{sharma2025openevolve}) is an open-source evolutionary coding agent inspired by AlphaEvolve~(\cite{novikov2025alphaevolve}). Rather than optimizing a prompt as a text string (as GEPA does via DSPy), OpenEvolve evolves \emph{source code}: specifically, a Python file containing two string variables---the scoring rubric (\texttt{CRITERIA}) and supplementary guidance (\texttt{ADDITIONAL\_INSTRUCTIONS})---delimited by \texttt{EVOLVE-BLOCK} markers. At each iteration the framework (i)~selects a parent program from a MAP-Elites quality-diversity archive with island-based populations, (ii)~prompts an LLM to produce a mutated variant of the evolvable code, and (iii)~evaluates the variant using a custom fitness function. Programs are stored in a population of size 30 across 2 islands, with an elite archive of size 15 and feature dimensions \texttt{mean\_gt\_deviation} and \texttt{avg\_std\_dev}. The exploitation ratio is 0.7, with an elite selection ratio of 0.2 and a similarity threshold of 0.95.

\paragraph{Evolution LLM.} Prompt mutations are generated by Claude~3.5~Sonnet (\texttt{temperature}$= 0.7$, \texttt{max\_tokens}$= 16{,}000$). Each metric's configuration file supplies a detailed system message describing the prompt structure, scoring function, evaluation scope, and code-output constraints so that the LLM is fully aware of the optimization objective.

\paragraph{Judge LLM.} All evaluations of candidate prompts are performed by Claude~4~Sonnet at \texttt{temperature}$= 0.0$ (via TruLens Cortex provider with \texttt{reasoning\_effort}$=$\texttt{high}), the same model used for the baseline reliability measurements.

\paragraph{Fitness function.} For each candidate prompt, 15 traces are sampled (fixed seed $= 42$) from the TRAIL/SWE-bench corpus. Each trace is evaluated 5 times (25 parallel workers) to obtain per-trace mean and standard deviation. The per-trace score is:
\[
  \text{base\_score}_t = \frac{1}{1 + \sigma_t}, \qquad
  \text{penalized\_score}_t = \text{base\_score}_t \times \text{penalty}(|\bar{s}_t - g_t|),
\]
where $g_t$ is the ground-truth mean score for trace $t$ (obtained from 10 initial-program runs over all 31 traces). The penalty factor is 1.0 when $|\bar{s}_t - g_t| \leq 0.2$, 0.5 when $|\bar{s}_t - g_t| \geq 0.5$, and linearly interpolated between. The fitness returned to OpenEvolve is $\texttt{combined\_score} = \frac{1}{T}\sum_t \text{penalized\_score}_t$, rewarding both low variance and fidelity to the initial prompt's score distribution.

\paragraph{Cascade evaluation.} To reduce wasted computation on syntactically invalid or degenerate prompts, we applied a two-stage cascade: Stage~1 evaluates a single trace with 2 runs (2 workers); only candidates exceeding a \texttt{combined\_score} threshold of 0.01 advance to Stage~2, which performs the full 15-trace, 5-run evaluation described above.

\paragraph{Seed prompts and variants.} We run OpenEvolve independently for each of the seed prompts in \cref{appendix-openevolve-prompts}

\begin{enumerate}
    \item \textbf{OpenEvolve (generic seed)}: Evolution seeded from the TruLens default scoring rubric for each metric (identical to the ``Generic'' baseline), with 50 iterations of OpenEvolve optimization..
    \item \textbf{OpenEvolve (GEPA seed)}: Evolution seeded from the GEPA-optimized prompt (DSPy \texttt{auto-medium}), decomposed into \texttt{CRITERIA} and \texttt{ADDITIONAL\_INSTRUCTIONS}. This variant tests whether OpenEvolve can further improve an already-optimized prompt. The \texttt{max\_code\_length} is increased from 15{,}000 to 25{,}000 to accommodate the longer GEPA-evolved instructions.
\end{enumerate}

\noindent All other settings (population size, archive size, number of islands, cascade thresholds, temperature, and judge model) are held constant across all six runs. Diff-based evolution is disabled to prevent merge-conflict markers in LLM output.

\section{Extended Future Work Directions}

Evaluation rubrics are critical for providing actionable feedback crucial for methods like GEPA (\cite{agrawal2025gepareflectivepromptevolution}) or MIPROv2 (\cite{opsahl2024optimizing}) to perform reflective prompt optimization. In a similar vein, systems like AlphaEvolve (\cite{novikov2025alphaevolve}) and OpenEvolve (\cite{sharma2025openevolve}) require evaluation feedback to drive their code evolution processes. By instrumenting our GPA evaluation rubrics within these evolutionary code systems, we may be able effectively leverage the quality of textual feedback to significantly improve the original underlying code of the agentic systems in use.

\section{Errata}

This section is intended to be updated with information regarding identification and remediation of data validation/formatting issues, if applicable.

\end{document}